%% file: cas-sc-template.tex
\documentclass[a4paper,11pt,fleqn]{cas-sc}
\input{math_cmds}

\usepackage[dvipsnames]{xcolor}
\usepackage{svg}
\usepackage{caption}
\usepackage{subcaption}
\usepackage[english]{babel}
\usepackage{amsthm}
\usepackage{mathtools}
\usepackage{xparse}
\usepackage{braket}
\usepackage{enumitem}
\usepackage{natbib}

\usepackage{soul}

\usepackage{algorithm}
\usepackage{algpseudocode}
\usepackage{algorithmicx}

\theoremstyle{plain}
\newtheorem{definition}{Definition}

\theoremstyle{plain}

\usepackage{booktabs}
\usepackage{threeparttable}
\usepackage{multirow}

\usepackage{tikz}
\usetikzlibrary{tikzmark}
\usepackage{graphicx}
\usetikzlibrary{arrows.meta}

\usepackage[toc,page]{appendix}

\def\tsc#1{\csdef{#1}{\textsc{\lowercase{#1}}\xspace}}
\tsc{WGM}
\tsc{QE}

\def \emphasize #1{{\textbf{\textit{#1}}}}

\def \revise #1{{#1}}

\def \SecondRevise #1{{#1}}

\begin{document}
\let\WriteBookmarks\relax
\def\floatpagepagefraction{1}
\def\textpagefraction{.001}

\shorttitle{Towards Human-like Perception: Learning Structural Causal Model in Heterogeneous Graph}

\shortauthors{Lin et~al.}

\title [mode = title]{Towards Human-like Perception: Learning Structural Causal Model in Heterogeneous Graph}

\author[1,3]{Tianqianjin Lin}
\ead{lintqj@zju.edu.cn}
\credit{Conceptualization, 
Formal analysis, 
Investigation, 
Methodology, 
Software, 
Validation, 
Visualization, 
Writing – original draft}

\author[3]{Kaisong Song}
\ead{kaisong.sks@alibaba-inc.com}
\credit{Resources, 
Supervision, 
Writing – review \& editing}

\author[1]{Zhuoren Jiang}
\cormark[1]
\ead{jiangzhuoren@zju.edu.cn}
\credit{Funding acquisition,
Methodology, 
Supervision, 
Visualization, 
Writing – review \& editing}

\author%
[3]
{Yangyang Kang}
\ead{yangyang.kangyy@alibaba-inc.com}
\credit{Project administration, 
Resources, 
Supervision}

\author%
[1]
{Weikang Yuan}
\ead{yuanwk@zju.edu.cn}
\credit{Data curation}

\author%
[3]
{Xurui Li}
\ead{xurui.lee@msn.com}
\credit{Data curation}

\author%
[3]
{Changlong Sun}
\ead{changlong.scl@taobao.com}
\credit{Project administration, 
Resources, 
Supervision}

\author%
[1]
{Cui Huang}
\ead{huangcui@zju.edu.cn}
\credit{Funding acquisition, Supervision}

\author%
[2]
{Xiaozhong Liu}
\cormark[1]
\ead{xliu14@wpi.edu}
\credit{Methodology, Project administration, Supervision, Writing - review \& editing}

\affiliation[1]{organization={Department of Information Resources Management, Zhejiang University}, 
    city={Hangzhou},
    postcode={310058}, 
    country={China}}

\affiliation[2]{organization={Computer Science Department, Worcester Polytechnic Institute},
    city={Worcester},
    postcode={01609-2280}, 
    state={Massachusetts},
    country={USA}}

\affiliation[3]{organization={Alibaba DAMO Academy},
    city={Hangzhou},
    postcode={311121}, 
    country={China}}

\cortext[cor1]{Corresponding author}



\begin{abstract}
Heterogeneous graph neural networks have become popular in various domains. 
However, their generalizability and interpretability are limited due to the discrepancy between their inherent inference flows and human reasoning logic or underlying causal relationships for the learning problem. 
This study introduces a novel solution, HG-SCM (Heterogeneous Graph as Structural Causal Model). It can mimic the human perception and decision process through two key steps: 
constructing intelligible variables based on semantics derived from the graph schema and automatically learning task-level causal relationships among these variables by incorporating advanced causal discovery techniques. 
We compared HG-SCM to seven state-of-the-art baseline models on three real-world datasets, under three distinct and ubiquitous out-of-distribution settings. 
HG-SCM achieved the highest average performance rank with minimal standard deviation, substantiating its effectiveness and superiority in terms of both predictive power and generalizability. 
Additionally, the visualization and analysis of the auto-learned causal diagrams for the three tasks aligned well with domain knowledge and human cognition, demonstrating prominent interpretability.  
HG-SCM's human-like nature and its enhanced generalizability and interpretability make it a promising solution for special scenarios where transparency and trustworthiness are paramount. 
\end{abstract}




\maketitle
\input{content/intro.tex}
\input{content/related.tex}
\input{content/prel.tex}

\input{content/method.tex}

\input{content/expr.tex}

\input{content/conclusion.tex}






\appendix
\input{content/appx.tex}


\printcredits

\bibliographystyle{cas-model2-names}

\bibliography{cas-refs}



\end{document}

%% file: math_cmds.tex
\usepackage{amsmath,amsfonts,bm}

\def\eqref#1{equation~\ref{#1}}

\def\1{\bm{1}}

\def\rvx{{\mathbf{x}}}
\def\rvy{{\mathbf{y}}}

\def\vh{{\bm{h}}}

\def\vv{{\bm{v}}}

\def\vx{{\bm{x}}}
\def\vy{{\bm{y}}}

\def\mA{{\bm{A}}}

\def\mH{{\bm{H}}}

\def\mX{{\bm{X}}}
\def\mY{{\bm{Y}}}

\DeclareMathAlphabet{\mathsfit}{\encodingdefault}{\sfdefault}{m}{sl}
\SetMathAlphabet{\mathsfit}{bold}{\encodingdefault}{\sfdefault}{bx}{n}



%% file: content/intro.tex
\section{Introduction}\label{sec:intro}

The surge and convergence of numerous readily available data sources have elicited exhilarating opportunities for investigation across various data-rich domains. Heterogeneous Graph Neural Network (HGNN), as a noteworthy approach in this realm, has garnered significant attention recently due to its effectiveness in modeling real-world complex systems with intricate relationships, \revise{such as academic networks, e-commerce networks, and social networks, in which different types of nodes and relations interplay following specific graph schema~\citep{openhgnn,dong2020heterogeneous,jiang2020task,ipm2023FactVerification,ipm2023sentimentanalysis,IPM2023MOOC,relation_aware_hgnn_rel_predict,ipm_reviewer2_image_text_graph}.}  

Despite the impressive results these models have achieved, three challenges have yet to be fully elucidated. 

\textbf{Generalizability:} 
The vast majority of studies on HGNN are conducted under the premise of identically and independently distributed (i.i.d) data splits, which can potentially lead to a disregard for the issue of generalizability. \emphasize{In practical applications, however, the i.i.d assumption is not always valid as the presence of distribution shifts in unseen data}, also known as out-of-distribution (o.o.d) problems~\citep{zhang2021deep,TKDE_Causal_Representations}. HGNNs that are susceptible to the presence of spurious correlations and connections in the training data may fail to generalize to unseen data~\citep{li2022ood,miao2022interpretable,ipm-Robust-gcn, robust_heterogeneous_embedding}. As~\citet{knyazev2019understanding} claimed, \emphasize{the popular neighborhood aggregation mechanisms used in current HGNNs are vulnerable to the presence of spurious edges and correlations} that mislead in how they attend to node neighbors, thereby resulting in inadequate generalizability of the models.

\textbf{Interpretability:} 
Building more accurate predictive models is not the only objective for graph models~\citep{ying2019gnnexplainer}. It is imperative for researchers to discover the patterns from the input graph that induce certain predictions~\citep{cranmer2020discovering}. While attention-based graph models~\citep{gat,hgt,simplehgn} are capable to assign weights to edges in the input graph, research has found that \emphasize{these estimated weights can not provide any reliable interpretation for the learning tasks}~\citep{yugraph,miao2022interpretable}. These methods mostly estimate the input-output relationships at the sample level from an associational perspective~\citep{wang2022reinforced}, which may overestimate the graph structure in a single input graph irrelevant to the outcome~\citep{moraffah2020causal}, \emphasize{rather than uncovering the plausible causation of the task itself}. 

\begin{figure*}[!htbp]
  \centering
  \includegraphics[width=1.0\textwidth]{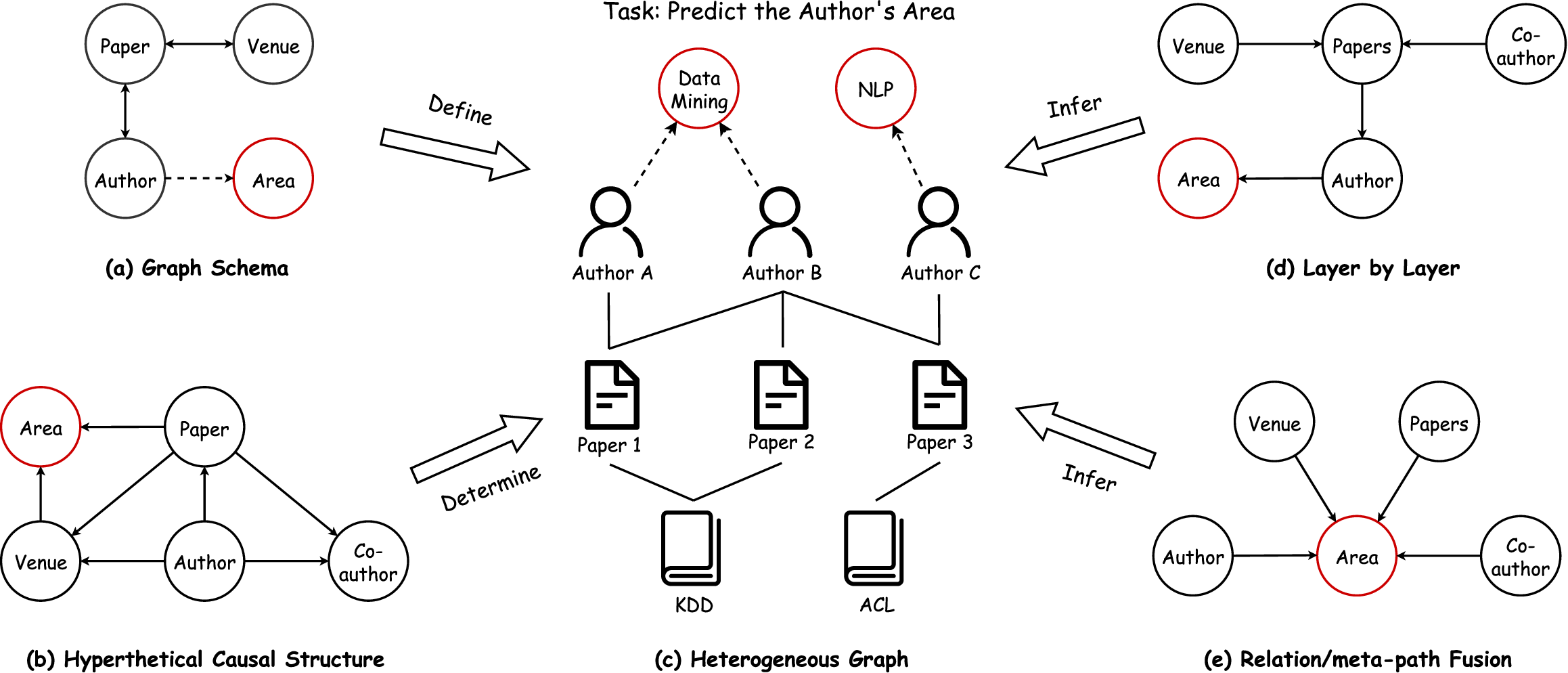}
  \caption{A toy academic heterogeneous graph is displayed in (c), where KDD and ACL are highly representative conferences in the fields of data mining and natural language processing, respectively. Its graph schema is shown in (a). For the task of predicting the author’s research area, a hypothetical causal structure is depicted in (b), where the research area of an author is only determined by two factors: the author’s papers and the venues of the author’s papers. The two mainstream paradigms of current HGNNs are illustrated in (d) and (e). It can be observed that \emphasize{the inherent associational inference flows of current HGNNs can not align well with human reasoning logic or underlying causal relationships. Therefore, they can face limitations in correct interpretation with respect to the learning tasks and are prone to get compromising generalizability due to inevitably introducing spurious correlations.}}
  \label{fig:intro_dag_mismatch}
\end{figure*}

\textbf{Learning level}. \revise{Perhaps due to the success of many models that employ sample-level dynamic adaptation~\citep{gat, transformerconv,ieee_reviwer2_path_selection}, existing HGNNs primarily focus on sample-adaptive learning by leveraging complex attention modules in the neighbor aggregation and relations/meta-paths fusion~\citep{rgcn, hgt, simplehgn, compgcn}. \emphasize{However, beyond the sample-level adaptation, for a given learning problem on a graph, there is always a relatively consistent principle for processing all samples.}} For instance, the importance of different relationships/meta-paths should be stable in a specific task~\citep{jiang2018cross}. As a result, for prediction problems in heterogeneous graphs, it can be critical to prioritize learning at the task level, rather than solely focusing on fine-grained sample-level adaptation. The empirical evidence from \citet{sehgnn} has indicated that \emphasize{the sample-wise neighbor attention has limited impacts on the performance of the models} in fact. Our supplementary experiment in Appendix~\ref{appx:learning-level} suggests that \emphasize{sample-wise semantic attention may also yield limited improvement in model performance}.

We provide a toy example in Figure~\ref{fig:intro_dag_mismatch} to explain these three challenges from the perspective of the design of the inference flow. 
\emphasize{The inference flow of HGNN models naturally introduces spurious correlations to reduce the generalizability of the model.} For instance, in the relation/meta-path fusion models (Figure~\ref{fig:intro_dag_mismatch}e), the co-author's information can always have a direct impact on the prediction. While information from co-author B (whose research area is data mining) is useful for predicting the research area of author A (data mining), for predicting author C's research area (NLP), co-author B's information would introduce noise. Thus, if the model attaches great importance to co-author information, the generalizability of the model will decrease. 
\emphasize{Current HGNN models carry limitations in interpretability because their inference flows are fixed.} For instance, in the layer-by-layer models (Figure~\ref{fig:intro_dag_mismatch}d), the venue's information can only indirectly impact the prediction of the research area through intermediate variables, i.e., paper and author. Additionally, the venue's information can become indistinguishable due to the blending with co-authors. As a result, it would be difficult for the model to provide a clear and explicit explanation: an author would be considered as a data mining scholar when this author publishes papers mainly at KDD. 
 
\emphasize{Under the constraints of these two mainstream paradigms, current efforts tend to focus on designing sophisticated aggregation or fusion within the inference flow, rather than reflecting on whether the paradigm of the inference flow needs to be changed.} 
Generally, ignoring achievable task-level causal relationships and strengthening the sample-level adaptive learning based on the inherent fixed associational inference logic may exacerbate the issues of interpretability and generalizability. 

\subsection{Research objectives and contributions}\label{sec:research_question}
The three challenges mentioned above pose significant obstacles to the trustworthiness and reliability of current HGNNs in real-world settings. To shed light on the reasons behind these challenges, we present an instructive toy example in Figure~\ref{fig:intro_dag_mismatch} and corresponding theoretical analysis that illustrate the deficiencies inherent in the pre-defined and fixed inference flows in existing HGNNs. 
  
As such, the focus of this work is to develop an HGNN architecture that aligns with human reasoning logic and causal mechanisms for the learning problem, achieving inherent and powerful interpretability and generalizability.  
Our motivation stems from a fundamental question: \emphasize{``What will humans do when facing a prediction task on a heterogeneous graph?''} Generally, when reasoning about a task defined in a system, humans will cognitively perceive meaningful variables involved in the system, select potentially causally-correlated factors from them, and estimate the causal effect of these factors on the task for task reasoning. From this viewpoint, we propose a novel solution for the node property prediction task, named HG-SCM (Heterogeneous Graph as Structural Causal Model). HG-SCM is designed to mimic the human perception and decision process and automatically learn task-level causal relationships by incorporating emerging causal techniques.  
  
Specifically, the following research questions will be addressed in this work: 
\begin{enumerate} 
    \item How to define, construct and represent human-understandable variables in the heterogeneous graph?\label{rq1}
    \item How to automatically discover the task-level causal relationships among the variables and identify the direct causes for the target variable? \label{rq2}
    \item How to make the prediction based on the learned task-level causal relationships? \label{rq3}
    \item How good is the task performance of the proposed solution? \label{rq4}
    \item How effective is the proposed solution for achieving better generalizability and interpretability?  \label{rq5}
\end{enumerate}  

Technically, HG-SCM constructs meaningful variables based on schema-level semantics in the heterogeneous graph. Specifically, the target node, the label of the target node, and neighbor sets of the target node based on different relations/meta-paths are considered available variables. HG-SCM embeds these variables via mutually-independent encoders without fitting the spurious correlations among them. By incorporating emerging causal structural learning techniques into the understanding of a heterogeneous graph task, HG-SCM further learns task-level causal relationships among these variables and makes predictions based only on variables that are likely to be causally correlated to the target variable. The learned causal structures can further provide clear interpretation with respect to the learning tasks. Own to such human-like reasoning logic, HG-SCM naturally equips enhanced interpretability and generalizability. Extensive experiments and in-depth analysis under both the i.i.d setting and the o.o.d setting effectively validate such hypotheses.

In summary, the contribution of this paper can be threefold: 
\begin{enumerate}
    \item We propose a novel heterogeneous graph algorithm HG-SCM. Unlike prior works, its inference flow aligns with human reasoning logic or underlying causal diagrams. To the best of our knowledge, this is a pioneer investigation to introduce the structural causal model into heterogeneous graph learning. 
    
    \item HG-SCM can consistently and significantly outperform various SOTA baselines in extensive experiments under the i.i.d setting and various types of the o.o.d settings, which verify the optimal efficiency and promising generalizability of HG-SCM. 

    \item HG-SCM can provide in-depth interpretations in accordance with the learning tasks by automatically discovering causal relationships among meaningful semantics hidden in a heterogeneous graph along with the graph schema. 
\end{enumerate}

%% file: content/related.tex
\section{Literature Review}\label{sec:related_work}

\subsection{Heterogeneous Graph Neural Networks}
Recent studies have attempted to explore algorithms for modeling heterogeneous graphs since many real-world problems can hardly be represented by homogeneous graphs. Among these efforts, message-passing-based heterogeneous graph neural networks (HGNNs), e.g., RGCN~\citep{rgcn}, HAN~\citep{han}, GTN~\citep{gtn}, CompGCN~\citep{compgcn}, HGT~\citep{hgt}, PGRA~\citep{PGRA}, SimpleHGN~\citep{simplehgn}, SeHGNN~\citep{sehgnn}, RHGCN~\citep{relation_aware_hgnn_rel_predict}, HetReGAT~\citep{HetReGAT-FC}, and R-HGNN~\citep{R-HGNN},  have emerged as the dominant approach because these methods can leverage complex encoders along with deep neural networks and enable the natural modeling of both spatial proximity and node attributes~\citep{dong2020heterogeneous}. 
HGNNs generally learn node representations from neighbors in two approaches: \textit{meta-path-based fusion} and \textit{layer-by-layer aggregation}. The meta-path-based methods~\citep{han, magnn, sehgnn} make in-depth use of the heterogeneous graph semantics. For example, two authors connected by a meta-path ``Author-Publication-Author'' suggest they have an academic partnership. Meanwhile, layer-by-layer methods~\citep{rgcn, compgcn, hgt, simplehgn} learn a node's representation by simultaneously aggregating all the directly connected neighbors belonging to all edge types, and they update the node representation recursively. \revise{The rapid development of heterogeneous graph neural networks has also led to a series of transformations and applications in specific fields, such as fact verification~\citep{ipm2023FactVerification}, recommendation~\citep{lichenliang_IPM_hgn, IPM_Bert_HGNN_citation_rec, ipm2023knowledgeGNN, hgnn_link_meta_learning_ipm}, sentiment analysis~\citep{ipm2023sentimentanalysis, esa_reviewer2_hgnn_sentiment}, video question answering~\citep{IPM_VQA_HGNN}, stock prediction~\citep{FinHGNN} and knowledge graph learning~\citep{IPM_KG_representation}.} \emphasize{Unfortunately, these approaches have limitations in terms of model generalizability and interpretability due to the deficiency of their inference flows}~\citep{wang2022reinforced,moraffah2020causal,yugraph,knyazev2019understanding}.

\SecondRevise{To address this issue, research on the generalizability and interpretability of graph neural networks has undergone notable advancements in recent years. 
In the realm of generalizability, prior efforts have not only expanded data augmentation~\citep{aaai_data_aug,cvpr_data_aug,nips_graphcl} and training strategies~\citep{ipm-Robust-gcn,www_confidence_cheat,tkde_adversarial_train} intrinsic to the broader field of machine learning but have also formulated novel model structures and prediction pipelines based on disentanglement techniques~\citep{icml_disentangled_gcn,aaai_independence} and causality tools~\citep{icml_gem,tnnls_debiased}. Despite these advancements, the inherent black-box nature of these methods remains a fundamental concern. This opacity hinders human comprehension and collaboration, thereby constraining the applicability of the models, particularly in high-stakes domains~\citep{nature_machine_intelligence_stable_learning}. 
On the interpretability front, prevailing methods primarily adopt post-hoc approaches, evaluating the significance of nodes or edges in the input graph for a trained graph model. \citet{tpami_explainability_survey} categorizes these methods into two branches: model-level and instance-level. While XGNN~\citep{xgnn} stands as the sole model-level method, instance-level techniques encompass gradients-based~\citep{cvpr_grad_cam, icml_2019_workshop_grad_based}, perturbation-based~\citep{ying2019gnnexplainer, nips_PGE}, decomposition~\citep{tpami_gnn_lrp}, and surrogate methods~\citep{GraphLime}. Additionally, a few approaches enhance built-in interpretability through learnable prototypes~\citep{aaai_ProtGNN,tpami_prototype_gnn} and concept distillation~\citep{xai_concept_distill}. However, all of these approaches only aid in understanding the dependency path of predictions, falling short of aligning with human cognitive processes. Consequently, trust issues persist in real-world applications. 
Moreover, it is noteworthy that the preponderance of extant generalizability and interpretability methods is tailored for homogeneous graphs. Therefore, they struggle to effectively handle or leverage the intricate semantics inherent in heterogeneous graphs. This limitation further underscores the need for continued advancements.}

\subsection{Causal Structural Learning}
Causal understanding is a fundamental problem of science~\citep{pearl2009causality,bareinboim2022pearl, nature2020causal} and is crucial for reasoning about the physical world~\citep{dag_review,AI_review_survey_dag}. In order to discover causal relations and acquire causal understanding, randomized experiments (REs) with interventions and manipulations can be carried out~\citep{dag_review, 2012FieldExperiments}. However, in real applications, REs tend to be costly or even impractical due to ethical concerns, etc~\citep{dag_review,active_nips20}. Therefore, researchers often discover causal structures from non-experimental and observational data. One can achieve this goal by learning a Bayesian networks (BNs), which encodes the conditional independencies between variables using directed acyclic graphs (DAGs), but learning such networks from data can be computationally intractable due to the combinatorial explosion in the search space~\citep{nature2020causal}. Recent work~\citep{notears, dag_gnn, 2020GradientDAG, pmlr_noparam_dag, NPS2021_BCD, 2020Causal_huawei_rl, nips20_dag_nofears, dag_sampling, jmlr_Adversarial_dag, AI_Markov_dag, ML_mix_dag, tnnls_Low_rank_DAG, nonlinear_confounder_dag} has made it possible to approximate this problem as a continuous optimization task~\citep{nature2020causal} by minimizing an innovative smooth function that quantifies the ``DAG-ness'' in both linear and non-linear cases~\citep{castle}. These techniques provide us with an unprecedented opportunity to develop neural models that could achieve both accuracy and interpretability simultaneously~\citep{nature2020causal,castle,CausPref}. For example, \citet{castle} regard such techniques as a regularization method in regression models, \citet{CausPref} incorporate DAG into the recommendation domain, and \citet{ipm2023causal} utilize  causality learning in the model in click-through rate prediction. \emphasize{However, to the best of our knowledge, no prior works have integrated this technique into the framework design of heterogeneous graph algorithms.}

%% file: content/prel.tex
\section{Definitions and Notations}

\begin{definition}\label{def:heteroG}
\textbf{Heterogeneous Graph:} A heterogeneous graph is defined as a directed graph $\mathcal{G} = (\mathcal{V}, \mathcal{E})$ with a node type mapping function $\phi: \mathcal{V} \rightarrow \mathcal{T}$ and an edge type mapping function $\varphi: \mathcal{E} \rightarrow \mathcal{R}$, where each node $n\in \mathcal{V}$ belongs to a particular node type $t \in \mathcal{T}$ and each edge $e\in \mathcal{E} $ belongs to a particular edge type $r \in \mathcal{R}$~\citep{sun2011pathsim}. Furthermore, in this work, we consider settings where nodes are associated with features. As a result, for each node $n_i\in \mathcal{V}$, a feature vector ${\vx}_{i}\in \mathbb{R}^D$ is assigned. $D$ is the supposed feature dimension.

\end{definition}

\begin{definition}\label{def:node_task}
\textbf{Node Property Prediction:} Generally, a node property prediction task is defined on a specific node type $t$ in a graph $\mathcal{G}$ and is to predict properties of a single node belonging to the node type $t$. In this work, we take the node classification task as an example. The task is to estimate a function $\psi_{\mathcal{G}}: \mathcal{V}_{t} \rightarrow \mathcal{Y}$ which can map each node $n_i\in \mathcal{V}_{t}$ to a categorical vector $\vy_i$ in the label space $\mathcal{Y}\in \mathbb{R}^C$ based on a given labeled node set $\mathcal{V}_t^* \subseteq \mathcal{V}_t$, where $\mathcal{V}_{t}$ denotes the node set $\set{n_i | n_i \in \mathcal{G} \wedge \phi(n_i)=t}$ and 
$C$ is the number of classes.
\end{definition}

\begin{definition}\label{def:metapath}
\textbf{Meta-Path:} A meta-path~\citep{dong2017metapath2vec} is a path in the form of $t_1 \xrightarrow[]{r_1} t_2 \xrightarrow[]{r_2} \cdots \xrightarrow[]{r_l}  t_{l+1}$, which defines a composite $l$-hop relation $p = r_1 \circ r_2\circ \cdots \circ r_{l}$ between the node type $t_1$ and $t_{l+1}$. 
\end{definition}

\begin{definition}\label{def:ego_graph}
\textbf{Ego-graph:} Ego-graphs are local graphs with the focal node (known as the ego), while all other nodes connected to the ego are called alters~\citep{ego_network_2009,ego_network_2007}. An ego-graph can be defined as a $k$-hop ego-graph when the maximum distance between the alters and the ego is $k$ and the alters contain all $k$-hop neighbors of the ego. 
\end{definition}

\begin{definition}\label{def:scm}
\textbf{Structural Causal Model:} A Structural Causal Model (SCM) can describe the causal mechanisms of a system. \revise{Specifically, assuming no unobserved variable exists, a SCM of $k$ variables $\set{\vv_i\mid 1 \le i \le k }$ consists of a collection $\set{f_i \mid 1\le i \le k}$ of structural assignments $\vv_i\coloneqq f_i(\mathrm{pa}_i)$ where $\mathrm{pa}_i$ is the set of direct causes of $\vv_i$. ``$\coloneqq$'' represents the assignment operation and it means the value of $\vv_i$ should be determined by its direct causes $\mathrm{pa}_i$ through the function $f_i$.} Usually, assignments are assumed acyclic and thus these assignments can be represented by a Directed Acyclic Graph (DAG)~\citep{dag1975} with edges pointing from causes to effects~\citep{peters2017elements, deep_scm_nips}.
\end{definition}

The notations are summarized in Table \ref{tab:notations} and vectors/matrices are indexed starting from zero in this paper.

\input{tables/notations.tex}

%% file: tables/notations.tex
\begin{table}[!h]
  \centering
  \caption{
    Notations and Explanations.
  }
  \label{tab:notations}%
    \begin{threeparttable}
  \renewcommand{\arraystretch}{1.}
    \begin{tabular}{cccc}
    \toprule
    Notation & Explanation & Notation & Explanation\\
    \midrule
    $r$         & an edge type & $\vx$       & a node feature vector\\
    $t$         & a node type& $\vy$       & a label vector of a node\\
    $p$         & a meta-path& $\vh$       & a hidden representation vector\\
    $q$         & the number of used relations/meta-paths & $\mX$       & a matrix of node feature vectors\\
    $C$         & the number of classes & $\mY$       & a matrix of node label vectors\\
    $B$         & the number of samples in a batch & $\mH$       & a matrix of hidden representation vectors\\
    $N$         & a neighbor set & $\mA$       & an adjacency matrix of a DAG \\
    
    \bottomrule
    \end{tabular}%

  \end{threeparttable}
\end{table}%

%% file: content/method.tex
\begin{figure}[ht]
  \centering
  \includegraphics[width=1.0\textwidth]{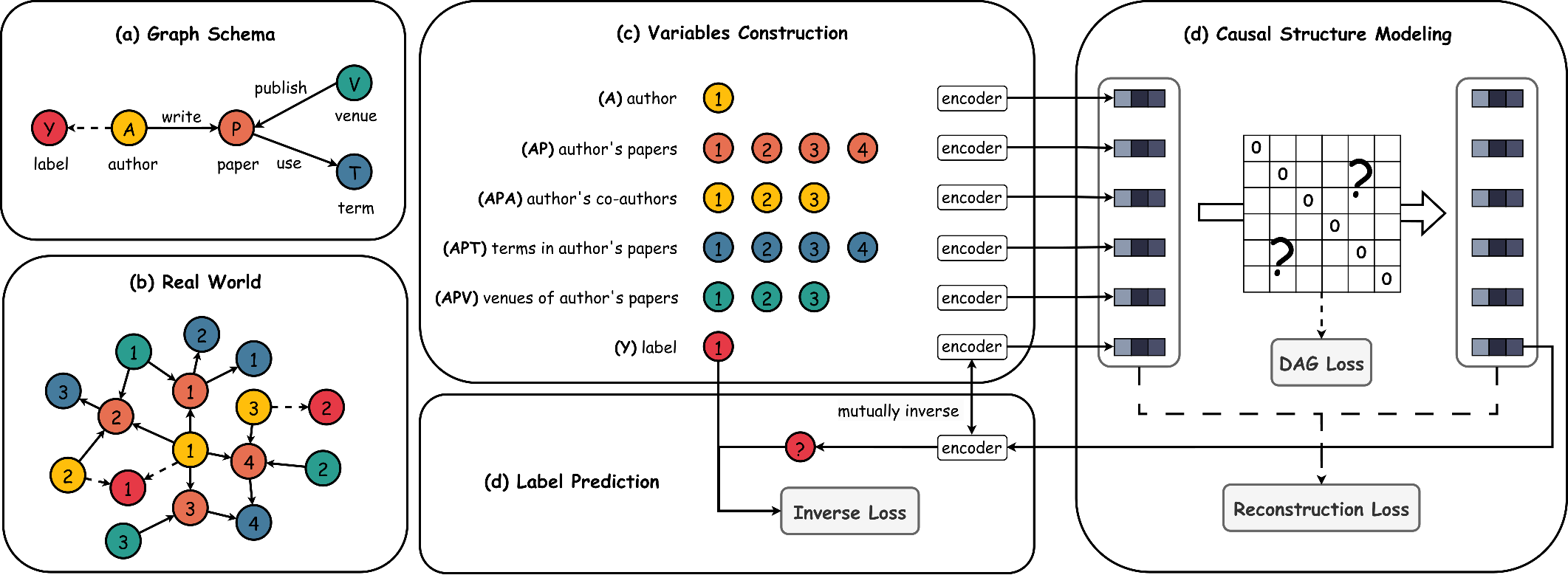}
  \caption{Overall architecture of HG-SCM.}
  \label{fig:overall}
\end{figure}
\section{Methodology}\label{sec:method}
In this section, we illustrate the proposed model, which can answer research questions \ref{rq1}, \ref{rq2} and \ref{rq3} in Section~\ref{sec:var_construct}, Section~\ref{sec:causal_structure_modeling} and Section~\ref{sec:label_prediction}, respectively. The overall HG-SCM architecture is depicted in Figure~\ref{fig:overall} and the general forward process of HG-SCM is described in Algorithm~\ref{overall_algorithm_training}. In addition, the optimization objectives are explained in Section~\ref{sec:optim-loss} and we provide the analysis of the computational complexity of the HG-SCM in Section~\ref{sec:complexity}.

\subsection{Variables Construction}\label{sec:var_construct}
As aforementioned, when given a task defined by a system, humans naturally tend to comprehend information and identify variables involved in the system. \revise{In the case of a node classification task, $\psi_{\mathcal{G}}: \mathcal{V}_t \rightarrow \mathcal{Y}$, which maps each node $n_i\in \mathcal{V}_{t}$ to a categorical vector $\vy_i$ in the label space $\mathcal{Y}\in \mathbb{R}^C$, two variables are naturally included, i.e., the node feature $\rvx$ of a target node $n$ and the node label $\rvy$ of the target node. }
In addition, we can construct meaningful variables based on various semantics that can be derived from the graph schema, i.e., relations and meta-paths. For example, as shown in Figure~\ref{fig:overall}, given the graph schema and the target node type ``author'', we can obtain semantics such as ``author's papers'' and ``co-authors'' through the relation ``Author-Paper'' and the meta-path ``Author-Paper-Author'', respectively. \revise{Based on the real graph $\mathcal{G}$, the variables based on these semantics can be embodied by the set of neighboring nodes. For instance, in Figure~\ref{fig:overall}, the value of the variable based on semantic ``author's papers'' of the author ``1'' can be represented by the set of papers ``1'', ``2'', ``3'' and ``4''.} \SecondRevise{\emphasize{To maintain the context integrity of the target node, we consider all meta-paths starting from the target node type within a specified length limit. Thereby, in this work, the variables inherent to the task $\psi_{\mathcal{G}}$ consist of three parts: the node itself, the node's label, and the neighbor sets corresponding to all available relations/meta-paths.}}  
 
\SecondRevise{Due to the heterogeneous nature, these variables typically have different forms and may exist in different feature spaces. For instance, nodes and labels are often vectors of different dimensional sizes, and the neighbor sets are collections of vectors. To address this, we need to embed all these variables into the same representation space.}  
  
For a node $n_i$, its feature $\vx_i$ and its label $\vy_i$ are encoded by two linear transformations as follows:

\begin{align}
    \vh^i_x  &= \mathrm{Linear}_x(\vx_i)\label{eq:ego_enc}\\
    \vh^i_y  &= \mathrm{Linear}_y(\vy_i)\label{eq:label_enc}
\end{align}

\revise{Then, for the $j$-th relation/meta-path based variable of the node $n_i$, we encode the corresponding neighbor set to a fixed-size vector representation $\vh^i_j$ via a function $\mathrm{NeighborSetEncoder}_j$}, i.e.,

\begin{equation}\begin{aligned} 
    \vh^i_j  &= \mathrm{NeighborSetEncoder}_j(\set{\vx_{k} | n_{k}\in N^i_j})= \frac{1}{|N^i_j|}\sum\vx_k, \label{eq:setenc} 
\end{aligned}\end{equation}
where $N^i_j$ is the neighbor set of the target node $n_i$ based on the $j$-th relation/meta-path given the graph $\mathcal{G}$ and $|N^i_j|$ is the number of nodes in $N^i_j$. The set $\set{\vx_{k} | n_{k}\in N^i_j}$ represents the collection of node feature vectors of all nodes contained in $N^i_j$. Practically, the $\mathrm{NeighborSetEncoder}$ can be implemented by any module that is capable of handling an unordered set of vectors, e.g., various pooling operators~\citep{graphsage}, Transformer~\citep{vaswani2017attention} and Set Transformer~\citep{settransformer}. For simplicity and motivated by~\citet{sehgnn}, we use a simple average pooling here.

Note that these encoders, i.e., $\mathrm{Linear}_x$, $\mathrm{Linear}_y$ and $\set{\mathrm{NeighborSetEncoder}_j}$, need to be mutually independent to avoid fitting the spurious correlations among variables. For example, the widely used encoding methods in previous works~\citep{gat,hgt,simplehgn}, which compute attention weights between neighboring nodes and target nodes, are not satisfactory since they introduce correlation between $\vx_i$ and $\set{\vx_{k} | n_{k}\in N^i_j}$ into the representation.

Assuming there are $q$ relations/meta-paths, we can obtain a set of $q+2$ variables, i.e., $\set{\vh^i_x, \vh^i_y}\cup \set{\vh^i_j|1 \le j \le q}$. For brevity, we ignore the superscript $i$ and use $\vh_0$ and $\vh_{q+1}$ to refer to $\vh_x$ and $\vh_y$, respectively.

\subsection{Causal Structure Modeling}\label{sec:causal_structure_modeling}
Based on the assumption that a \revise{causal directed acyclic graph (DAG)} can exist among the above-constructed variables, in this section, we illustrate how to leverage emerging causal discovery techniques to learn a structural causal model from the constructed variables. \revise{According to Definition~\ref{def:scm}, we need to learn a collection of $\set{f_k \mid 0\le k \le q+1}$ of structural assignments function $\vh_k\coloneqq f_k(\mathrm{pa}_k)$ where $\mathrm{pa}_k$ is the set of direct causes of $\vh_k$, and its value should be determined by the causes.}  

We initialize a trainable matrix $\mA \in \mathbb{R}^{(q+2) \times (q+2)}$ to represent the causal DAG, where $\mA_{i,j}$ represent the probability that $\vh_i$ is one of the directed causes of $\vh_j$. 
Note that the diagonal of the matrix $\mA$ is constrained to be zero since a variable cannot be its own cause, i.e., $\mA_{i,i}=0$.  

Based on the matrix $\mA$, we define the structural assignment $f_k$ of a variable $\vh_k$ as:

\begin{equation}\begin{aligned}
    f_k(\mathrm{pa}_k) = \mathrm{VariableDecoder}_k\left(\sum_{i=0}^{q+1}\mA_{i,k} \cdot \mathrm{EffectEncoder}_{ik}(\vh_i)\right), \label{eq:assign} 
\end{aligned}\end{equation}

\revise{where $\mathrm{EffectEncoder}_{ik}$ is a function to computes the hidden state of the causal effect of $\vh_i$ on $\vh_k$ and $\mathrm{VariableDecoder}_{k}$ is a function to reconstruct the variable $\vh_k$ based on the received causal effects. 
We use multi-layer perceptrons (MLPs) to model and learn $\mathrm{EffectEncoder}_{ik}$ and $\mathrm{VariableDecoder}_k$ without any assumption on the underlying functions of them, thanks to the universal approximation theorem~\citep{mlp}. In detail, we implement Equation~\ref{eq:assign} as following: } 

\begin{equation}\begin{aligned}
    \hat{\vh}_k = \mathrm{MLP}_k\left(\sum_{i=0}^{q+1}\mA_{i,k} \cdot \mathrm{Linear}_{ik}\left(\mathrm{MLP}_{i}\left(\vh_i\right)\right)\right), \label{eq:assign_detail}
\end{aligned}\end{equation}
where $\mathrm{MLP}_k$ is a three-layer MLP, $\mathrm{MLP}_{i}$ is a shared two-layer MLP when variable $\vh_i$ serves as a cause and imposes influence on other variables, and $\mathrm{Linear}_{ik}$ between each pair of variables, i.e., $\vh_i$ and $\vh_k$, are mutually independent because the function of the causal relationships between different variables can be different. Generally, Equation~\ref{eq:assign} can also be implemented based on many other powerful modules, as long as these modules have the ability to process the unordered set of weighted vectors. Our implementation is one of the simple modules, and we found that it can achieve good performance.  

\subsection{Label Prediction}\label{sec:label_prediction} 
\revise{Once the structural assignments collection, i.e., $\set{f_k \mid 0\le k \le q+1}$ has been learned, we can have the reconstructed label representation $\hat{\vh}_{y}$, i.e., $\hat{\vh}_{q+1}$, through $f_{q+1}\in\set{f_k \mid 0\le k \le q+1}$, i.e., Equation~\ref{eq:assign} and Equation~\ref{eq:assign_detail} with $k=q+1$. Thereby, we can achieve the label prediction under the causal constraints as }

\revise{
\begin{equation}\begin{aligned}
    \hat{\vy} = \mathrm{Linear}_{y}^{-1}(\hat{\vh}_{y}), \label{eq:label_inverse}
\end{aligned}\end{equation}
}

where $\mathrm{Linear}_{y}^{-1}$ is the inverse function of $\mathrm{Linear}_{y}$ mentioned in Equation~\ref{eq:label_enc}. However, $\mathrm{Linear}_{y}^{-1}$ may not be solvable mathematically. Therefore, we use a two-layer MLP with a shortcut to approximate this inverse linear transformation as follows:
\revise{
\begin{equation}\begin{aligned}
    \hat{\vy} = \mathrm{Linear}^2_{inv}\left(\hat{\vh}_{y} + \sigma\left(\mathrm{Linear}_{inv}^1(\hat{\vh}_{y})\right)\right),. \label{eq:label_predictor}
\end{aligned}\end{equation}}
\revise{where $\mathrm{Linear}^2_{inv}$ and $\mathrm{Linear}_{inv}^1$ are the two linear transformation layers and $\sigma$ is the activation function, such as $\mathrm{Sigmoid}(x)=\frac{1}{1+e^{-x}}$, $\mathrm{ReLU}(x)=\mathrm{max}(0,x)$.}

\subsection{Optimization Objectives}\label{sec:optim-loss}
\emphasize{To ensure the structural assignment effectiveness}, the least-squares loss is applied on the reconstructed variables, i.e.,
\begin{align}
    \mathcal{L}_{rec} = \frac{1}{B} \sum_{i=1}^{B} \frac{1}{q+2}\sum_{k=0}^{q+1} ||\vh^i_k - \hat{\vh^i_k}||_F^2, \label{eq:rec}
\end{align}
where $B$ is the number of training samples and $||\cdot||_F$ is the Frobenius norm. \revise{$\vh^i_k$ and $\hat{\vh^i_k}$ are the raw value of the variable $i$ and the reconstructed value of the variable $i$, respectively.} Furthermore, \emphasize{to satisfy the directed acyclic constraint of the matrix $\mA$}, motivated by ~\citet{notears}, a smooth optimizable objective can be minimized as follows:

\begin{align}
    \mathcal{L}_{dag} &= \frac{\rho}{2} |\mathcal{L}_{acy}|^2 + \alpha\mathcal{L}_{acy},\label{eq:dag}
\end{align}
where $\rho$ and $\alpha$ can be hyper-parameters (while we set them to 1 in this work for the sake of simplicity), and $\mathcal{L}_{acy}$ is calculated by:
\begin{align}
    \mathcal{L}_{acy} &= \left(\mathrm{Tr}(e^{\mA\odot \mA}) - q - 2\right)^{2},\\
    \mathrm{Tr}(e^{\mA\odot \mA})&=\sum_{i=0}^{q+1}\sum _{k=0}^{\infty }{\frac {1}{k!}}(\mA\odot \mA)^{k}_{i,i},\label{eq:loss_trace_dag}
\end{align}

\revise{where $\mathrm{Tr(\cdot)}$ is the trace of a square matrix, which is defined to be the sum of elements on the main diagonal (from the upper left to the lower right) of the square matrix. $e^{\mA\odot \mA}$ is the matrix exponential of $\mA\odot \mA$. $(\mA\odot \mA)^{k}_{i,j}$ denotes the probability that $\vh_i$ can influence $\vh_j$ through $k$ steps and thus the term $(\mA\odot \mA)^{k}_{i, i}$ represents the probability of the existence of a cycle of length $k$ starting from node $i$ and returning to node $i$, which indicates the probability that variable $\vh_i$ is its own cause. By allowing ``$k$'' to take values from 0 to infinity, we can ensure that the result is equal to $q+2$ only if there are no cycles with any length in the causal DAG $\mA$. This implies that there are no causal relationships between variables that can create loops. In other words, there is no variable that is its own cause or two variables are mutually causes and effects of each other. }

In addition, a cross-entropy loss should be minimized to learn the inverse function of $\mathrm{Linear}_y$ in the Equation~\ref{eq:label_predictor}, i.e., 
\begin{align}
    \mathcal{L}_{inv} = -\frac{1}{B} \sum_{i=1}^{B} \sum_{j=0}^{C-1} \vy^i_j \cdot \mathrm{log}(\hat{\vy}^i_j), \label{eq:inv_ce}
\end{align}
\revise{where $C$ is the number of classes. $\vy^i_j$ and $\hat{\vy}^i_j$ are the probability of the class $j$ in the ground truth and the prediction, respectively.} 
Finally, these objectives can be jointly optimized by:
\begin{align}
    \mathcal{L}_{joint} = \underbrace{\vphantom{\beta\mathcal{L}_{rec} + \gamma \mathcal{L}_{dag}} \mathcal{L}_{inv}}_{\textup{Task}} + \underbrace{\beta\mathcal{L}_{rec} + \gamma \mathcal{L}_{dag}}_{\textup{Causal Structure}}. \label{eq:final_loss}
\end{align}
$\beta$ and $\gamma$ can adjust the weight ratios of these three objectives. Generally, the larger values of $\beta$ and $\gamma$ indicate that we have more confidence in the existence of an underlying causal DAG among the constructed variables.

\input{content/train.tex}
\subsection{Computational Complexity}\label{sec:complexity}
In the training stage, the computational cost mainly comes from the Equations~\ref{eq:ego_enc}, \ref{eq:label_enc},~\ref{eq:assign_detail} and~\ref{eq:label_predictor}. Suppose that the hidden dimension is $D$, the complexity for HG-SCM is $O(B\times D\times (2\times C) + (q^2 + 8q +15)\times B\times D^2 )$. In the evaluation stage, we only need to reconstruct the label variable, i.e., $k$ is set to $q+1$ in Equation~\ref{eq:assign_detail} and the loop defined from line~\ref{algorithm:line:loop_start} to line~\ref{algorithm:line:end} in Algorithm~\ref{overall_algorithm_training} is no longer needed. Therefore, the complexity for HG-SCM can be $O(B\times D\times (2\times C) + (7q +15)\times B\times D^2 )$. \emphasize{Compared to other models, HG-SCM has comparable computational complexity during the training stage.} For example, the computational complexities of SimpleHGN and SeHGNN are around $O(B\times E\times D^2)$ and $O(B\times (q^2 + 2q +1) \times D^2)$~\citep{sehgnn}, respectively. $E$ is the number of processed neighbors during the multi-layer aggregation and it generally exceeds $q^2$. However, \emphasize{HG-SCM can be faster during the evaluation stage due to the linear computational complexity regarding the number of relations/meta-paths.}

%% file: content/train.tex
\begin{algorithm}[ht]

	\renewcommand{\algorithmicrequire}{\textbf{Input:}}
	\renewcommand{\algorithmicensure}{\textbf{Output:}}
	\caption{The training forward process of HG-SCM.}
	\label{overall_algorithm_training}
	
	\begin{algorithmic}[1]
		\Require 
            Graph $\mathcal{G}=(\mathcal{V},\mathcal{E})$; 
            Features of all nodes $\mX \in \mathbb{R}^{|\mathcal{V}|\times D}$; 
		Target node type $t$; 
            $B$ training nodes $\mathcal{V}_t^*$ and their labels $\mY \in \mathbb{R}^{B\times C}$; 
            $q$ valid relations/meta-paths.
	    \Ensure 
            Task Prediction $\hat{\mY}\in \mathbb{R}^{B\times C}$.
	    \Statex{\color{OliveGreen}{// define model parameters}}
	    \State{\revise{Ego transformation $\mathrm{Linear}_x$; 
                Label transformation $\mathrm{Linear}_y$; 
                Causal DAG matrix $\mA$; 
                $q$ neighbor set encoder $\set{\mathrm{NeighborSetEncoder}_{i}|1\le i \le q}$ for each relation/meta-path;
                $q+2$ variable-wise causal effect encoders $\set{\mathrm{EffectEncoder}_{i}|0\le i \le q+1}$; 
                $(q+2)\times (q+1)$ pair-wise causal effect transformation $\set{\mathrm{Linear}_{ik}|0\le i,k \le q+1 \wedge i\neq k}$; 
                $q+2$ variable decoders $\set{\mathrm{VariableDecoder}_{i}|0\le i \le q+1}$;
                Inverse label encoder $\mathrm{Linear}_{inv}$.}
              
            }
	   \Statex{\color{OliveGreen}{// define function}}
	    \Function{StructuralAssignment}{
                $k$
        }
            \Statex{\color{OliveGreen}{\quad \;\,\,// reconstruct a variable based on a DAG}}
	   
    	\State{\revise{$\hat{\mH}_k\in \mathbb{R}^{B\times D} \gets$ Equation~\ref{eq:assign_detail}, given $k$, $\mA$, $\set{\mathrm{EffectEncoder}_{i}|i\neq k}$, $\set{\mathrm{Linear}_{ik}|i\neq k}$ and $\mathrm{VariableDecoder}_{k}$}}
            
            \State \Return $\hat{\mH}_k$
        \EndFunction

        \Statex{\color{OliveGreen}{// main process}}
        \State{$\mH_0\in{\mathbb{R}^{B\times D}} \gets$ Equation~\ref{eq:ego_enc}, given $\mathcal{V}^*$ and $\mX$}
        \State{$\mH_{q+1}\in \mathbb{R}^{B\times D} \gets$ Equation~\ref{eq:label_enc}}, given $\mY$
        \State{$\set{\mH_{k}\in \mathbb{R}^{B\times D}| 1\leq k\leq q} \gets$ 
        Equation~\ref{eq:setenc}}, given $\mathcal{G}$ and $\mX$
        \Statex{\color{OliveGreen}{// This loop is implemented by parallel matrix multiplication}}
        \For{$0 \leq k \leq q+1 $}\label{algorithm:line:loop_start}
            \State{$\hat{\mH}_{k}\in \mathbb{R}^{B\times D} \gets$\Call{StructuralAssignment}{$k$}}
        \EndFor\label{algorithm:line:end}
        \State{$\hat{\mY}\in \mathbb{R}^{B\times C} \gets$ 
        Equation~\ref{eq:label_inverse}}, given $\hat{\mH}_{q+1}$ and $\mathrm{Linear}_{inv}$
        \Statex{\color{OliveGreen}{// optimization objectives}}
        \State{$\mathcal{L}_{rec}\gets$ Equation~\ref{eq:rec}}, given $\set{(\mH_k, \hat{\mH}_k)| 0\leq k\leq q+1}$

        \State{$\mathcal{L}_{dag}\gets$ Equation~\ref{eq:dag}}, given $\mA$

        \State{$\mathcal{L}_{inv}\gets$ Equation~\ref{eq:inv_ce}}, given $\mY$ and $\hat{\mY}$

	\end{algorithmic}
\end{algorithm}

%% file: content/expr.tex
\section{Experiments}
To answer research questions~\ref{rq4} and \ref{rq5}, in this section, we conduct extensive experiments to validate HG-SCM task performance as well as promising generalizability and interoperability. In Section~\ref{sec:setup}, we will describe the experimental setup, including datasets, baselines and reproducibility. In Section~\ref{sec:results}, we will report the experimental results. Based on comprehensive experiment outcomes, HG-SCM outperformed a series of strong baselines when applied to the independent and identical distribution (i.i.d) setting and also showed its superiority and stability under multiple out-of-distribution (o.o.d) settings. In-depth analyses of the learned DAGs in HG-SCM further demonstrated its potential in model interpretation with respect to the learning tasks. 
\input{tables/datasets.tex}

\subsection{Experimental Setup}\label{sec:setup}
\subsubsection{Dataset}\label{sec:dataset} 
\begin{figure}[ht]
    \centering
    \begin{subfigure}[b]{0.3\textwidth}
        \centering
        \includegraphics[width=1.0\textwidth]{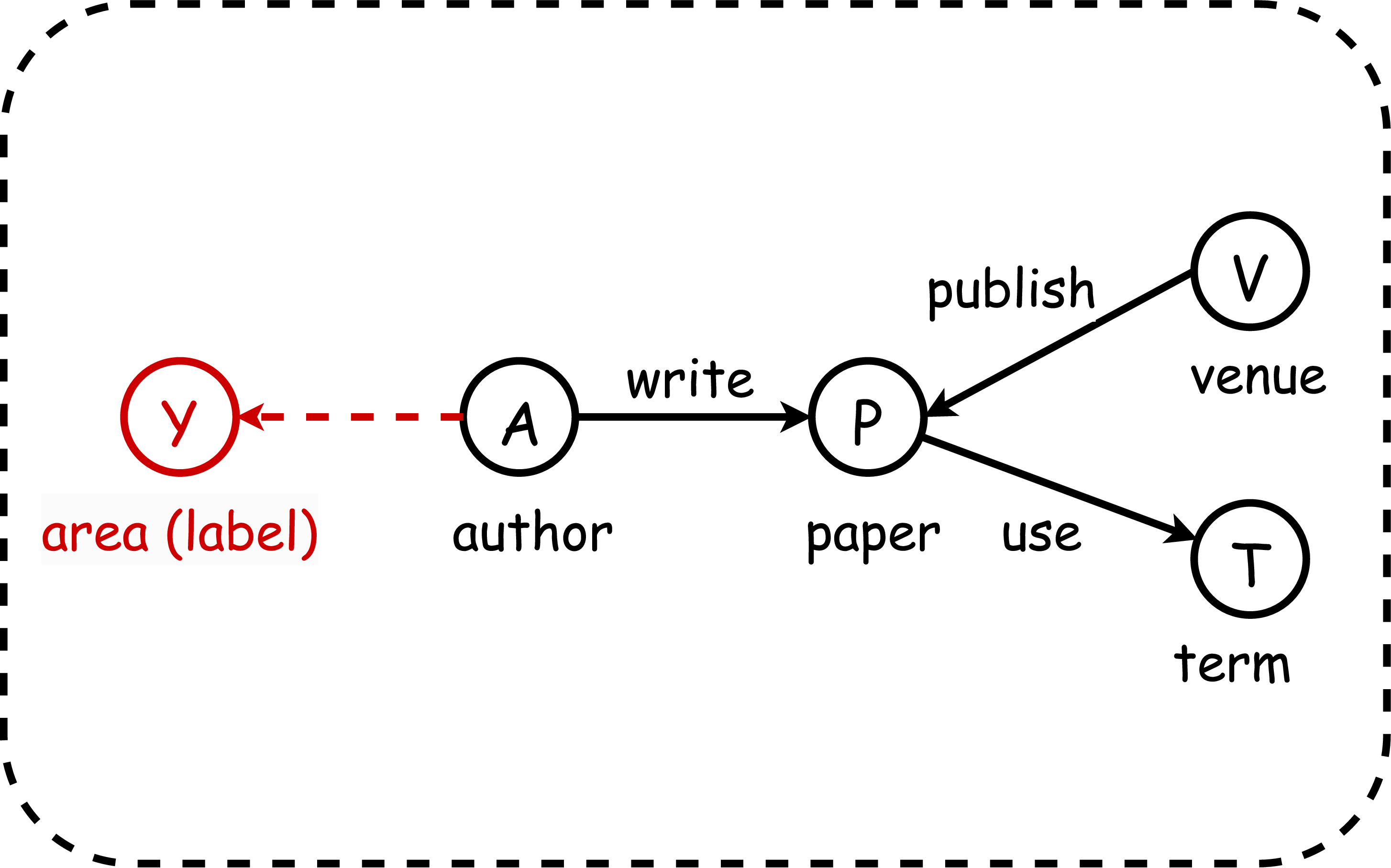}
        
        \caption{DBLP}
        \label{fig:dblp-schema}
    \end{subfigure}
    \hfill
    \begin{subfigure}[b]{0.3\textwidth}
        \centering
        \includegraphics[width=1.0\textwidth]{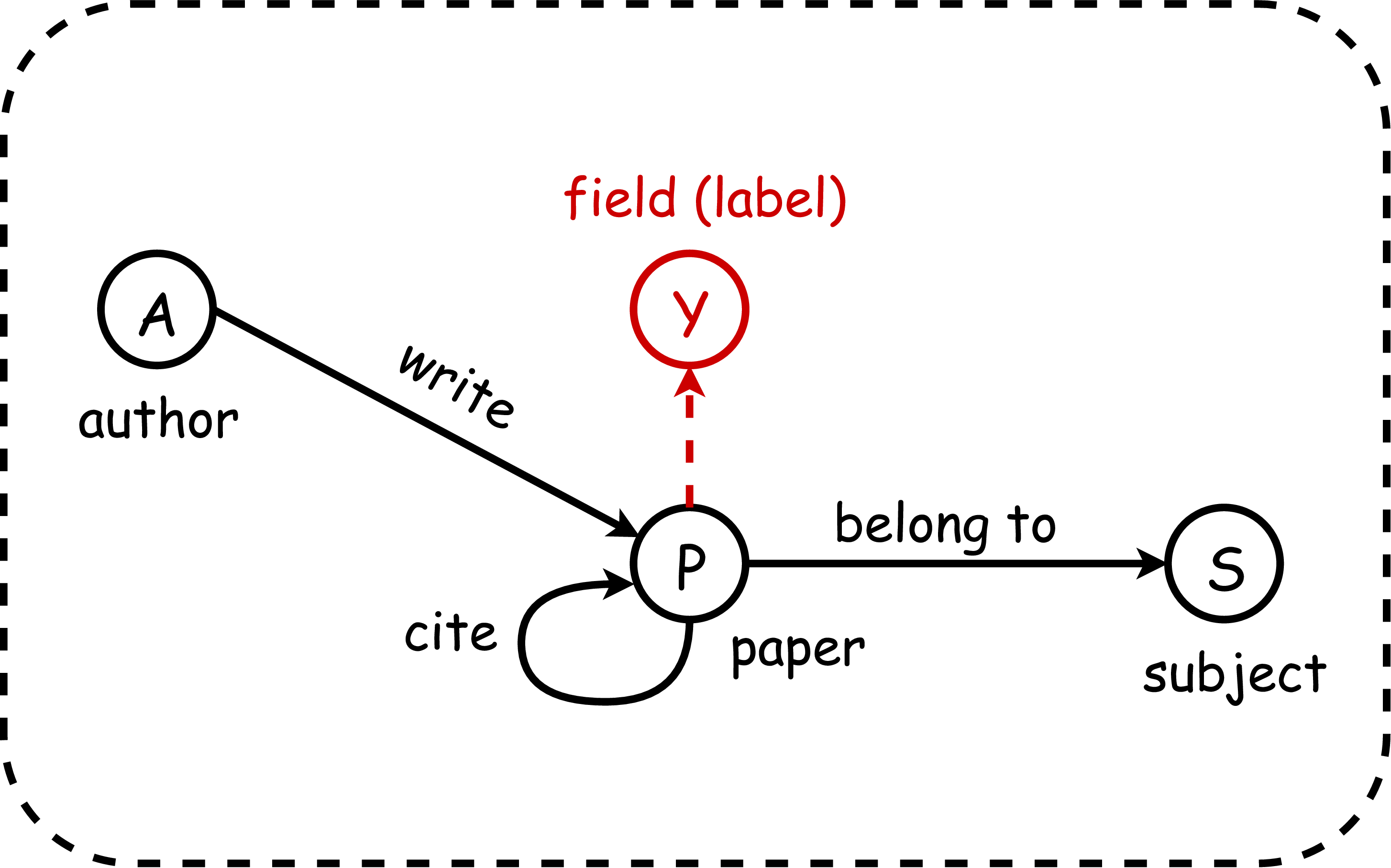}
        
        \caption{ACM}
        \label{fig:acm-schema}
    \end{subfigure}
    \hfill
    \begin{subfigure}[b]{0.3\textwidth}
        \centering
        \includegraphics[width=1.0\textwidth]{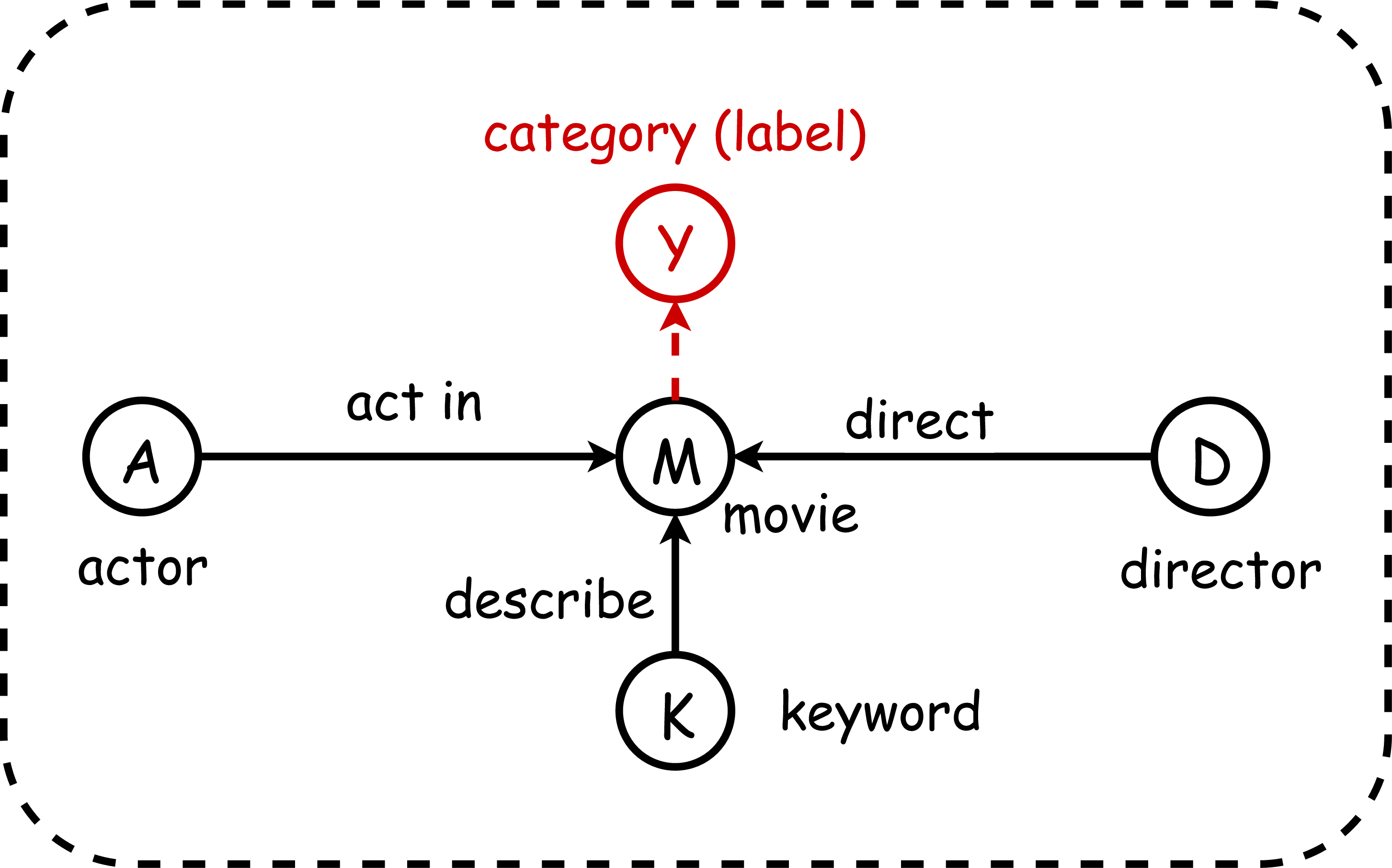}
        
        \caption{IMDB}
        \label{fig:imdb-schema}
    \end{subfigure}
    \caption{Graph Schemas. Black circles represent node types in the datasets. Black arrows describe directional relations between node types. The shown relations and their inverse relations together form all the edge types. Red circles and arrows define the node property prediction tasks, starting from the target node types and ending with the predicted properties. Generally, relations can be represented by two consecutive node types without confusion. For example, ``AP'' in DBLP means $Author \xrightarrow[]{write} Paper$. However, there exists one special case. Since the relation $Paper \xrightarrow[]{cite} Paper$ in ACM starts and ends with the same node type, ``PP'' can not clearly identify the relation. Therefore, we use ``PcP'' and ``PrP'' to represent $Paper \xrightarrow[]{cite} Paper$ and $Paper \xrightarrow[]{cited\; by} Paper$, respectively. Based on this denotation convention, we can further represent a meta-path by a sequence of node types. For instance, ``APA'' can represent $Author \xrightarrow[]{write} Paper\xrightarrow[]{written\; by} Author$ in DBLP and ``PcPrP'' can represent $Paper \xrightarrow[]{cite} Paper\xrightarrow[]{cited\; by} Paper$ in ACM.}
    \label{fig:graph-schema}
\end{figure}

Three open benchmark datasets including DBLP, ACM, and IMDB from the Heterogeneous Graph Benchmark (HGB)~\citep{simplehgn} are employed in this work. 
These datasets were chosen for their ability to present a relatively complete system. 
Table~\ref{tab:datasets} summarizes the brief statistical information of the three datasets, while Figure~\ref{fig:graph-schema} displays the graph schema for each dataset. Below are specific introductions for the three datasets:  
    
\begin{enumerate}
    \item \textbf{DBLP} is a computer science bibliography website~\footnote{\url{https://dblp.org/}}. The used dataset is a subset of it. Its graph schema is shown in Figure~\ref{fig:dblp-schema}. The dataset comprises four types of nodes: \textit{author} (N=4,057), \textit{paper} (N=14,328), \textit{term} (N=7,723), and \textit{venue} (N=20). Additionally, there are three types of directed relations connecting two node types: \textit{an author writes a paper} (N=19,645), \textit{a venue publishes a paper} (N=14,328), and \textit{a paper uses a term} (N=85,810). The papers' feature vectors are created based on the bag-of-words representation of their titles, while the authors' feature vectors are constructed based on the bag-of-words representation of their research keywords. The terms' feature vectors are represented by pre-trained GloVe vectors~\citep{pennington-etal-2014-glove}, and the feature vectors of venues are represented by one-hot encoded vectors. The authors are manually labeled into four areas: \textit{Database}, \textit{Data Mining}, \textit{Machine Learning}, and \textit{Information Retrieval}. The task is to predict the author's area. 
    \item \textbf{ACM} is an international learned society for computing~\footnote{\url{https://www.acm.org/}}. The used dataset is a subset of the ACM Digital Library. Its graph schema is shown in Figure~\ref{fig:acm-schema}. The dataset contains three types of nodes: \textit{paper} (N=3,025), \textit{author} (N=5,959), and \textit{subject} (N=56). Additionally, there are three types of directed relations connecting two node types: \textit{a paper cites a paper} (N=5,343), \textit{an author writes a paper} (N=9,949), and \textit{a paper belongs to a subject} (N=3,025). Each paper or author or subject node is associated with a bag-of-words vector formed by 1,902 representative keywords. The papers are categorized into three classes, \textit{Database}, \textit{Wireless Communication}, and \textit{Data Mining}. The task is to predict the paper's category. 
    \item \textbf{IMDB} is an online database of information related to films~\footnote{\url{https://www.imdb.com/}}. The used dataset was a subset of it. Its graph schema is shown in Figure~\ref{fig:imdb-schema}. The dataset contains four types of nodes: \textit{movie} (N=4,932), \textit{director} (N=2,393), \textit{actor} (N=6,124), and \textit{keyword} (N=7,971). In addition, there are three types of directed relations connecting two node types: \textit{an actor act in a movie} (N=14,779), \textit{a director directs a movie} (N=4,932), and \textit{a keyword describes a movie} (N=23,610). The movie's feature vectors are bag-of-word representations of their plot keywords. The features of director and actor nodes are aggregated features from their associated movies. The movies are divided into five classes, namely \textit{Action}, \textit{Comedy}, \textit{Drama}, \textit{Romance}, and \textit{Thriller}. The task is to predict the movie's category. 
\end{enumerate}

In the i.i.d setting, we follow the data splits used in HGB\citep{simplehgn,sehgnn}, where  node labels are split according to 24\% for training, 6\% for validation and
70\% for test in each dataset. For the concerned o.o.d settings, we consider three types of bias that are ubiquitous in graph mining:
\begin{enumerate} 
    \item \textbf{Homophily bias:} Homophily is a principle of graphs whereby linked nodes often belong to the same class or have similar features. Many existing graph algorithms implicitly assume strong homophily, thus they can fail to generalize to graphs with heterophily (or low/medium level of homophily)~\citep{nips_homo,homo_aaai}. \SecondRevise{In the context of heterogeneous graphs, we can define the homophily level of a target node as the label consistencies between the target node and its neighbors based on various meta-paths, where the meta-paths start with and end with the same node type as the target node.} 
    
    \item \textbf{Degree bias:} Degrees of nodes often obey a long-tailed or skewed distribution~\citep{newman2003social,tail-gnn-kdd}. A model may have inferior performance on unseen nodes that have a different degree distribution compared to nodes in the training graph. \SecondRevise{In heterogeneous graphs, a node type can be associated with multiple relation types. Therefore, we define the degree size of a node as the degrees of the target nodes in terms of different relations.} 
    
    \item \textbf{Feature bias:} Collected nodes in many real scenes are inherently imbalanced on features or classes~\citep{GraphENS, CausPref}, hence HGNNs can be biased toward the dominant feature groups. \SecondRevise{The original node features in the above datasets are represented as bag-of-words, which are mostly sparse and high-dimensional. Therefore, we performed a Principal Component Analysis (PCA) on the node feature and we conducted the bias analysis based on the top 128 principal components.} 
\end{enumerate}

\begin{figure}[!htbp]
  \centering
  \includegraphics[width=0.9\linewidth]{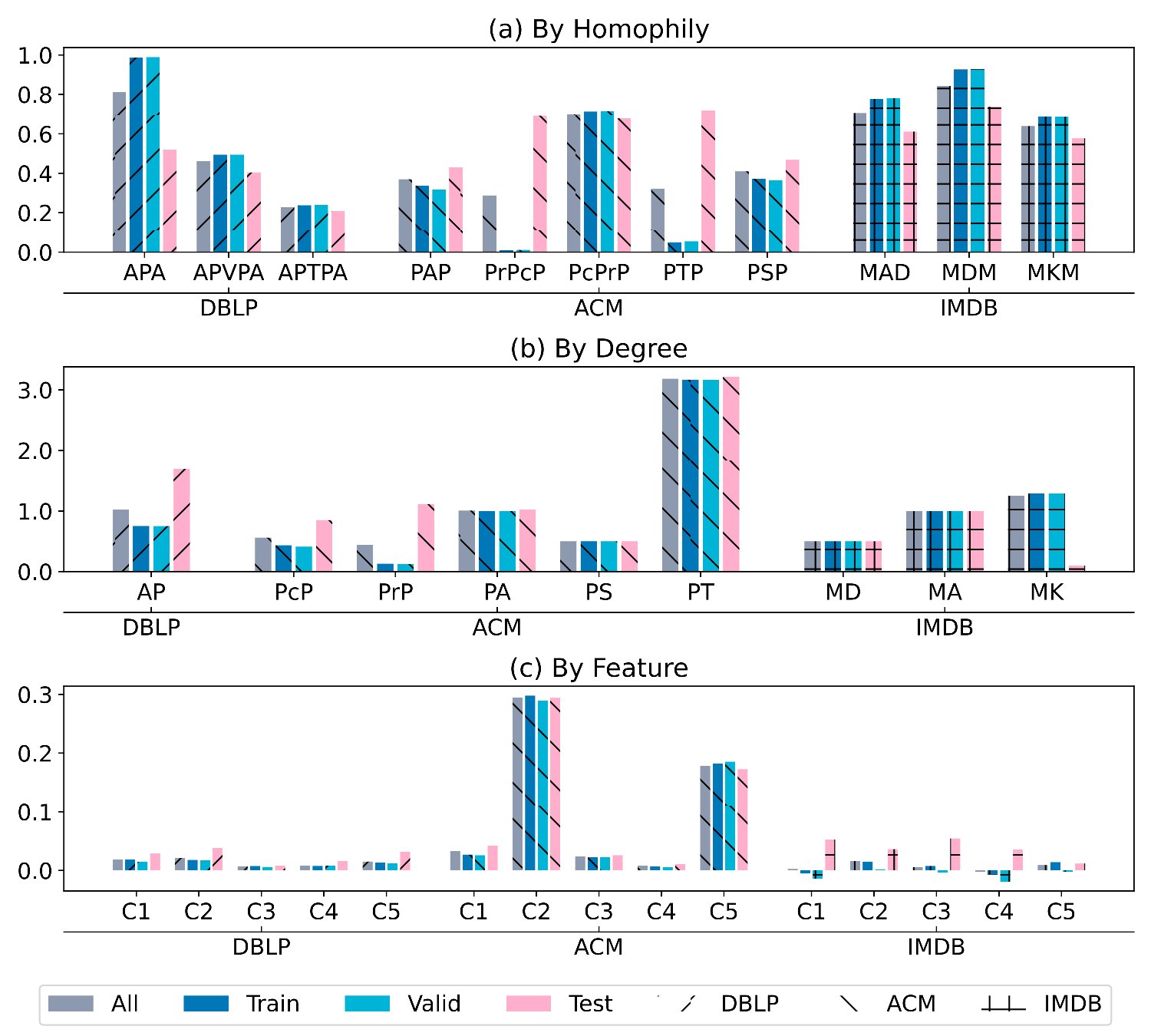}
  \caption{Visualization of data distribution difference in o.o.d settings. The greater the difference in height between the pink bar and the two corresponding light and deep blue bars, the greater the difference in the data distribution between the test set and the training/validation set. (a) homophily level (y-axis) in terms of specific meta-path (x-axis). (b) log-transformed degree size (y-axis) based on specific relation (x-axis). (c) \SecondRevise{component values (y-axis) of the top five principal components (denoted by C1 to C5) of the PCA-decomposed features of the target nodes (x-axis).} Please refer to Figure~\ref{fig:graph-schema} for interpretation of the abbreviated relations and meta-paths. 
  }
  \label{fig:ood_dist_vis}
\end{figure}

To simulate the above potential bias, for each dataset, we cluster the labeled nodes into two clusters by K-MEANS~\citep{kmeans} based on their homophily level, degree size, and feature value. 
Then, the cluster with a larger sample size is randomly divided into the training and the validation sets in a six-to-four ratio, and the other cluster is regarded as the test set. \SecondRevise{In Figrue~\ref{fig:ood_dist_vis}, we visualize the mean homophily levels, mean degree sizes, and mean feature values of all samples, training set samples, validation set samples, and test set samples under the o.o.d data splits for each dataset.} For example, in Figure~\ref{fig:ood_dist_vis}a, on the DBLP dataset, based on the meta-path ``Author-Paper-Author'', the proportion of co-authors with the same label as the target authors is around 80\% for all labeled authors. This proportion is nearly 100\% for the authors in the training and validation sets, but only about 50\% for the authors in the test set.

\subsubsection{Baselines}\label{sec:baselines} \revise{We have validated the superiority of HG-SCM by comparing it with the following two categories of HGNNs: (1) relation/meta-path fusion methods including the Heterogeneous Graph Neural Network (HetGNN)~\citep{hetgnn}, Heterogeneous Graph Attention Network (HAN)~\cite{han}, the Graph Transformer Network (GTN)~\cite{gtn}, and the Simple and Efficient Heterogeneous Graph Neural Network (SeHGNN)~\cite{sehgnn} and (2) layer-by-layer methods including the Relational Graph Convolutional Network (RGCN)~\cite{rgcn}, the Composition-based Multi-Relational Graph Convolutional Networks (CompGCN)~\cite{compgcn}, the Relation Structure-Aware Heterogeneous Graph Neural Network (RSHN)~\citep{rshn}, the Metapath Aggregated Graph Neural Network for Heterogeneous Graph Embedding (MAGNN)~\citep{magnn}, the Heterogeneous Graph Structural Attention Neural Network (HetSANN)~\citep{HetSANN}, the Heterogeneous Graph Transformer (HGT)~\cite{hgt} and the Simple Heterogeneous Graph Neural Network (SimpleHGN)~\cite{simplehgn}.} Note that HGT, SimpleHGN, and SeHGNN are the strongest state-of-the-art models recently.

\subsubsection{Reproducibility} All the experiments were run five times with random seeds from 0 to 4. In each experiment, \textit{we used the training set to fit the model, then used the validation set to tune the model's hyperparameters, and finally assessed and reported the performance of the model on the test set}. For a fair comparison, for all models, the hidden dimension was set to 64, the number of graph layers was searched from 1 to 4, and the batch size was searched in 128, 256, 512, and 1024. 
All experiments are conducted on a Ubuntu (18.04) server with a Tesla V100 GPU. Baseline models except SeHGNN were implemented via the DGL~\citep{dgl} package with the PyTorch (1.10) backend based on OpenHGNN~\citep{han2022openhgnn}. SeHGNN was implemented according to its official code~\citep{sehgnn}. We set other hyperparameters of baselines, e.g., negative slope in the LeakyReLU activation and dropout ratio, following the \citet{simplehgn} and \citet{sehgnn}. An optimizer AdamW~\cite{adamw} with a learning rate of 0.001 was utilized in all experiments, and an early-stopping strategy with a patience of 50 epochs based on the evaluation on the validation sets was applied. 
   
\SecondRevise{Following~\citet{simplehgn,sehgnn,ipm2023sentimentanalysis}, we will use Macro F1 and accuracy as evaluation metrics to present and discuss the experimental results. }

\subsection{Results and Discussion}\label{sec:results}

\subsubsection{Comparison in i.i.d Setting}\label{sec:iid-results}
Table~\ref{tab:all-iid} displays the comparison of HG-SCM with other baselines on the three benchmark datasets with their official i.i.d data splits, where the experimental results of all baselines except SeHGNN are referenced from \citet{simplehgn}. The experimental results of SeHGNN are obtained using its official code with the hidden dimension set to 64 (for a fair comparison). \emphasize{It is observed that HG-SCM can achieve best performance on almost all metrics in three datasets}. These experimental results prove that HG-SCM has a promising fitting ability to achieve competitive efficacy under conventional data distribution settings. 

\input{tables/ALL-iid.tex}
\input{tables/ood_concat.tex}

\begin{figure*}[t]
  \centering
  \includegraphics[width=1.0\linewidth]{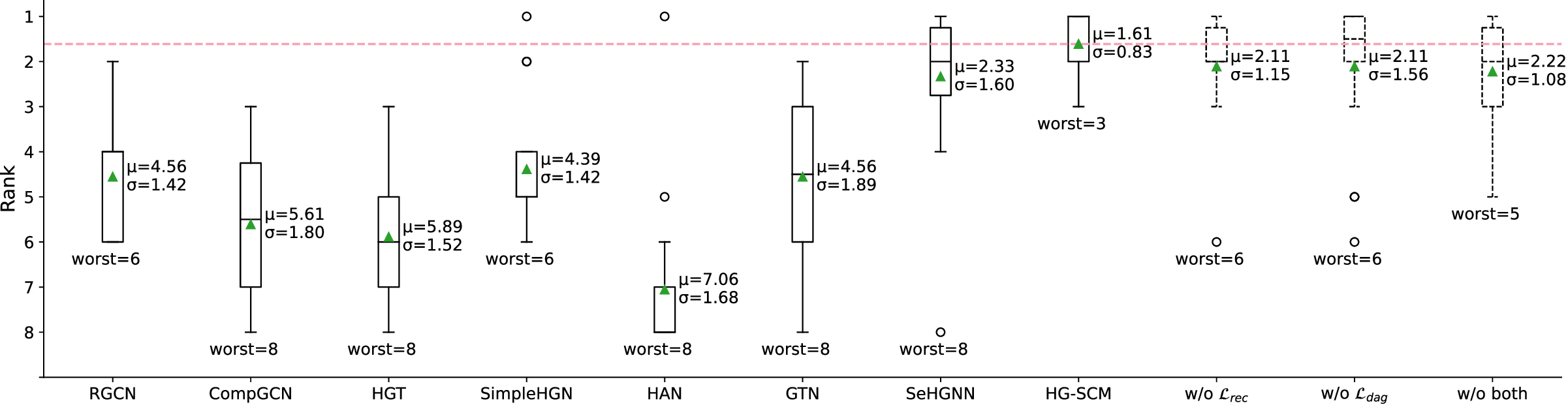}
  \caption{Box plot of the statistics of the performance rankings. $\mu$ and $\sigma$ represent the mean and the standard deviation of rankings in all the o.o.d experiments. $\circ$ represents fliers and ``\textit{worst}'' represents the lowest ranking. Dashed boxes are variants in the ablation study and their rankings are what they would get if they replaced the normal HG-SCM in the experiments.}
  \label{fig:rank}
\end{figure*}

\subsubsection{Comparison in o.o.d Setting}\label{sec:ood-results}
Tables~\ref{tab:ood-concat} report the experimental results in the three o.o.d data splits (introduced in Section~\ref{sec:dataset}) on three benchmark datasets. HG-SCM can outperform current SOTA models in most settings. Especially in the setting of o.o.d data split by homophily, HG-SCM consistently achieves the optimal performance. 
Notably, It can be observed that, though the SOTA methods have already achieved a high performance, HG-SCM can still push the boundary forward to a higher level. For example, on ACM dataset (under the o.o.d data split by homophily), the best baseline's performance is 96.49\% in Micro F1, HG-SCM can achieve 97.32\%.

Moreover, \emphasize{HG-SCM is the stablest model across these experiments, demonstrating its promising generalizability.} In contrast, the baseline models usually show significant differences in performance under different o.o.d settings. For example, GTN achieved very competitive performance in the degree o.o.d setting while it became a relatively weak baseline in other settings. As shown in Figure~\ref{fig:rank}, HG-SCM has kept its competitive performance in all settings. In detail, HG-SCM's ranking standard deviation is only 0.83, far smaller than the ranking standard deviation of other models. Additionally, the worst ranking HG-SCM reached is third, while all other models always have their worse ranking at a much lower level, e.g., sixth or even eighth place.

\subsubsection{Ablation Study}\label{sec:ablation-study}
We carried out additional experiments to explore the impacts of the two objectives introduced in Section~\ref{sec:optim-loss}: the reconstruction loss of the structural assignments, denoted by $\mathcal{L}_{rec}$, and the directed acyclic constraint loss, denoted by $\mathcal{L}_{dag}$.  

\begin{figure}[!htbp]
  \centering
  \includegraphics[width=0.8\textwidth]{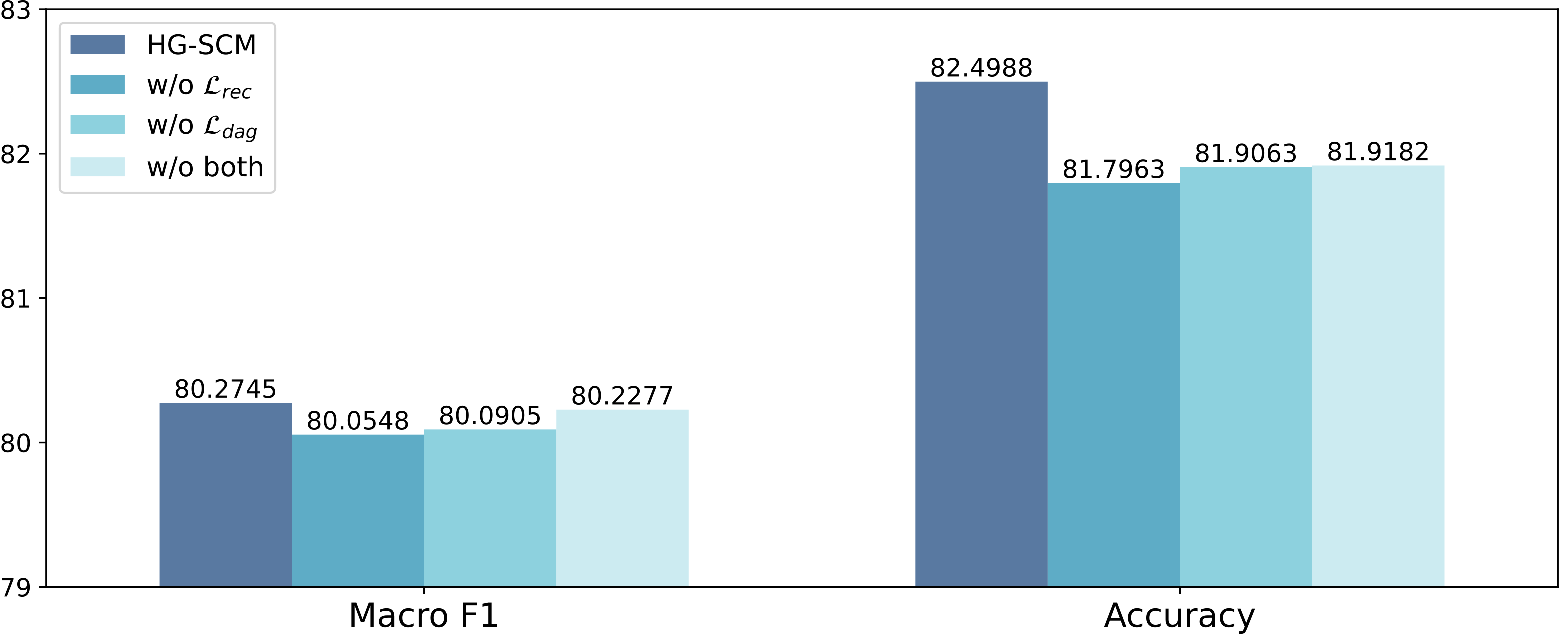}
  \caption{Ablation Study: the averaged performance of variants of HG-SCM under all o.o.d settings.}
  \label{fig:ablation}
\end{figure}

As shown in Figure~\ref{fig:ablation}, without the optimization objective(s) of causal structure, the average of the task performances would decline. Furthermore, Figure~\ref{fig:rank} suggests that, overall, removing any of the two optimization objectives can decrease the average ranking and increase the standard deviation of the ranking. Specifically, removing $\mathcal{L}_{rec}$ got an average ranking of 2.11 with a standard deviation of 1.15, and removing $\mathcal{L}_{dag}$ got an average ranking of 2.11 with a standard deviation of 1.56. Moreover, removing both of the two objectives led to worse results, i.e., an average ranking of 2.22. \emphasize{These results demonstrate the importance of the objective of learning an underlying causal structure for the generalizability of the HG-SCM.}

\revise{
An intriguing observation is that the average performance is higher when neither optimization objective is used compared to using only one of them. This highlights the interconnected nature of these two objectives. It is only when both objectives are simultaneously considered that the modeling of a Structural Causal Model (SCM) becomes possible. Therefore, when employing a single objective in isolation, it does not represent a weaker causal structure optimization objective and can even potentially have a negative impact, leading to inferior performance. 
We can infer the potential reasons for the decline in performance. Firstly, when solely considering the Directed Acyclic Graph (DAG) loss without incorporating the reconstruction loss, the model will tend to learn simple causal relationships that satisfy the DAG conditions but do not accurately reflect the true underlying causal relationships as deep learning models have a tendency to learn shortcuts~\citep{nature2020shortcut}. For instance, the model might learn that all variables are causes of the label variable. However, such a causal relationship deviates significantly from the true one and can detrimentally affect predictions on unseen data, particularly in out-of-distribution scenarios. 
Secondly, when solely considering the reconstruction loss without incorporating the DAG loss, the model essentially becomes an autoencoder trained on the training set. This increases the risk of overfitting the training data, resulting in diminished performance on the test set, especially in out-of-distribution scenarios. 
Therefore, using a single loss alone does not achieve the desired optimization process for the causal modeling process. 
}

\begin{figure*}[!htbp]
  \centering
  \includegraphics[width=1.0\linewidth]{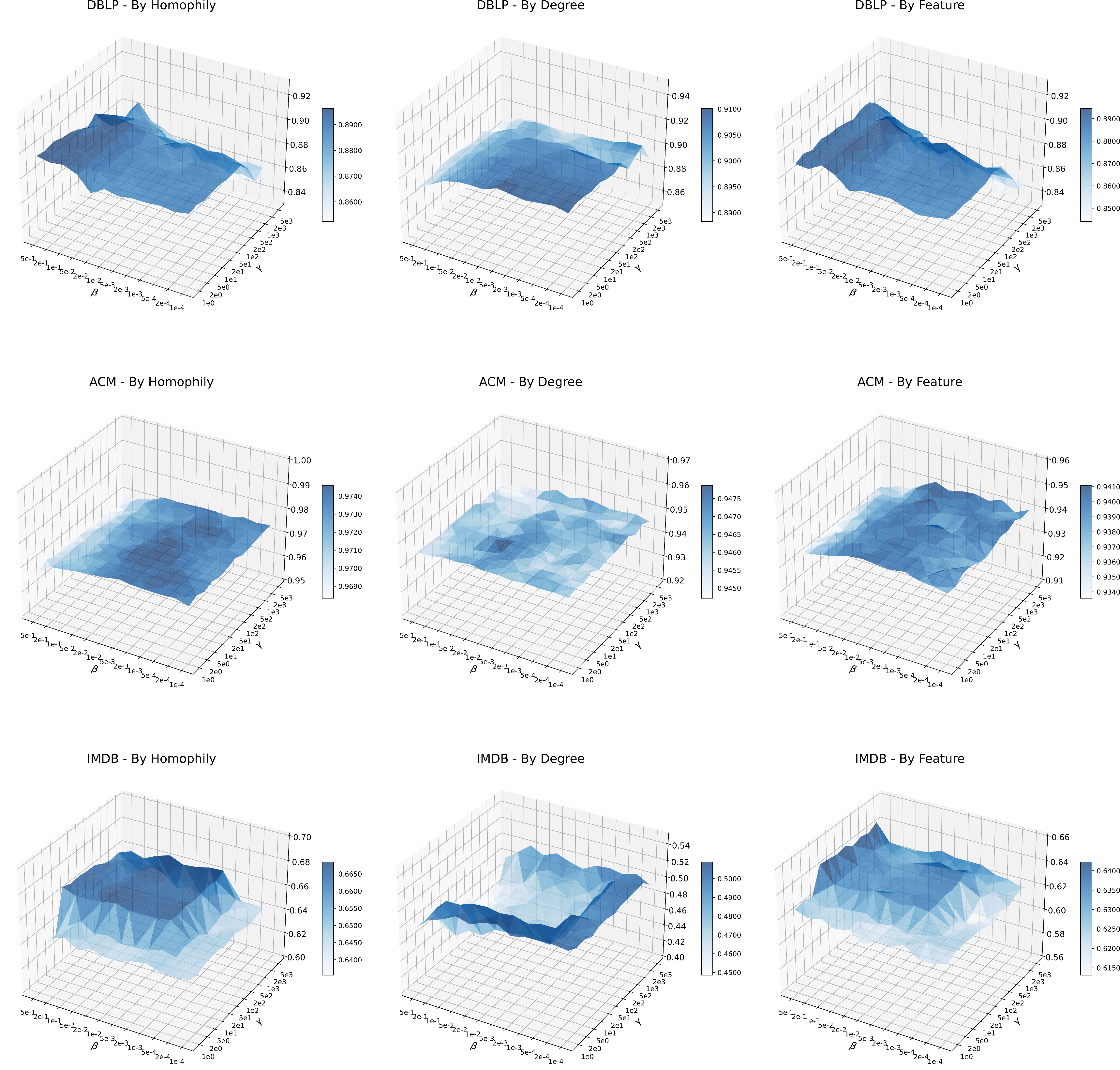}
  \caption{Sensitivity Analysis of $\beta$ and $\gamma$.}
  \label{fig:sensitivity}
\end{figure*}

\subsubsection{Hyperparameters Sensitivity Analysis}\label{sec:hyperparam-study}
Two hyper-parameters can be tuned in this work: the weight of the objective $\mathcal{L}_{rec}$, i.e., $\beta$, and the weight of the objective $\mathcal{L}_{dag}$, i.e., $\gamma$. HG-SCM’s sensitivity with respect to $\beta$ and $\gamma$ is presented in Figure~\ref{fig:sensitivity}. \revise{We only change the values of these two hyperparameters in this analysis, and other hyperparameters remain in the same setting as stated in Section~\ref{sec:ood-results}.} The average of the Micro F1 and Macro F1 under fixed hyperparameters were reported. We varied the values of $\beta$ from 1e-4 to 5e-1 and varied the values of $\gamma$ from 1e0 to 5e3. 

Although the degree of impact of hyperparameters on task performance varies across datasets, the improvement of model performance has a clear direction in all settings. For example, on the setting of DBLP with an o.o.d data split based on homophily, model performance is positively correlated with the value of $\beta$ and negatively correlated with the values of $\gamma$.  Based on this observation, \emphasize{within the appropriate hyperparameter interval, HG-SCM can be quite robust to the adjustment of the hyperparameters.}

\begin{figure}[htbp]
\centering
\includegraphics[width=1.0\textwidth]{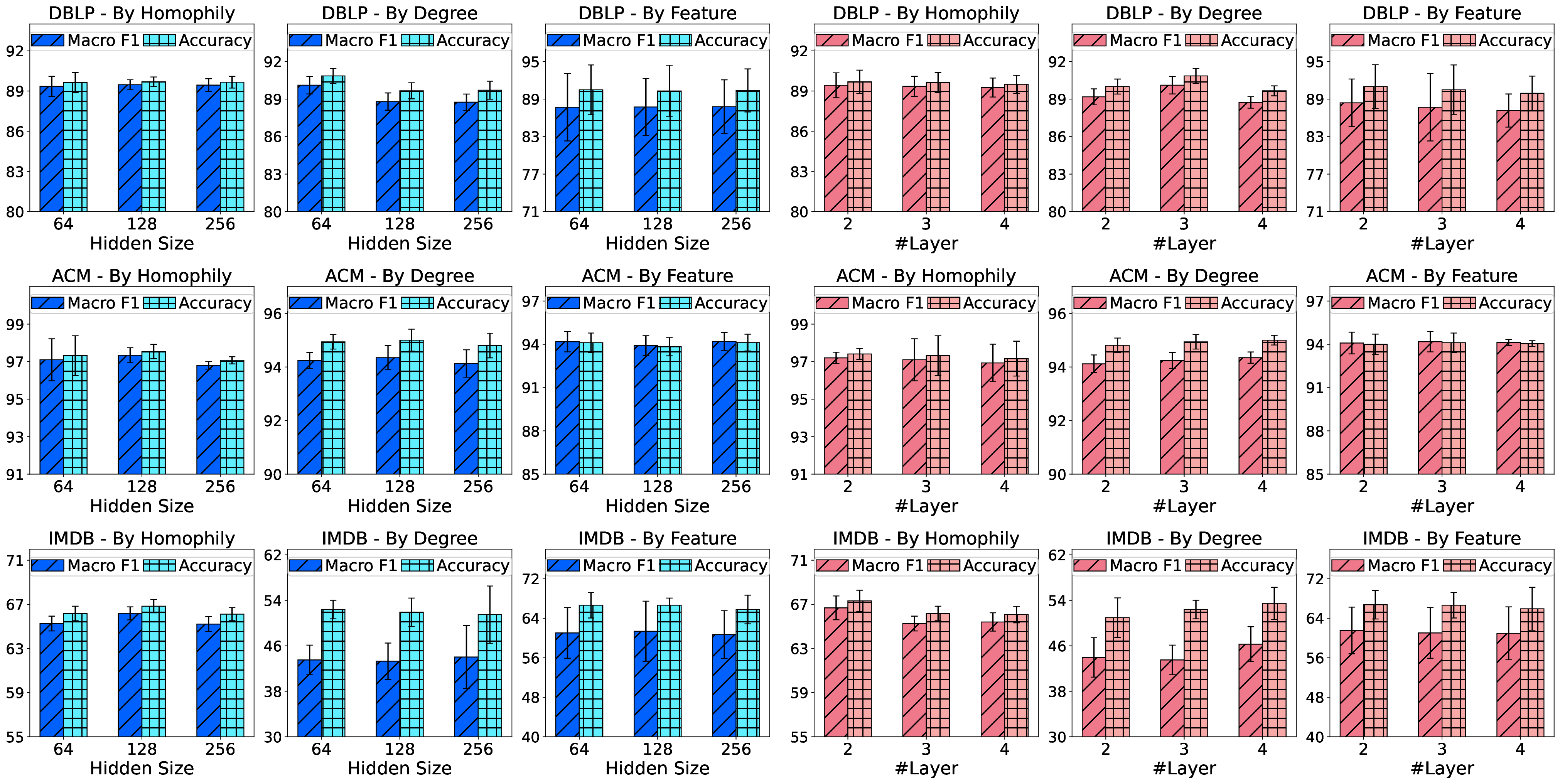}
\caption[]{
    \revise{Sensitivity Analysis of the hidden size of the hidden layer and the number of hidden layers (\#Layer) in MLP.}
} 
\label{fig:mlp_sensitivity}
\end{figure}

\revise{
We conducted a sensitivity analysis to examine the impact of the MLP configuration on our proposed model. In the experiments described in section~\ref{sec:ood-results} and Table~\ref{tab:ood-concat}, we kept the hidden size in hidden layers of the MLP fixed at 64 and the number of hidden layers fixed at 3. Here, we varied the hidden size from 64 to 256 and the number of hidden layers from 2 to 4 to explore different settings. 
The experimental results of all the different MLP configurations across all the datasets and o.o.d scenarios can be found in Figure~\ref{fig:mlp_sensitivity}. We observed the following:
\begin{itemize}
	\item Our proposed model demonstrates robustness to changes in MLP settings. In most cases, the model exhibits consistent performance across different MLP configurations. In addition, the original experiment in Table~\ref{tab:ood-concat} yielded an average performance of 81.39. In the additional experiments here, where the hidden size was 128 or 256 and the number of hidden layers was 2 or 4, the average performance was 81.32. These two values are very close, which indicates that the model maintains state-of-the-art or competitive performance. 
	\item By tuning the MLP settings, we can further enhance the performance of our proposed model. In many cases shown in Figure~\ref{fig:mlp_sensitivity}, the model with a new MLP configuration outperforms the reported results in Table~\ref{tab:ood-concat}. For example, in the IMDB dataset with a homophily o.o.d setting, increasing the hidden size from 64 to 128 improves the model's Macro F1 score from 65.26 to 66.18. Similarly, reducing the number of hidden layers from 3 to 2 increases the model's Macro F1 score from 65.26 to 66.68.
\end{itemize}
}

\subsubsection{Variants Exploration}\label{sec:variants}
In this section, we analyze the impact of replacing simple modules with more complex ones on the performance of the original model. We introduce two variants:
\begin{itemize}
    \item HG-SCM-TC: We change the encoding method of the neighbor set in Equation~\ref{eq:setenc} from simple average pooling to a Transformer Convolution operator derived from \citep{transformerconv}. For each sample, this variant dynamically adjusts the weights of nodes in a neighbor set based on self-attention mechanisms~\citep{vaswani2017attention} among the nodes. 
    \item HG-SCM-ST: We replace Equation~\ref{eq:assign_detail} with an assignment function inspired by SetTransformer~\citep{settransformer}. SetTransformer is designed to model interactions among elements in an input set, utilizing attention mechanisms in both its encoder and decoder. This aligns well with the purpose of Equation~\ref{eq:assign} which encodes the causes of a variable and then decodes its value. For each sample, we conduct an element-wise product between the causal DAG $\mA$ and the self-attention matrix~\citep{vaswani2017attention} among the variables in the SetTransformer. 
\end{itemize}
\input{tables/variants}

The results in Table~\ref{tab:variant} indicate that HG-SCM-ST can outperform HG-SCM in terms of overall performance and stability across experimental settings. In addition, HG-SCM-TC achieved optimal performance under a few settings but showed a decrease in overall performance, and it also exhibited significant deterioration in several settings, particularly on the DBLP dataset. Hence, the following conclusions can be drawn:
\begin{itemize}
    \item The use of more complex modules in neighbor set encoding may lead to reduced generalizability and stability of the model across different datasets or experimental conditions. This could be attributed to the fine-grained nature of changes in the neighbor set, with complex modules being prone to overfitting such information~\cite{nature2020shortcut}, thereby diminishing generalizability. Furthermore, this overfitting tendency may amplify the model's preference for specific datasets or data distributions, leading to reduced stability across conditions.
    \item Employing more complex modules in the assignment function of SCM has the potential to enhance the overall generalizability and stability of the model. This is likely because complex assignment functions are better equipped to capture intricate causal mechanisms in the real world, thereby facilitating the overall learning process of the structural causal model. 
\end{itemize}

\begin{figure}[h]
  \centering
  \includegraphics[width=1.0\textwidth]{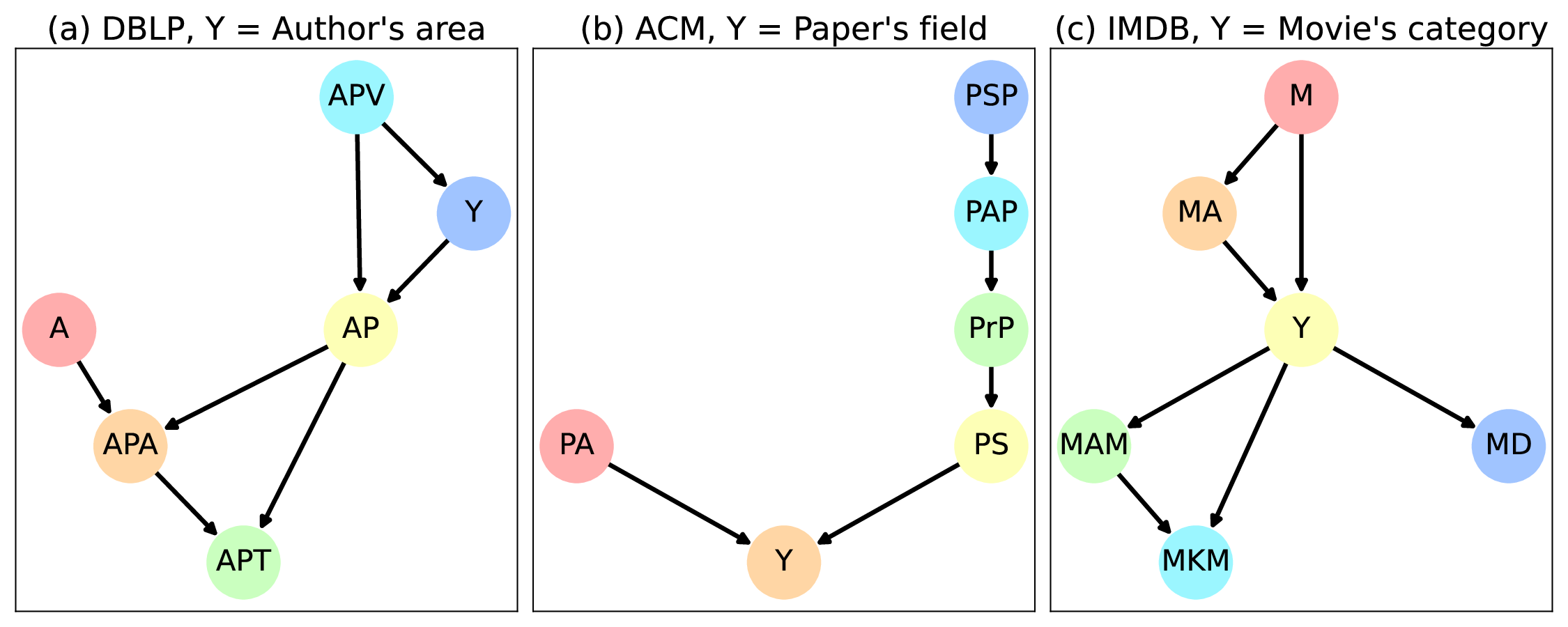}
  \caption{Learned DAGs for different tasks: these DAGs can provide explanations of the possible causal relationships in the task. Please refer to Figure~\ref{fig:graph-schema} for interpretation of the abbreviated relations and meta-paths.}
  \label{fig:learned-dag}
\end{figure}

\subsubsection{Case Study}\label{sec:case-study}
As aforementioned, HG-SCM can provide in-depth interpretations on the task level through the learned causal DAGs. Specifically, after obtaining the optimized matrix $\mA$ (mentioned in Section~\ref{sec:causal_structure_modeling}), we remove edges in $\mA$ in order of increasing absolute values of the weights until $\mA$ satisfies the directed acyclic property. This ensures that the remaining DAGs characterize relatively important and reliable causal structures. These trimmed DAGs are displayed in Figure~\ref{fig:learned-dag}.\emphasize{ We can find that they exhibit extremely high interpretability and can be partially aligned with our human knowledge/cognition.}  

For instance, in Figure~\ref{fig:learned-dag}a, based on the learned DAG, an author's research area is considered to be labeled based on the venues in which this author has published papers, i.e., APV $\rightarrow$ Y (author's area). \SecondRevise{When authors publish their papers in specific venues, it usually signifies their active engagement within specific academic communities, as well as the recognition they garner within those communities~\citep{jasist_author_jounral_preference,sciento_journal_selection}. Therefore, using venues to predict an author's primary research area can be quite intuitive and widely accepted. This rule probably becomes more reasonable in the current landscape of academia, where interdisciplinary research is prevalent and papers across different venues often exhibit considerable semantic similarity~\citep{joi_phrase_intersection, joi_interdisciplinary_author_or_reference}. Figuring out an author's primary research area through the venues they contribute can be advantageous in avoiding confusion arising from subtle differences in the meanings of their papers.} 
Moreover, the author's research area, along with the target venue, can affect the content of the author's papers, i.e.,  APV $\rightarrow$ AP $\leftarrow$ Y. \SecondRevise{As demonstrated by \citet{joi_linguistic_and_journal}, venues within the same field still exhibit significant differences in linguistic features. In practice, authors tailor their writing styles to align with the preferences of specific venues, thereby enhancing the likelihood of paper acceptance by the respective readership. }  
Additionally, the author and the content of the papers may determine the academic partners of the author, i.e.,  A $\rightarrow$ APA $\leftarrow$ AP. \SecondRevise{This is supported by research such as that conducted by \citet{jasist_factors_collaboration}, which reveals that personal compatibility and work connections can impact collaboration. } 
Lastly, the authors and the content of the papers may influence the terms used in writing, i.e., APA $\rightarrow$ APT $\leftarrow$ AP, \SecondRevise{which is consistent with prior studies indicating that the choice of author keywords is not solely shaped by the paper content but is also substantially influenced by authors' prior knowledge and backgrounds~\citep{joi_keyword_selection}.} 
Notably, the causality embodied in this learned DAG is consistent with the logic of reality and human perception. Similarly, Figure~\ref{fig:learned-dag}b and Figure~\ref{fig:learned-dag}c provide explanations of the possible causal reasoning logic for predicting a paper's field and a movie's category, respectively. A paper's field could be determined by its subjects and authors. The plot of a movie and its actors usually determine the movie’s category.  

It is crucial to highlight that the automatically learned DAGs not only provide a high degree of interpretability but also present researchers with the opportunity to refine and improve the model through such feedback. Based on the DAGs, one can check whether the inference logic of the model is correct, thereby guaranteeing the model trustworthiness, which is of paramount importance for some real-world applications.

%% file: tables/datasets.tex
\begin{table}
  \centering
  \caption{
    Dataset statistics.
  }
  \label{tab:datasets}%
    \begin{threeparttable}
    \setlength{\tabcolsep}{8pt}
  \renewcommand{\arraystretch}{1.}
    \begin{tabular}{ccccccc}
    \toprule
    Dataset & \#Nodes & \#Node Types & \#Edges & \#Edge Types & Target  & \#Classes \\
    \midrule
    DBLP  & 26,128 & 4     & 239,566 & 6     & author & 4 \\
    ACM   & 9,040 & 3     & 36,634 & 6     & paper & 3 \\
    IMDB  & 21,420 & 4     & 86,642 & 6     & movie & 5 \\
    \bottomrule
    \end{tabular}%

  \end{threeparttable}
\end{table}%

%% file: tables/ALL-iid.tex
\begin{table}[!hb]
  \centering
  \caption{
    Comparison on the three benchmark datasets under the official i.i.d data splits in HGB~\citep{simplehgn}. \underline{\textbf{Bold}} and \underline{underline} indicate the best and the top 3 performance, respectively.
  }
  \label{tab:all-iid}%
  
    \begin{tabular}{ccccccc}
    \toprule
         \multirow{2}{*}{Methods} & \multicolumn{2}{c}{DBLP} & \multicolumn{2}{c}{ACM} & \multicolumn{2}{c}{IMDB} \\
    \cmidrule{2-7}  & Macro F1 & \SecondRevise{Accuracy} & Macro F1 & \SecondRevise{Accuracy} & Macro F1 & \SecondRevise{Accuracy} \\
    \midrule
    RGCN  & 91.52±0.50 & 92.07±0.50 & 91.55±0.74 & 91.41±0.75 & 58.85±0.26 & 62.05±0.15  \\
    HAN   & 91.67±0.49 & 92.05±0.62 & 90.89±0.43 & 90.79±0.43 & 57.74±0.96 & 64.63±0.58  \\
    GTN   & 93.52±0.55 & 93.97±0.54 & 91.31±0.70 & 91.20±0.71 & 60.47±0.98 & 65.14±0.45  \\
    RSHN  & 93.34±0.58 & 93.81±0.55 & 90.50±1.51 & 90.32±1.54 & 59.85±3.21 & 64.22±1.03 \\
    HetGNN & 91.76±0.43 & 92.33±0.41& 85.91±0.25 & 85.05±0.25 & 48.25±0.67 & 51.16±0.65 \\
    MAGNN & 93.28±0.51 & 93.76±0.45 & 90.88±0.64 & 90.77±0.65  & 56.49±3.20 & 64.67±1.67 \\
    HetSANN & 78.55±2.42 & 80.56±1.50 & 90.02±0.35 & 89.91±0.37 & 49.47±1.21 & 57.68±0.44 \\
    HGT   & 93.01±0.23 & 93.49±0.25 & 91.12±0.76 & 91.00±0.76 & 63.00±1.19 & \underline{67.20±0.57} \\
    SimpleHGN & \underline{94.01±0.24}  & \underline{94.46±0.22} & \underline{93.42±0.44} & \underline{93.35±0.45}  & \underline{63.53±1.36} & \underline{\textbf{67.36±0.57}} \\
    SeHGNN & \underline{94.49±0.20} & \underline{94.89±0.18} & \underline{93.60±0.44} & \underline{93.51±0.45}  & \underline{64.67±0.29} & 65.98±0.12 \\
    \midrule
    HG-SCM   & \underline{\textbf{94.51±0.15}} & \underline{\textbf{94.90±0.15}}& \underline{\textbf{93.64±0.31}} & \underline{\textbf{93.56±0.32}}  & \underline{\textbf{65.34±0.33}} & \underline{66.90±0.61} \\
    \bottomrule
    \end{tabular}%
  
\end{table}%

%% file: tables/ood_concat.tex
\begin{table}[htbp]
  \centering
  \caption{
    Comparison on the three benchmark datasets under the three o.o.d data splits. \underline{\textbf{Bold}} and \underline{underline} indicate the best and the top 3 performance, respectively.
  }
  \label{tab:ood-concat}%
    \begin{threeparttable}
    \setlength{\tabcolsep}{3pt}
  \renewcommand{\arraystretch}{1.}
    \begin{tabular}{ccccccc}
        \toprule
    \multicolumn{1}{c}{\multirow{2}[2]{*}{DBLP}} & \multicolumn{2}{c}{By Homophily} & \multicolumn{2}{c}{By Degree} & \multicolumn{2}{c}{By Feature	} \\
\cmidrule{2-7}          & Macro F1 & \SecondRevise{Accuracy} & Macro F1 & \SecondRevise{Accuracy} & Macro F1 & \SecondRevise{Accuracy} \\
    \midrule
    RGCN  & 88.23±0.83 & 88.55±0.73 & 88.87±0.37 & 89.76±0.27 & \underline{87.00±3.47} & \underline{90.20±3.05} \\
    CompGCN & 85.27±1.86 & 85.82±1.61 & 86.65±0.89 & 87.59±0.95 & 85.65±3.89 & \underline{89.35±2.67} \\
    HGT   & 88.48±1.10 & 88.98±1.02 & 89.37±0.80 & 90.27±0.77 & 85.00±4.77 & 88.65±3.64 \\
    SimpleHGN & \underline{89.06±1.17} & \underline{89.40±1.08} & 89.42±1.29 & 90.32±1.24 & 85.27±5.22 & 89.11±4.31 \\
    \midrule
    HAN   & 82.42±1.33 & 82.81±1.22 & 85.41±1.39 & 86.42±1.38 & 83.72±4.33 & 87.79±3.43 \\
    GTN   & 88.38±1.09 & 88.70±1.07 & \underline{90.26±0.80} & \underline{90.98±0.74} & 84.29±2.71 & 88.00±3.13 \\
    SeHGNN & \underline{88.85±0.30} & \underline{89.14±0.29} & \underline{\textbf{90.88±0.32}} & \underline{\textbf{91.56±0.31}} & \underline{86.05±5.12} & 89.31±4.37 \\
    \midrule
    HG-SCM   & \underline{\textbf{89.35±0.75}} & \underline{\textbf{89.63±0.75}} & \underline{90.11±0.70} & \underline{90.85±0.60} & \underline{\textbf{87.69±5.39}} & \underline{\textbf{90.48±3.98}} \\
    \midrule
    \midrule
    \multicolumn{1}{c}{\multirow{2}[2]{*}{ACM}} & \multicolumn{2}{c}{By Homophily} & \multicolumn{2}{c}{By Degree} & \multicolumn{2}{c}{By Feature	} \\
\cmidrule{2-7}          & Macro F1 & \SecondRevise{Accuracy} & Macro F1 & \SecondRevise{Accuracy} & Macro F1 & \SecondRevise{Accuracy} \\
    \midrule
    RGCN  & 92.65±1.16 & 93.13±1.10 & 92.93±0.93 & 93.72±0.80 & 92.38±0.93 & 92.25±0.96 \\
    CompGCN & \underline{94.42±0.58} & \underline{94.84±0.55} & 92.86±0.65 & 93.65±0.55 & 91.89±0.22 & 91.79±0.26 \\
    HGT   & 91.87±0.74 & 92.28±0.73 & 92.07±0.75 & 92.76±0.73 & 90.72±1.37 & 90.63±1.38 \\
    SimpleHGN & 93.43±1.43 & 93.86±1.36 & 92.54±1.01 & 93.33±0.83 & 91.65±1.12 & 91.57±1.00 \\
    \midrule
    HAN   & 91.72±1.64 & 92.18±1.66 & 89.69±1.47 & 90.80±1.51 & 91.11±1.26 & 90.97±1.29 \\
    GTN   & 93.94±1.96 & 94.36±1.87 & \underline{93.56±0.50} & \underline{94.29±0.43} & \underline{92.54±0.48} & \underline{92.43±0.56} \\
    SeHGNN & \underline{96.21±0.66} & \underline{96.49±0.62} & \underline{\textbf{94.26±0.50}} & \underline{94.90±0.44} & \underline{93.77±0.94} & \underline{93.66±0.93} \\
    \midrule
    HG-SCM   & \underline{\textbf{97.10±1.12}} & \underline{\textbf{97.32±1.06}} & \underline{94.24±0.30} & \underline{\textbf{94.94±0.27}} & \underline{\textbf{94.18±0.70}} & \underline{\textbf{94.11±0.67}} \\
    \midrule
    \midrule
    \multicolumn{1}{c}{\multirow{2}[2]{*}{IMDB}} & \multicolumn{2}{c}{By Homophily} & \multicolumn{2}{c}{By Degree} & \multicolumn{2}{c}{By Feature	} \\
\cmidrule{2-7}          & Macro F1 & \SecondRevise{Accuracy} & Macro F1 & \SecondRevise{Accuracy} & Macro F1 & \SecondRevise{Accuracy} \\
    \midrule
    RGCN  & \underline{50.49±2.49} & \underline{55.77±1.43} & 34.55±3.89 & 51.85±2.10 & 55.79±9.63 & 64.52±6.07 \\
    CompGCN & 41.67±0.78 & 49.74±0.95 & 30.63±3.82 & 49.76±4.07 & \underline{56.93±6.21} & 62.79±5.49 \\
    HGT   & 49.01±0.96 & 54.85±0.70 & 30.04±1.91 & 49.75±2.42 & 56.77±6.05 & \underline{65.59±4.11} \\
    SimpleHGN & 49.47±2.39 & 55.24±1.46 & 43.49±4.68 & \underline{\textbf{54.70±3.62}} & 56.56±5.91 & 63.99±5.91 \\
    \midrule
    HAN   & 48.92±2.04 & 54.07±1.40 & \underline{\textbf{44.33±5.06}} & 51.58±3.91 & 55.71±4.08 & 62.92±4.50 \\
    GTN   & 47.90±1.35 & 54.23±1.12 & 40.70±3.80 & \underline{54.13±2.49} & 55.54±5.25 & 63.97±5.63 \\
    SeHGNN & \underline{65.20±0.56} & \underline{65.86±0.58} & \underline{43.87±1.42} & 48.04±1.29 & \underline{\textbf{62.88±4.00}} & \underline{\textbf{67.31±3.43}} \\
    \midrule
    HG-SCM   & \underline{\textbf{65.26±0.67}} & \underline{\textbf{66.17±0.66}} & \underline{43.51±2.61} & \underline{52.37±1.63} & \underline{61.03±5.14} & \underline{66.64±2.58} \\
    \bottomrule
    \end{tabular}%

  \end{threeparttable}
\end{table}%

%% file: tables/variants.tex
\begin{table}[htbp]
  \centering
  \caption{\revise{Variants comparison on the three benchmark datasets under the three o.o.d data splits. Bold and underline indicate the best Macro F1 and \SecondRevise{Accuracy} among the three variants in each row.}}
    \begin{tabular}{cccccccc}
    \toprule
    \multirow{2}[4]{*}{Dataset} & \multirow{2}[4]{*}{OOD Type} & \multicolumn{2}{c}{HG-SCM} & \multicolumn{2}{c}{HG-SCM-TC} & \multicolumn{2}{c}{HG-SCM-ST} \\
\cmidrule{3-8}          &       & Macro F1 & \SecondRevise{Accuracy} & Macro F1 & \SecondRevise{Accuracy} & Macro F1 & \SecondRevise{Accuracy} \\
    \midrule
    ACM   & By Homophily & 97.10±1.12 & 97.32±1.06 & \underline{\textbf{97.19±0.33}} & \underline{\textbf{97.44±0.31}} & 96.97±0.80 & 97.22±0.76 \\
    ACM   & By Degree & \underline{\textbf{94.24±0.30}} & \underline{\textbf{94.94±0.27}} & 94.06±0.10 & 94.78±0.09 & 94.10±0.44 & 94.81±0.37 \\
    ACM   & By Feature & \underline{\textbf{94.18±0.70}} & \underline{\textbf{94.11±0.67}} & 92.73±0.33 & 92.69±0.28 & 94.15±1.02 & 94.09±1.01 \\
    \midrule
    DBLP  & By Homophily & 89.35±0.75 & 89.63±0.75 & 85.18±0.40 & 85.70±0.42 & \underline{\textbf{89.92±0.72}} & \underline{\textbf{90.13±0.68}} \\
    DBLP  & By Degree & \underline{\textbf{90.11±0.70}} & \underline{\textbf{90.85±0.60}} & 84.26±1.42 & 85.58±1.35 & 89.11±0.98 & 90.03±0.89 \\
    DBLP  & By Feature & 87.69±5.39 & 90.48±3.98 & 82.24±5.25 & 86.04±3.85 & \underline{\textbf{88.44±3.10}} & \underline{\textbf{90.95±2.62}} \\
    \midrule
    IMDB  & By Homophily & 65.26±0.67 & 66.17±0.66 & 64.95±0.55 & 65.67±0.56 & \underline{\textbf{65.31±0.73}} & \underline{\textbf{66.45±0.58}} \\
    IMDB  & By Degree & 43.51±2.61 & 52.37±1.63 & \underline{\textbf{43.70±2.74}} & 51.50±2.60 & 41.82±5.53 & \underline{\textbf{54.09±5.90}} \\
    IMDB  & By Feature & 61.03±5.14 & 66.64±2.58 & \underline{\textbf{63.25±1.99}} & 66.86±4.28 & 63.19±3.71 & \underline{\textbf{67.46±3.60}} \\
    \midrule
    \multicolumn{2}{c}{Average} & 80.27  & 82.50  & 78.62  & 80.70  & \underline{\textbf{80.33}}  & \underline{\textbf{82.80}}  \\
    \bottomrule
    \end{tabular}%
  \label{tab:variant}%
\end{table}%

%% file: content/conclusion.tex
\section{Implications}\label{sec:implications}
\subsection{Theoretical implications}\label{sec:theoretical-implications}
\emphasize{From the theoretical perspective, this work analyzes and breaks through the limitations on the inference flow in current mainstream paradigms of heterogeneous graph neural networks and offers a new paradigm for heterogeneous graph learning that can dynamically learn causality-based inference flow for distinct tasks.} As our discussion with the instructive example and theoretical analysis in Section~\ref{sec:intro}, under the current mainstream paradigm, the generalizability of the model is compromised because the inference flow naturally introduces spurious correlations and the interpretability of the model is limited because the inference flow is fixed. In contrast, the proposed methodology aligns well with human perception and decision-making processes and thus has the potential to achieve more satisfactory generalizability and interpretability than previous studies.  Unlike previous studies that focus on the sophisticated design of sample-level aggregation and fusion modules, we highlight the importance of reflecting on and optimizing the heterogeneous graph learning paradigm. It is a more fundamental and critical issue to explore a paradigm that is causal and aligns with human knowledge or social theory. 

\subsection{Practical implications}\label{sec:practical-implications}
\emphasize{This work has significant practical implications for real-world scenarios, including finance, policy, and healthcare, where transparency and trustworthiness are critical.} Prior research often compromises task performance for better generalizability and their approaches to achieve generalizability can not enable human understanding. In contrast, HG-SCM achieves the best task performance and optimal generalizability according to the experimental results. The generalizability constrained by the causal mechanism is highly trustworthy because it aligns with human cognition and causal mechanisms are stable across environments. In addition, prior work can only provide explanations at the sample level, such as which subgraph of a sample influences the prediction. Such explanations are unreliable and difficult for humans to comprehend. In contrast, HG-SCM has the capacity to provide in-depth interpretations in accordance with the learning tasks by generating the learned causal relationships among expressive semantics involved in the heterogeneous graph. This type of interpretability aligns well with human cognitive habits, enabling researchers and developers to gain a deeper understanding of the phenomenon or task and reflect on the acceptability and reasonableness of the logic behind the phenomenon.

\section{Conclusion}\label{sec:conclusion}

\textit{``What will humans do when facing a prediction task?''} Facing this challenge, in this work, we proposed a novel heterogeneous graph model, HG-SCM (Heterogeneous Graph as Structural Causal Model). HG-SCM aligns with human cognition and learns reasoning logic at the task level by leveraging causal discovery and inference techniques. Through extensive experiments on multiple datasets with i.i.d and o.o.d settings, we found that HG-SCM earns task performance superiority while accomplishing inherent interpretability and enhanced generalizability. The DAGs learned by HG-SCM perceive strong interpretations of the learning tasks. In the future, we will further enhance the model's generalizability by incorporating advanced causal techniques. \SecondRevise{Furthermore, to improve the comprehensiveness and persuasiveness of the assessments regarding model interpretability, we will undertake additional quantitative evaluations with human assistance. }

%% file: content/appx.tex
\section{Supplementary Learning Level Experiment}\label{appx:learning-level}
As shown in Table~\ref{tab:motivation}, there is no significant performance difference on test sets when using the average of estimated values of the self-attention matrices of training samples, instead of dynamic sample-wise self-attention in the transformer-based semantics fusion module in SeHGNN~\citep{sehgnn}. These empirical results suggest that, compared with conventional sample-wise learning modules, an advanced task-level semantic fusion module may have more potential. 
\input{tables/motivation.tex}

%% file: tables/motivation.tex
\begin{table}[htbp]
  \centering
  \caption{
    The effect of using the average of estimated values of the self-attention matrices of training samples (global-mean), instead of dynamic sample-wise self-attention in the transformer-based semantic fusion module in SeHGNN~\citep{sehgnn} on test sets.
  }
  \label{tab:motivation}%
    \begin{threeparttable}
    \setlength{\tabcolsep}{4pt}
  \renewcommand{\arraystretch}{1.}
    \begin{tabular}{ccccccc}
    \toprule
          & \multicolumn{2}{c}{DBLP} & \multicolumn{2}{c}{ACM} & \multicolumn{2}{c}{IMDB}\\
    \midrule
    Metric & Micro F1 & Macro F1 & Micro F1 & Macro F1 & Micro F1 & Macro F1\\
    \midrule
    Sample-wise & 94.97±0.33 & 94.62±0.36 & 93.81±0.41 & 93.86±0.40 & 66.06±0.12 & 64.68±0.24\\
    Global-mean & 94.97±0.43 & 94.62±0.45 & 93.82±0.42 & 93.89±0.41 & 66.05±0.12 & 64.67±0.24\\
    \bottomrule
    \end{tabular}%

  \end{threeparttable}
\end{table}%

%% file: cas-sc-template.bbl
\begin{thebibliography}{118}
\expandafter\ifx\csname natexlab\endcsname\relax\def\natexlab#1{#1}\fi
\providecommand{\url}[1]{\texttt{#1}}
\providecommand{\href}[2]{#2}
\providecommand{\path}[1]{#1}
\providecommand{\DOIprefix}{doi:}
\providecommand{\ArXivprefix}{arXiv:}
\providecommand{\URLprefix}{URL: }
\providecommand{\Pubmedprefix}{pmid:}
\providecommand{\doi}[1]{\href{http://dx.doi.org/#1}{\path{#1}}}
\providecommand{\Pubmed}[1]{\href{pmid:#1}{\path{#1}}}
\providecommand{\bibinfo}[2]{#2}
\ifx\xfnm\relax \def\xfnm[#1]{\unskip,\space#1}\fi
\bibitem[{Abramo et~al.(2018)Abramo, D’Angelo and Zhang}]{joi_interdisciplinary_author_or_reference}
\bibinfo{author}{Abramo, G.}, \bibinfo{author}{D’Angelo, C.A.}, \bibinfo{author}{Zhang, L.}, \bibinfo{year}{2018}.
\newblock \bibinfo{title}{A comparison of two approaches for measuring interdisciplinary research output: The disciplinary diversity of authors vs the disciplinary diversity of the reference list}.
\newblock \bibinfo{journal}{Journal of Informetrics} \bibinfo{volume}{12}, \bibinfo{pages}{1182--1193}.
\newblock \URLprefix \url{https://www.sciencedirect.com/science/article/pii/S1751157718301639}, \DOIprefix\doi{https://doi.org/10.1016/j.joi.2018.09.001}.
\bibitem[{Alan(2012)}]{2012FieldExperiments}
\bibinfo{author}{Alan, G.}, \bibinfo{year}{2012}.
\newblock \bibinfo{title}{Field experiments: Design, analysis, and interpretation}.
\newblock \bibinfo{journal}{W.W. Norton} .
\bibitem[{Amon and Hornik(2022)}]{joi_linguistic_and_journal}
\bibinfo{author}{Amon, J.}, \bibinfo{author}{Hornik, K.}, \bibinfo{year}{2022}.
\newblock \bibinfo{title}{Is it all bafflegab? – linguistic and meta characteristics of research articles in prestigious economics journals}.
\newblock \bibinfo{journal}{Journal of Informetrics} \bibinfo{volume}{16}, \bibinfo{pages}{101284}.
\newblock \URLprefix \url{https://www.sciencedirect.com/science/article/pii/S1751157722000360}, \DOIprefix\doi{https://doi.org/10.1016/j.joi.2022.101284}.
\bibitem[{Baldassarre and Azizpour(2019)}]{icml_2019_workshop_grad_based}
\bibinfo{author}{Baldassarre, F.}, \bibinfo{author}{Azizpour, H.}, \bibinfo{year}{2019}.
\newblock \bibinfo{title}{Explainability techniques for graph convolutional networks}, in: \bibinfo{booktitle}{International Conference on Machine Learning (ICML) Workshops, 2019 Workshop on Learning and Reasoning with Graph-Structured Representations}.
\bibitem[{Bareinboim et~al.(2022)Bareinboim, Correa, Ibeling and Icard}]{bareinboim2022pearl}
\bibinfo{author}{Bareinboim, E.}, \bibinfo{author}{Correa, J.D.}, \bibinfo{author}{Ibeling, D.}, \bibinfo{author}{Icard, T.}, \bibinfo{year}{2022}.
\newblock \bibinfo{title}{On Pearl’s Hierarchy and the Foundations of Causal Inference}. \bibinfo{edition}{1} ed.. \bibinfo{publisher}{Association for Computing Machinery}, \bibinfo{address}{New York, NY, USA}.
\newblock p. \bibinfo{pages}{507–556}.
\newblock \URLprefix \url{https://doi.org/10.1145/3501714.3501743}.
\bibitem[{Borgatti et~al.(2009)Borgatti, Mehra, Brass and Labianca}]{ego_network_2009}
\bibinfo{author}{Borgatti, S.P.}, \bibinfo{author}{Mehra, A.}, \bibinfo{author}{Brass, D.J.}, \bibinfo{author}{Labianca, G.}, \bibinfo{year}{2009}.
\newblock \bibinfo{title}{Network analysis in the social sciences}.
\newblock \bibinfo{journal}{science} \bibinfo{volume}{323}, \bibinfo{pages}{892--895}.
\bibitem[{Chairatanakul et~al.(2021)Chairatanakul, Liu and Murata}]{PGRA}
\bibinfo{author}{Chairatanakul, N.}, \bibinfo{author}{Liu, X.}, \bibinfo{author}{Murata, T.}, \bibinfo{year}{2021}.
\newblock \bibinfo{title}{Pgra: Projected graph relation-feature attention network for heterogeneous information network embedding}.
\newblock \bibinfo{journal}{INFORMATION SCIENCES} \bibinfo{volume}{570}, \bibinfo{pages}{769--794}.
\newblock \DOIprefix\doi{10.1016/j.ins.2021.04.070}.
\bibitem[{Chang et~al.(2023)Chang, Zhou, Cai, Fan, Hu and Wen}]{ipm2023knowledgeGNN}
\bibinfo{author}{Chang, Y.}, \bibinfo{author}{Zhou, W.}, \bibinfo{author}{Cai, H.}, \bibinfo{author}{Fan, W.}, \bibinfo{author}{Hu, L.}, \bibinfo{author}{Wen, J.}, \bibinfo{year}{2023}.
\newblock \bibinfo{title}{Meta-relation assisted knowledge-aware coupled graph neural network for recommendation}.
\newblock \bibinfo{journal}{Information Processing \& Management} \bibinfo{volume}{60}, \bibinfo{pages}{103353}.
\newblock \URLprefix \url{https://www.sciencedirect.com/science/article/pii/S0306457323000900}, \DOIprefix\doi{https://doi.org/10.1016/j.ipm.2023.103353}.
\bibitem[{Charpentier et~al.(2022)Charpentier, Kibler and G{\"{u}}nnemann}]{dag_sampling}
\bibinfo{author}{Charpentier, B.}, \bibinfo{author}{Kibler, S.}, \bibinfo{author}{G{\"{u}}nnemann, S.}, \bibinfo{year}{2022}.
\newblock \bibinfo{title}{Differentiable {DAG} sampling}, in: \bibinfo{booktitle}{The Tenth International Conference on Learning Representations, {ICLR} 2022, Virtual Event, April 25-29, 2022}, \bibinfo{publisher}{OpenReview.net}.
\newblock \URLprefix \url{https://openreview.net/forum?id=9wOQOgNe-w}.
\bibitem[{Chen et~al.(2021)Chen, Cai, Hu, Chen and Chen}]{ipm2023FactVerification}
\bibinfo{author}{Chen, C.}, \bibinfo{author}{Cai, F.}, \bibinfo{author}{Hu, X.}, \bibinfo{author}{Chen, W.}, \bibinfo{author}{Chen, H.}, \bibinfo{year}{2021}.
\newblock \bibinfo{title}{Hhgn: A hierarchical reasoning-based heterogeneous graph neural network for fact verification}.
\newblock \bibinfo{journal}{Information Processing \& Management} \bibinfo{volume}{58}, \bibinfo{pages}{102659}.
\newblock \URLprefix \url{https://www.sciencedirect.com/science/article/pii/S0306457321001473}, \DOIprefix\doi{https://doi.org/10.1016/j.ipm.2021.102659}.
\bibitem[{Christofides(1975)}]{dag1975}
\bibinfo{author}{Christofides, N.}, \bibinfo{year}{1975}.
\newblock \bibinfo{title}{Graph theory: An algorithmic approach (Computer science and applied mathematics)}.
\newblock \bibinfo{publisher}{Academic Press, Inc.}
\bibitem[{Cranmer et~al.(2020)Cranmer, Sanchez~Gonzalez, Battaglia, Xu, Cranmer, Spergel and Ho}]{cranmer2020discovering}
\bibinfo{author}{Cranmer, M.}, \bibinfo{author}{Sanchez~Gonzalez, A.}, \bibinfo{author}{Battaglia, P.}, \bibinfo{author}{Xu, R.}, \bibinfo{author}{Cranmer, K.}, \bibinfo{author}{Spergel, D.}, \bibinfo{author}{Ho, S.}, \bibinfo{year}{2020}.
\newblock \bibinfo{title}{Discovering symbolic models from deep learning with inductive biases}.
\newblock \bibinfo{journal}{Advances in Neural Information Processing Systems} \bibinfo{volume}{33}, \bibinfo{pages}{17429--17442}.
\bibitem[{Cui and Athey(2022)}]{nature_machine_intelligence_stable_learning}
\bibinfo{author}{Cui, P.}, \bibinfo{author}{Athey, S.}, \bibinfo{year}{2022}.
\newblock \bibinfo{title}{Stable learning establishes some common ground between causal inference and machine learning}.
\newblock \bibinfo{journal}{Nat. Mach. Intell.} \bibinfo{volume}{4}, \bibinfo{pages}{110--115}.
\newblock \URLprefix \url{https://doi.org/10.1038/s42256-022-00445-z}, \DOIprefix\doi{10.1038/S42256-022-00445-Z}.
\bibitem[{Cundy et~al.(2021)Cundy, Grover and Ermon}]{NPS2021_BCD}
\bibinfo{author}{Cundy, C.}, \bibinfo{author}{Grover, A.}, \bibinfo{author}{Ermon, S.}, \bibinfo{year}{2021}.
\newblock \bibinfo{title}{Bcd nets: Scalable variational approaches for bayesian causal discovery}, in: \bibinfo{editor}{Ranzato, M.}, \bibinfo{editor}{Beygelzimer, A.}, \bibinfo{editor}{Dauphin, Y.}, \bibinfo{editor}{Liang, P.}, \bibinfo{editor}{Vaughan, J.W.} (Eds.), \bibinfo{booktitle}{Advances in Neural Information Processing Systems}, \bibinfo{publisher}{Curran Associates, Inc.}. pp. \bibinfo{pages}{7095--7110}.
\newblock \URLprefix \url{https://proceedings.neurips.cc/paper/2021/file/39799c18791e8d7eb29704fc5bc04ac8-Paper.pdf}.
\bibitem[{Dai et~al.(2023)Dai, Zhao, Li, Tian, Zhao and Pan}]{IPM_Bert_HGNN_citation_rec}
\bibinfo{author}{Dai, T.}, \bibinfo{author}{Zhao, J.}, \bibinfo{author}{Li, D.}, \bibinfo{author}{Tian, S.}, \bibinfo{author}{Zhao, X.}, \bibinfo{author}{Pan, S.}, \bibinfo{year}{2023}.
\newblock \bibinfo{title}{Heterogeneous deep graph convolutional network with citation relational bert for covid-19 inline citation recommendation}.
\newblock \bibinfo{journal}{EXPERT SYSTEMS WITH APPLICATIONS} \bibinfo{volume}{213}.
\newblock \DOIprefix\doi{10.1016/j.eswa.2022.118841}.
\bibitem[{Daly and Haahr(2007)}]{ego_network_2007}
\bibinfo{author}{Daly, E.M.}, \bibinfo{author}{Haahr, M.}, \bibinfo{year}{2007}.
\newblock \bibinfo{title}{Social network analysis for routing in disconnected delay-tolerant manets}, in: \bibinfo{booktitle}{Proceedings of the 8th ACM international symposium on Mobile ad hoc networking and computing}, pp. \bibinfo{pages}{32--40}.
\bibitem[{Dong et~al.(2017)Dong, Chawla and Swami}]{dong2017metapath2vec}
\bibinfo{author}{Dong, Y.}, \bibinfo{author}{Chawla, N.V.}, \bibinfo{author}{Swami, A.}, \bibinfo{year}{2017}.
\newblock \bibinfo{title}{metapath2vec: Scalable representation learning for heterogeneous networks}, in: \bibinfo{booktitle}{Proceedings of the 23rd ACM SIGKDD International Conference on Knowledge Discovery and Data Mining}, \bibinfo{organization}{ACM}. pp. \bibinfo{pages}{135--144}.
\bibitem[{Durmusoglu and Durmusoglu(2021)}]{sciento_journal_selection}
\bibinfo{author}{Durmusoglu, Z.D.U.}, \bibinfo{author}{Durmusoglu, A.}, \bibinfo{year}{2021}.
\newblock \bibinfo{title}{A {TOPSIS} model for understanding the authors choice of journal selection}.
\newblock \bibinfo{journal}{Scientometrics} \bibinfo{volume}{126}, \bibinfo{pages}{521--543}.
\newblock \URLprefix \url{https://doi.org/10.1007/s11192-020-03770-5}, \DOIprefix\doi{10.1007/S11192-020-03770-5}.
\bibitem[{Fan et~al.(2022)Fan, Wang, Shi, Kuang, Liu and Wang}]{tnnls_debiased}
\bibinfo{author}{Fan, S.}, \bibinfo{author}{Wang, X.}, \bibinfo{author}{Shi, C.}, \bibinfo{author}{Kuang, K.}, \bibinfo{author}{Liu, N.}, \bibinfo{author}{Wang, B.}, \bibinfo{year}{2022}.
\newblock \bibinfo{title}{Debiased graph neural networks with agnostic label selection bias}.
\newblock \bibinfo{journal}{IEEE Transactions on Neural Networks and Learning Systems} , \bibinfo{pages}{1--12}\DOIprefix\doi{10.1109/TNNLS.2022.3141260}.
\bibitem[{Fang et~al.(2022)Fang, Liu, Geng, Zhu and He}]{AI_Markov_dag}
\bibinfo{author}{Fang, Z.}, \bibinfo{author}{Liu, Y.}, \bibinfo{author}{Geng, Z.}, \bibinfo{author}{Zhu, S.}, \bibinfo{author}{He, Y.}, \bibinfo{year}{2022}.
\newblock \bibinfo{title}{A local method for identifying causal relations under markov}.
\newblock \bibinfo{journal}{ARTIFICIAL INTELLIGENCE} \bibinfo{volume}{305}.
\newblock \DOIprefix\doi{10.1016/j.artint.2022.103669}.
\bibitem[{Fang et~al.(2023)Fang, Zhu, Zhang, Liu, Chen and He}]{tnnls_Low_rank_DAG}
\bibinfo{author}{Fang, Z.}, \bibinfo{author}{Zhu, S.}, \bibinfo{author}{Zhang, J.}, \bibinfo{author}{Liu, Y.}, \bibinfo{author}{Chen, Z.}, \bibinfo{author}{He, Y.}, \bibinfo{year}{2023}.
\newblock \bibinfo{title}{On low-rank directed acyclic graphs and causal structure learning}.
\newblock \bibinfo{journal}{IEEE Transactions on Neural Networks and Learning Systems} , \bibinfo{pages}{1--14}\DOIprefix\doi{10.1109/TNNLS.2023.3273353}.
\bibitem[{Feng et~al.(2021)Feng, He, Tang and Chua}]{tkde_adversarial_train}
\bibinfo{author}{Feng, F.}, \bibinfo{author}{He, X.}, \bibinfo{author}{Tang, J.}, \bibinfo{author}{Chua, T.}, \bibinfo{year}{2021}.
\newblock \bibinfo{title}{Graph adversarial training: Dynamically regularizing based on graph structure}.
\newblock \bibinfo{journal}{{IEEE} Trans. Knowl. Data Eng.} \bibinfo{volume}{33}, \bibinfo{pages}{2493--2504}.
\newblock \URLprefix \url{https://doi.org/10.1109/TKDE.2019.2957786}, \DOIprefix\doi{10.1109/TKDE.2019.2957786}.
\bibitem[{Fu et~al.(2020)Fu, Zhang, Meng and King}]{magnn}
\bibinfo{author}{Fu, X.}, \bibinfo{author}{Zhang, J.}, \bibinfo{author}{Meng, Z.}, \bibinfo{author}{King, I.}, \bibinfo{year}{2020}.
\newblock \bibinfo{title}{{MAGNN:} metapath aggregated graph neural network for heterogeneous graph embedding}, in: \bibinfo{editor}{Huang, Y.}, \bibinfo{editor}{King, I.}, \bibinfo{editor}{Liu, T.}, \bibinfo{editor}{van Steen, M.} (Eds.), \bibinfo{booktitle}{{WWW} '20: The Web Conference 2020, Taipei, Taiwan, April 20-24, 2020}, \bibinfo{publisher}{{ACM} / {IW3C2}}. pp. \bibinfo{pages}{2331--2341}.
\newblock \URLprefix \url{https://doi.org/10.1145/3366423.3380297}, \DOIprefix\doi{10.1145/3366423.3380297}.
\bibitem[{Fu et~al.(2023)Fu, Yu, Wu, Ding and Zhao}]{robust_heterogeneous_embedding}
\bibinfo{author}{Fu, Y.}, \bibinfo{author}{Yu, X.}, \bibinfo{author}{Wu, Y.}, \bibinfo{author}{Ding, X.}, \bibinfo{author}{Zhao, S.}, \bibinfo{year}{2023}.
\newblock \bibinfo{title}{Robust representation learning for heterogeneous attributed networks}.
\newblock \bibinfo{journal}{INFORMATION SCIENCES} \bibinfo{volume}{628}, \bibinfo{pages}{22--49}.
\newblock \DOIprefix\doi{10.1016/j.ins.2023.01.038}.
\bibitem[{Gamella and Heinze-Deml(2020)}]{active_nips20}
\bibinfo{author}{Gamella, J.L.}, \bibinfo{author}{Heinze-Deml, C.}, \bibinfo{year}{2020}.
\newblock \bibinfo{title}{Active invariant causal prediction: Experiment selection through stability}, in: \bibinfo{editor}{Larochelle, H.}, \bibinfo{editor}{Ranzato, M.}, \bibinfo{editor}{Hadsell, R.}, \bibinfo{editor}{Balcan, M.}, \bibinfo{editor}{Lin, H.} (Eds.), \bibinfo{booktitle}{Advances in Neural Information Processing Systems}, \bibinfo{publisher}{Curran Associates, Inc.}. pp. \bibinfo{pages}{15464--15475}.
\newblock \URLprefix \url{https://proceedings.neurips.cc/paper/2020/file/b197ffdef2ddc3308584dce7afa3661b-Paper.pdf}.
\bibitem[{Geirhos et~al.(2020)Geirhos, Jacobsen, Michaelis, Zemel, Brendel, Bethge and Wichmann}]{nature2020shortcut}
\bibinfo{author}{Geirhos, R.}, \bibinfo{author}{Jacobsen, J.H.}, \bibinfo{author}{Michaelis, C.}, \bibinfo{author}{Zemel, R.}, \bibinfo{author}{Brendel, W.}, \bibinfo{author}{Bethge, M.}, \bibinfo{author}{Wichmann, F.A.}, \bibinfo{year}{2020}.
\newblock \bibinfo{title}{Shortcut learning in deep neural networks}.
\newblock \bibinfo{journal}{Nature Machine Intelligence} \bibinfo{volume}{2}, \bibinfo{pages}{665--673}.
\bibitem[{Hamilton et~al.(2017)Hamilton, Ying and Leskovec}]{graphsage}
\bibinfo{author}{Hamilton, W.L.}, \bibinfo{author}{Ying, R.}, \bibinfo{author}{Leskovec, J.}, \bibinfo{year}{2017}.
\newblock \bibinfo{title}{Inductive representation learning on large graphs}, in: \bibinfo{booktitle}{Proceedings of the 31st International Conference on Neural Information Processing Systems}, \bibinfo{publisher}{Curran Associates Inc.}, \bibinfo{address}{Red Hook, NY, USA}. p. \bibinfo{pages}{1025–1035}.
\bibitem[{Han et~al.(2022a)Han, Zhao, Yang, Zhang, Liu, Wang and Shi}]{openhgnn}
\bibinfo{author}{Han, H.}, \bibinfo{author}{Zhao, T.}, \bibinfo{author}{Yang, C.}, \bibinfo{author}{Zhang, H.}, \bibinfo{author}{Liu, Y.}, \bibinfo{author}{Wang, X.}, \bibinfo{author}{Shi, C.}, \bibinfo{year}{2022}a.
\newblock \bibinfo{title}{Openhgnn: An open source toolkit for heterogeneous graph neural network}, in: \bibinfo{booktitle}{Proceedings of the 31st ACM International Conference on Information \& Knowledge Management}, \bibinfo{publisher}{Association for Computing Machinery}, \bibinfo{address}{New York, NY, USA}. p. \bibinfo{pages}{3993–3997}.
\newblock \URLprefix \url{https://doi.org/10.1145/3511808.3557664}, \DOIprefix\doi{10.1145/3511808.3557664}.
\bibitem[{Han et~al.(2022b)Han, Zhao, Yang, Zhang, Liu, Wang and Shi}]{han2022openhgnn}
\bibinfo{author}{Han, H.}, \bibinfo{author}{Zhao, T.}, \bibinfo{author}{Yang, C.}, \bibinfo{author}{Zhang, H.}, \bibinfo{author}{Liu, Y.}, \bibinfo{author}{Wang, X.}, \bibinfo{author}{Shi, C.}, \bibinfo{year}{2022}b.
\newblock \bibinfo{title}{Openhgnn: An open source toolkit for heterogeneous graph neural network}, in: \bibinfo{booktitle}{CIKM}.
\bibitem[{Hara et~al.(2003)Hara, Solomon, Kim and Sonnenwald}]{jasist_factors_collaboration}
\bibinfo{author}{Hara, N.}, \bibinfo{author}{Solomon, P.}, \bibinfo{author}{Kim, S.L.}, \bibinfo{author}{Sonnenwald, D.H.}, \bibinfo{year}{2003}.
\newblock \bibinfo{title}{An emerging view of scientific collaboration: Scientists' perspectives on collaboration and factors that impact collaboration}.
\newblock \bibinfo{journal}{Journal of the American Society for Information Science and Technology} \bibinfo{volume}{54}, \bibinfo{pages}{952--965}.
\newblock \URLprefix \url{https://onlinelibrary.wiley.com/doi/abs/10.1002/asi.10291}, \DOIprefix\doi{https://doi.org/10.1002/asi.10291}, \href{http://arxiv.org/abs/https://onlinelibrary.wiley.com/doi/pdf/10.1002/asi.10291}{\tt arXiv:https://onlinelibrary.wiley.com/doi/pdf/10.1002/asi.10291}.
\bibitem[{He et~al.(2022)He, Wang, Cui, Zou, Zhang, Cui and Jiang}]{CausPref}
\bibinfo{author}{He, Y.}, \bibinfo{author}{Wang, Z.}, \bibinfo{author}{Cui, P.}, \bibinfo{author}{Zou, H.}, \bibinfo{author}{Zhang, Y.}, \bibinfo{author}{Cui, Q.}, \bibinfo{author}{Jiang, Y.}, \bibinfo{year}{2022}.
\newblock \bibinfo{title}{Causpref: Causal preference learning for out-of-distribution recommendation}, in: \bibinfo{booktitle}{Proceedings of the ACM Web Conference 2022}, \bibinfo{publisher}{Association for Computing Machinery}, \bibinfo{address}{New York, NY, USA}. p. \bibinfo{pages}{410–421}.
\newblock \URLprefix \url{https://doi.org/10.1145/3485447.3511969}, \DOIprefix\doi{10.1145/3485447.3511969}.
\bibitem[{Hong et~al.(2020)Hong, Guo, Lin, Yang, Li and Ye}]{HetSANN}
\bibinfo{author}{Hong, H.}, \bibinfo{author}{Guo, H.}, \bibinfo{author}{Lin, Y.}, \bibinfo{author}{Yang, X.}, \bibinfo{author}{Li, Z.}, \bibinfo{author}{Ye, J.}, \bibinfo{year}{2020}.
\newblock \bibinfo{title}{An attention-based graph neural network for heterogeneous structural learning}, in: \bibinfo{booktitle}{The Thirty-Fourth {AAAI} Conference on Artificial Intelligence, {AAAI} 2020, The Thirty-Second Innovative Applications of Artificial Intelligence Conference, {IAAI} 2020, The Tenth {AAAI} Symposium on Educational Advances in Artificial Intelligence, {EAAI} 2020, New York, NY, USA, February 7-12, 2020}, \bibinfo{publisher}{{AAAI} Press}. pp. \bibinfo{pages}{4132--4139}.
\newblock \URLprefix \url{https://doi.org/10.1609/aaai.v34i04.5833}, \DOIprefix\doi{10.1609/aaai.v34i04.5833}.
\bibitem[{Hornik et~al.(1989)Hornik, Stinchcombe and White}]{mlp}
\bibinfo{author}{Hornik, K.}, \bibinfo{author}{Stinchcombe, M.B.}, \bibinfo{author}{White, H.}, \bibinfo{year}{1989}.
\newblock \bibinfo{title}{Multilayer feedforward networks are universal approximators}.
\newblock \bibinfo{journal}{Neural Networks} \bibinfo{volume}{2}, \bibinfo{pages}{359--366}.
\newblock \URLprefix \url{https://doi.org/10.1016/0893-6080(89)90020-8}, \DOIprefix\doi{10.1016/0893-6080(89)90020-8}.
\bibitem[{Hsieh(2017)}]{jasist_author_jounral_preference}
\bibinfo{author}{Hsieh, J.L.}, \bibinfo{year}{2017}.
\newblock \bibinfo{title}{Author publication preferences and journal competition}.
\newblock \bibinfo{journal}{Journal of the Association for Information Science and Technology} \bibinfo{volume}{68}, \bibinfo{pages}{365--377}.
\newblock \URLprefix \url{https://asistdl.onlinelibrary.wiley.com/doi/abs/10.1002/asi.23657}, \DOIprefix\doi{https://doi.org/10.1002/asi.23657}, \href{http://arxiv.org/abs/https://asistdl.onlinelibrary.wiley.com/doi/pdf/10.1002/asi.23657}{\tt arXiv:https://asistdl.onlinelibrary.wiley.com/doi/pdf/10.1002/asi.23657}.
\bibitem[{Hu et~al.(2020)Hu, Dong, Wang and Sun}]{hgt}
\bibinfo{author}{Hu, Z.}, \bibinfo{author}{Dong, Y.}, \bibinfo{author}{Wang, K.}, \bibinfo{author}{Sun, Y.}, \bibinfo{year}{2020}.
\newblock \bibinfo{title}{Heterogeneous graph transformer}, in: \bibinfo{booktitle}{Proceedings of The Web Conference 2020}, pp. \bibinfo{pages}{2704--2710}.
\bibitem[{Huang et~al.(2023)Huang, Yamada, Tian, Singh and Chang}]{GraphLime}
\bibinfo{author}{Huang, Q.}, \bibinfo{author}{Yamada, M.}, \bibinfo{author}{Tian, Y.}, \bibinfo{author}{Singh, D.}, \bibinfo{author}{Chang, Y.}, \bibinfo{year}{2023}.
\newblock \bibinfo{title}{Graphlime: Local interpretable model explanations for graph neural networks}.
\newblock \bibinfo{journal}{{IEEE} Trans. Knowl. Data Eng.} \bibinfo{volume}{35}, \bibinfo{pages}{6968--6972}.
\newblock \URLprefix \url{https://doi.org/10.1109/TKDE.2022.3187455}, \DOIprefix\doi{10.1109/TKDE.2022.3187455}.
\bibitem[{Jiang et~al.(2020)Jiang, Gao, Lan, Yang, Lu and Liu}]{jiang2020task}
\bibinfo{author}{Jiang, Z.}, \bibinfo{author}{Gao, Z.}, \bibinfo{author}{Lan, J.}, \bibinfo{author}{Yang, H.}, \bibinfo{author}{Lu, Y.}, \bibinfo{author}{Liu, X.}, \bibinfo{year}{2020}.
\newblock \bibinfo{title}{Task-oriented genetic activation for large-scale complex heterogeneous graph embedding}, in: \bibinfo{booktitle}{Proceedings of The Web Conference 2020}, pp. \bibinfo{pages}{1581--1591}.
\bibitem[{Jiang et~al.(2018)Jiang, Yin, Gao, Lu and Liu}]{jiang2018cross}
\bibinfo{author}{Jiang, Z.}, \bibinfo{author}{Yin, Y.}, \bibinfo{author}{Gao, L.}, \bibinfo{author}{Lu, Y.}, \bibinfo{author}{Liu, X.}, \bibinfo{year}{2018}.
\newblock \bibinfo{title}{Cross-language citation recommendation via hierarchical representation learning on heterogeneous graph}, in: \bibinfo{booktitle}{The 41st International ACM SIGIR Conference on Research \& Development in Information Retrieval}, pp. \bibinfo{pages}{635--644}.
\bibitem[{Kalainathan et~al.(2022)Kalainathan, Goudet, Guyon, Lopez-Paz and Sebag}]{jmlr_Adversarial_dag}
\bibinfo{author}{Kalainathan, D.}, \bibinfo{author}{Goudet, O.}, \bibinfo{author}{Guyon, I.G.}, \bibinfo{author}{Lopez-Paz, D.}, \bibinfo{author}{Sebag, M.}, \bibinfo{year}{2022}.
\newblock \bibinfo{title}{Structural agnostic modeling: Adversarial learning of causal graphs}.
\newblock \bibinfo{journal}{JOURNAL OF MACHINE LEARNING RESEARCH} \bibinfo{volume}{23}.
\bibitem[{Kitson et~al.(2023)Kitson, Constantinou, Guo, Liu and Chobtham}]{AI_review_survey_dag}
\bibinfo{author}{Kitson, N.K.}, \bibinfo{author}{Constantinou, A.C.C.}, \bibinfo{author}{Guo, Z.}, \bibinfo{author}{Liu, Y.}, \bibinfo{author}{Chobtham, K.}, \bibinfo{year}{2023}.
\newblock \bibinfo{title}{A survey of bayesian network structure learning}.
\newblock \bibinfo{journal}{ARTIFICIAL INTELLIGENCE REVIEW} \bibinfo{volume}{56}, \bibinfo{pages}{8721--8814}.
\newblock \DOIprefix\doi{10.1007/s10462-022-10351-w}.
\bibitem[{Knyazev et~al.(2019)Knyazev, Taylor and Amer}]{knyazev2019understanding}
\bibinfo{author}{Knyazev, B.}, \bibinfo{author}{Taylor, G.W.}, \bibinfo{author}{Amer, M.}, \bibinfo{year}{2019}.
\newblock \bibinfo{title}{Understanding attention and generalization in graph neural networks}.
\newblock \bibinfo{journal}{Advances in neural information processing systems} \bibinfo{volume}{32}.
\bibitem[{Kong et~al.(2022)Kong, Li, Ding, Wu, Zhu, Ghanem, Taylor and Goldstein}]{cvpr_data_aug}
\bibinfo{author}{Kong, K.}, \bibinfo{author}{Li, G.}, \bibinfo{author}{Ding, M.}, \bibinfo{author}{Wu, Z.}, \bibinfo{author}{Zhu, C.}, \bibinfo{author}{Ghanem, B.}, \bibinfo{author}{Taylor, G.}, \bibinfo{author}{Goldstein, T.}, \bibinfo{year}{2022}.
\newblock \bibinfo{title}{Robust optimization as data augmentation for large-scale graphs}, in: \bibinfo{booktitle}{{IEEE/CVF} Conference on Computer Vision and Pattern Recognition, {CVPR} 2022, New Orleans, LA, USA, June 18-24, 2022}, \bibinfo{publisher}{{IEEE}}. pp. \bibinfo{pages}{60--69}.
\newblock \URLprefix \url{https://doi.org/10.1109/CVPR52688.2022.00016}, \DOIprefix\doi{10.1109/CVPR52688.2022.00016}.
\bibitem[{Kyono et~al.(2020)Kyono, Zhang and van~der Schaar}]{castle}
\bibinfo{author}{Kyono, T.}, \bibinfo{author}{Zhang, Y.}, \bibinfo{author}{van~der Schaar, M.}, \bibinfo{year}{2020}.
\newblock \bibinfo{title}{Castle: Regularization via auxiliary causal graph discovery}, in: \bibinfo{booktitle}{Proceedings of the 34th International Conference on Neural Information Processing Systems}, \bibinfo{publisher}{Curran Associates Inc.}, \bibinfo{address}{Red Hook, NY, USA}.
\bibitem[{Lachapelle et~al.(2020)Lachapelle, Brouillard, Deleu and Lacoste-Julien}]{2020GradientDAG}
\bibinfo{author}{Lachapelle, S.}, \bibinfo{author}{Brouillard, P.}, \bibinfo{author}{Deleu, T.}, \bibinfo{author}{Lacoste-Julien, S.}, \bibinfo{year}{2020}.
\newblock \bibinfo{title}{Gradient-based neural dag learning}, in: \bibinfo{booktitle}{International Conference on Learning Representations}.
\bibitem[{Lee et~al.(2019)Lee, Lee, Kim, Kosiorek, Choi and Teh}]{settransformer}
\bibinfo{author}{Lee, J.}, \bibinfo{author}{Lee, Y.}, \bibinfo{author}{Kim, J.}, \bibinfo{author}{Kosiorek, A.}, \bibinfo{author}{Choi, S.}, \bibinfo{author}{Teh, Y.W.}, \bibinfo{year}{2019}.
\newblock \bibinfo{title}{Set transformer: A framework for attention-based permutation-invariant neural networks}, in: \bibinfo{editor}{Chaudhuri, K.}, \bibinfo{editor}{Salakhutdinov, R.} (Eds.), \bibinfo{booktitle}{Proceedings of the 36th International Conference on Machine Learning}, \bibinfo{publisher}{PMLR}. pp. \bibinfo{pages}{3744--3753}.
\newblock \URLprefix \url{https://proceedings.mlr.press/v97/lee19d.html}.
\bibitem[{Li et~al.(2023a)Li, Shen and Pan}]{nonlinear_confounder_dag}
\bibinfo{author}{Li, C.}, \bibinfo{author}{Shen, X.}, \bibinfo{author}{Pan, W.}, \bibinfo{year}{2023}a.
\newblock \bibinfo{title}{Nonlinear causal discovery with confounders}.
\newblock \bibinfo{journal}{Journal of the American Statistical Association} \bibinfo{volume}{0}, \bibinfo{pages}{1--10}.
\newblock \URLprefix \url{https://doi.org/10.1080/01621459.2023.2179490}, \DOIprefix\doi{10.1080/01621459.2023.2179490}, \href{http://arxiv.org/abs/https://doi.org/10.1080/01621459.2023.2179490}{\tt arXiv:https://doi.org/10.1080/01621459.2023.2179490}.
\bibitem[{Li et~al.(2023b)Li, Yan, Fu, Zhao and Zeng}]{HetReGAT-FC}
\bibinfo{author}{Li, C.}, \bibinfo{author}{Yan, Y.}, \bibinfo{author}{Fu, J.}, \bibinfo{author}{Zhao, Z.}, \bibinfo{author}{Zeng, Q.}, \bibinfo{year}{2023}b.
\newblock \bibinfo{title}{Hetregat-fc: Heterogeneous residual graph attention network via feature completion}.
\newblock \bibinfo{journal}{INFORMATION SCIENCES} \bibinfo{volume}{632}, \bibinfo{pages}{424--438}.
\newblock \DOIprefix\doi{10.1016/j.ins.2023.03.034}.
\bibitem[{Li et~al.(2022)Li, Wang, Zhang and Zhu}]{li2022ood}
\bibinfo{author}{Li, H.}, \bibinfo{author}{Wang, X.}, \bibinfo{author}{Zhang, Z.}, \bibinfo{author}{Zhu, W.}, \bibinfo{year}{2022}.
\newblock \bibinfo{title}{Ood-gnn: Out-of-distribution generalized graph neural network}.
\newblock \bibinfo{journal}{IEEE Transactions on Knowledge and Data Engineering} .
\bibitem[{Lin et~al.(2021)Lin, Lan and Li}]{icml_gem}
\bibinfo{author}{Lin, W.}, \bibinfo{author}{Lan, H.}, \bibinfo{author}{Li, B.}, \bibinfo{year}{2021}.
\newblock \bibinfo{title}{Generative causal explanations for graph neural networks}, in: \bibinfo{editor}{Meila, M.}, \bibinfo{editor}{Zhang, T.} (Eds.), \bibinfo{booktitle}{Proceedings of the 38th International Conference on Machine Learning, {ICML} 2021, 18-24 July 2021, Virtual Event}, \bibinfo{publisher}{{PMLR}}. pp. \bibinfo{pages}{6666--6679}.
\newblock \URLprefix \url{http://proceedings.mlr.press/v139/lin21d.html}.
\bibitem[{Liu et~al.(2022)Liu, Hu, Wang, Shi, Zhang and Zhou}]{www_confidence_cheat}
\bibinfo{author}{Liu, H.}, \bibinfo{author}{Hu, B.}, \bibinfo{author}{Wang, X.}, \bibinfo{author}{Shi, C.}, \bibinfo{author}{Zhang, Z.}, \bibinfo{author}{Zhou, J.}, \bibinfo{year}{2022}.
\newblock \bibinfo{title}{Confidence may cheat: Self-training on graph neural networks under distribution shift}, in: \bibinfo{editor}{Laforest, F.}, \bibinfo{editor}{Troncy, R.}, \bibinfo{editor}{Simperl, E.}, \bibinfo{editor}{Agarwal, D.}, \bibinfo{editor}{Gionis, A.}, \bibinfo{editor}{Herman, I.}, \bibinfo{editor}{M{\'{e}}dini, L.} (Eds.), \bibinfo{booktitle}{{WWW} '22: The {ACM} Web Conference 2022, Virtual Event, Lyon, France, April 25 - 29, 2022}, \bibinfo{publisher}{{ACM}}. pp. \bibinfo{pages}{1248--1258}.
\newblock \URLprefix \url{https://doi.org/10.1145/3485447.3512172}, \DOIprefix\doi{10.1145/3485447.3512172}.
\bibitem[{Liu et~al.(2020)Liu, Wang, Wu and Xiao}]{aaai_independence}
\bibinfo{author}{Liu, Y.}, \bibinfo{author}{Wang, X.}, \bibinfo{author}{Wu, S.}, \bibinfo{author}{Xiao, Z.}, \bibinfo{year}{2020}.
\newblock \bibinfo{title}{Independence promoted graph disentangled networks}, in: \bibinfo{booktitle}{The Thirty-Fourth {AAAI} Conference on Artificial Intelligence, {AAAI} 2020, The Thirty-Second Innovative Applications of Artificial Intelligence Conference, {IAAI} 2020, The Tenth {AAAI} Symposium on Educational Advances in Artificial Intelligence, {EAAI} 2020, New York, NY, USA, February 7-12, 2020}, \bibinfo{publisher}{{AAAI} Press}. pp. \bibinfo{pages}{4916--4923}.
\newblock \URLprefix \url{https://doi.org/10.1609/aaai.v34i04.5929}, \DOIprefix\doi{10.1609/AAAI.V34I04.5929}.
\bibitem[{Liu et~al.(2021)Liu, Nguyen and Fang}]{tail-gnn-kdd}
\bibinfo{author}{Liu, Z.}, \bibinfo{author}{Nguyen, T.}, \bibinfo{author}{Fang, Y.}, \bibinfo{year}{2021}.
\newblock \bibinfo{title}{Tail-gnn: Tail-node graph neural networks}, in: \bibinfo{editor}{Zhu, F.}, \bibinfo{editor}{Ooi, B.C.}, \bibinfo{editor}{Miao, C.} (Eds.), \bibinfo{booktitle}{{KDD} '21: The 27th {ACM} {SIGKDD} Conference on Knowledge Discovery and Data Mining, Virtual Event, Singapore, August 14-18, 2021}, \bibinfo{publisher}{{ACM}}. pp. \bibinfo{pages}{1109--1119}.
\newblock \URLprefix \url{https://doi.org/10.1145/3447548.3467276}, \DOIprefix\doi{10.1145/3447548.3467276}.
\bibitem[{Lloyd(1982)}]{kmeans}
\bibinfo{author}{Lloyd, S.}, \bibinfo{year}{1982}.
\newblock \bibinfo{title}{Least squares quantization in pcm}.
\newblock \bibinfo{journal}{IEEE Transactions on Information Theory} \bibinfo{volume}{28}, \bibinfo{pages}{129--137}.
\newblock \DOIprefix\doi{10.1109/TIT.1982.1056489}.
\bibitem[{Loshchilov and Hutter(2018)}]{adamw}
\bibinfo{author}{Loshchilov, I.}, \bibinfo{author}{Hutter, F.}, \bibinfo{year}{2018}.
\newblock \bibinfo{title}{Decoupled weight decay regularization}, in: \bibinfo{booktitle}{International Conference on Learning Representations}.
\bibitem[{Lu et~al.(2022)Lu, Li and Wei}]{ipm2023sentimentanalysis}
\bibinfo{author}{Lu, G.}, \bibinfo{author}{Li, J.}, \bibinfo{author}{Wei, J.}, \bibinfo{year}{2022}.
\newblock \bibinfo{title}{Aspect sentiment analysis with heterogeneous graph neural networks}.
\newblock \bibinfo{journal}{Information Processing \& Management} \bibinfo{volume}{59}, \bibinfo{pages}{102953}.
\newblock \URLprefix \url{https://www.sciencedirect.com/science/article/pii/S0306457322000747}, \DOIprefix\doi{https://doi.org/10.1016/j.ipm.2022.102953}.
\bibitem[{Lu et~al.(2020)Lu, Liu, Huang, Bu, Li and Cheng}]{joi_keyword_selection}
\bibinfo{author}{Lu, W.}, \bibinfo{author}{Liu, Z.}, \bibinfo{author}{Huang, Y.}, \bibinfo{author}{Bu, Y.}, \bibinfo{author}{Li, X.}, \bibinfo{author}{Cheng, Q.}, \bibinfo{year}{2020}.
\newblock \bibinfo{title}{How do authors select keywords? a preliminary study of author keyword selection behavior}.
\newblock \bibinfo{journal}{Journal of Informetrics} \bibinfo{volume}{14}, \bibinfo{pages}{101066}.
\newblock \URLprefix \url{https://www.sciencedirect.com/science/article/pii/S1751157720300134}, \DOIprefix\doi{https://doi.org/10.1016/j.joi.2020.101066}.
\bibitem[{Luo et~al.(2020a)Luo, Cheng, Xu, Yu, Zong, Chen and Zhang}]{nips_PGE}
\bibinfo{author}{Luo, D.}, \bibinfo{author}{Cheng, W.}, \bibinfo{author}{Xu, D.}, \bibinfo{author}{Yu, W.}, \bibinfo{author}{Zong, B.}, \bibinfo{author}{Chen, H.}, \bibinfo{author}{Zhang, X.}, \bibinfo{year}{2020}a.
\newblock \bibinfo{title}{Parameterized explainer for graph neural network}, in: \bibinfo{editor}{Larochelle, H.}, \bibinfo{editor}{Ranzato, M.}, \bibinfo{editor}{Hadsell, R.}, \bibinfo{editor}{Balcan, M.}, \bibinfo{editor}{Lin, H.} (Eds.), \bibinfo{booktitle}{Advances in Neural Information Processing Systems 33: Annual Conference on Neural Information Processing Systems 2020, NeurIPS 2020, December 6-12, 2020, virtual}.
\newblock \URLprefix \url{https://proceedings.neurips.cc/paper/2020/hash/e37b08dd3015330dcbb5d6663667b8b8-Abstract.html}.
\bibitem[{Luo et~al.(2020b)Luo, Peng and Ma}]{nature2020causal}
\bibinfo{author}{Luo, Y.}, \bibinfo{author}{Peng, J.}, \bibinfo{author}{Ma, J.}, \bibinfo{year}{2020}b.
\newblock \bibinfo{title}{When causal inference meets deep learning}.
\newblock \bibinfo{journal}{Nature Machine Intelligence} \bibinfo{volume}{2}, \bibinfo{pages}{426--427}.
\bibitem[{Lv et~al.(2021)Lv, Ding, Liu, Chen, Feng, He, Zhou, Jiang, Dong and Tang}]{simplehgn}
\bibinfo{author}{Lv, Q.}, \bibinfo{author}{Ding, M.}, \bibinfo{author}{Liu, Q.}, \bibinfo{author}{Chen, Y.}, \bibinfo{author}{Feng, W.}, \bibinfo{author}{He, S.}, \bibinfo{author}{Zhou, C.}, \bibinfo{author}{Jiang, J.}, \bibinfo{author}{Dong, Y.}, \bibinfo{author}{Tang, J.}, \bibinfo{year}{2021}.
\newblock \bibinfo{title}{Are we really making much progress? revisiting, benchmarking and refining heterogeneous graph neural networks}, in: \bibinfo{booktitle}{Proceedings of the 27th ACM SIGKDD Conference on Knowledge Discovery; Data Mining}, \bibinfo{publisher}{Association for Computing Machinery}, \bibinfo{address}{New York, NY, USA}. p. \bibinfo{pages}{1150–1160}.
\newblock \URLprefix \url{https://doi.org/10.1145/3447548.3467350}, \DOIprefix\doi{10.1145/3447548.3467350}.
\bibitem[{Ma et~al.(2019)Ma, Cui, Kuang, Wang and Zhu}]{icml_disentangled_gcn}
\bibinfo{author}{Ma, J.}, \bibinfo{author}{Cui, P.}, \bibinfo{author}{Kuang, K.}, \bibinfo{author}{Wang, X.}, \bibinfo{author}{Zhu, W.}, \bibinfo{year}{2019}.
\newblock \bibinfo{title}{Disentangled graph convolutional networks}, in: \bibinfo{editor}{Chaudhuri, K.}, \bibinfo{editor}{Salakhutdinov, R.} (Eds.), \bibinfo{booktitle}{Proceedings of the 36th International Conference on Machine Learning, {ICML} 2019, 9-15 June 2019, Long Beach, California, {USA}}, \bibinfo{publisher}{{PMLR}}. pp. \bibinfo{pages}{4212--4221}.
\newblock \URLprefix \url{http://proceedings.mlr.press/v97/ma19a.html}.
\bibitem[{Magister et~al.(2023)Magister, Barbiero, Kazhdan, Siciliano, Ciravegna, Silvestri, Jamnik and Li{\`{o}}}]{xai_concept_distill}
\bibinfo{author}{Magister, L.C.}, \bibinfo{author}{Barbiero, P.}, \bibinfo{author}{Kazhdan, D.}, \bibinfo{author}{Siciliano, F.}, \bibinfo{author}{Ciravegna, G.}, \bibinfo{author}{Silvestri, F.}, \bibinfo{author}{Jamnik, M.}, \bibinfo{author}{Li{\`{o}}, P.}, \bibinfo{year}{2023}.
\newblock \bibinfo{title}{Concept distillation in graph neural networks}, in: \bibinfo{editor}{Longo, L.} (Ed.), \bibinfo{booktitle}{Explainable Artificial Intelligence - First World Conference, xAI 2023, Lisbon, Portugal, July 26-28, 2023, Proceedings, Part {III}}, \bibinfo{publisher}{Springer}. pp. \bibinfo{pages}{233--255}.
\newblock \URLprefix \url{https://doi.org/10.1007/978-3-031-44070-0\_12}, \DOIprefix\doi{10.1007/978-3-031-44070-0\_12}.
\bibitem[{Miao et~al.(2022)Miao, Liu and Li}]{miao2022interpretable}
\bibinfo{author}{Miao, S.}, \bibinfo{author}{Liu, M.}, \bibinfo{author}{Li, P.}, \bibinfo{year}{2022}.
\newblock \bibinfo{title}{Interpretable and generalizable graph learning via stochastic attention mechanism}, in: \bibinfo{booktitle}{International Conference on Machine Learning}, \bibinfo{organization}{PMLR}. pp. \bibinfo{pages}{15524--15543}.
\bibitem[{Mo et~al.(2023)Mo, Tang and Liu}]{relation_aware_hgnn_rel_predict}
\bibinfo{author}{Mo, X.}, \bibinfo{author}{Tang, R.}, \bibinfo{author}{Liu, H.}, \bibinfo{year}{2023}.
\newblock \bibinfo{title}{A relation-aware heterogeneous graph convolutional network for relationship prediction}.
\newblock \bibinfo{journal}{INFORMATION SCIENCES} \bibinfo{volume}{623}, \bibinfo{pages}{311--323}.
\newblock \DOIprefix\doi{10.1016/j.ins.2022.12.059}.
\bibitem[{Moraffah et~al.(2020)Moraffah, Karami, Guo, Raglin and Liu}]{moraffah2020causal}
\bibinfo{author}{Moraffah, R.}, \bibinfo{author}{Karami, M.}, \bibinfo{author}{Guo, R.}, \bibinfo{author}{Raglin, A.}, \bibinfo{author}{Liu, H.}, \bibinfo{year}{2020}.
\newblock \bibinfo{title}{Causal interpretability for machine learning-problems, methods and evaluation}.
\newblock \bibinfo{journal}{ACM SIGKDD Explorations Newsletter} \bibinfo{volume}{22}, \bibinfo{pages}{18--33}.
\bibitem[{Newman and Park(2003)}]{newman2003social}
\bibinfo{author}{Newman, M.E.}, \bibinfo{author}{Park, J.}, \bibinfo{year}{2003}.
\newblock \bibinfo{title}{Why social networks are different from other types of networks}.
\newblock \bibinfo{journal}{Physical review E} \bibinfo{volume}{68}, \bibinfo{pages}{036122}.
\bibitem[{Park et~al.(2022)Park, Song and Yang}]{GraphENS}
\bibinfo{author}{Park, J.}, \bibinfo{author}{Song, J.}, \bibinfo{author}{Yang, E.}, \bibinfo{year}{2022}.
\newblock \bibinfo{title}{Graphens: Neighbor-aware ego network synthesis for class-imbalanced node classification}, in: \bibinfo{booktitle}{The Tenth International Conference on Learning Representations, {ICLR} 2022, Virtual Event, April 25-29, 2022}, \bibinfo{publisher}{OpenReview.net}.
\newblock \URLprefix \url{https://openreview.net/forum?id=MXEl7i-iru}.
\bibitem[{Pawlowski et~al.(2020)Pawlowski, Coelho~de Castro and Glocker}]{deep_scm_nips}
\bibinfo{author}{Pawlowski, N.}, \bibinfo{author}{Coelho~de Castro, D.}, \bibinfo{author}{Glocker, B.}, \bibinfo{year}{2020}.
\newblock \bibinfo{title}{Deep structural causal models for tractable counterfactual inference}, in: \bibinfo{editor}{Larochelle, H.}, \bibinfo{editor}{Ranzato, M.}, \bibinfo{editor}{Hadsell, R.}, \bibinfo{editor}{Balcan, M.}, \bibinfo{editor}{Lin, H.} (Eds.), \bibinfo{booktitle}{Advances in Neural Information Processing Systems}, \bibinfo{publisher}{Curran Associates, Inc.}. pp. \bibinfo{pages}{857--869}.
\newblock \URLprefix \url{https://proceedings.neurips.cc/paper/2020/file/0987b8b338d6c90bbedd8631bc499221-Paper.pdf}.
\bibitem[{Pearl(2009)}]{pearl2009causality}
\bibinfo{author}{Pearl, J.}, \bibinfo{year}{2009}.
\newblock \bibinfo{title}{Causality}.
\newblock \bibinfo{publisher}{Cambridge university press}.
\bibitem[{Pennington et~al.(2014)Pennington, Socher and Manning}]{pennington-etal-2014-glove}
\bibinfo{author}{Pennington, J.}, \bibinfo{author}{Socher, R.}, \bibinfo{author}{Manning, C.}, \bibinfo{year}{2014}.
\newblock \bibinfo{title}{{G}lo{V}e: Global vectors for word representation}, in: \bibinfo{booktitle}{Proceedings of the 2014 Conference on Empirical Methods in Natural Language Processing ({EMNLP})}, \bibinfo{publisher}{Association for Computational Linguistics}, \bibinfo{address}{Doha, Qatar}. pp. \bibinfo{pages}{1532--1543}.
\newblock \URLprefix \url{https://aclanthology.org/D14-1162}, \DOIprefix\doi{10.3115/v1/D14-1162}.
\bibitem[{Peters et~al.(2017)Peters, Janzing and Sch{\"o}lkopf}]{peters2017elements}
\bibinfo{author}{Peters, J.}, \bibinfo{author}{Janzing, D.}, \bibinfo{author}{Sch{\"o}lkopf, B.}, \bibinfo{year}{2017}.
\newblock \bibinfo{title}{Elements of causal inference: foundations and learning algorithms}.
\newblock \bibinfo{publisher}{The MIT Press}.
\bibitem[{Pope et~al.(2019)Pope, Kolouri, Rostami, Martin and Hoffmann}]{cvpr_grad_cam}
\bibinfo{author}{Pope, P.E.}, \bibinfo{author}{Kolouri, S.}, \bibinfo{author}{Rostami, M.}, \bibinfo{author}{Martin, C.E.}, \bibinfo{author}{Hoffmann, H.}, \bibinfo{year}{2019}.
\newblock \bibinfo{title}{Explainability methods for graph convolutional neural networks}, in: \bibinfo{booktitle}{{IEEE} Conference on Computer Vision and Pattern Recognition, {CVPR} 2019, Long Beach, CA, USA, June 16-20, 2019}, \bibinfo{publisher}{Computer Vision Foundation / {IEEE}}. pp. \bibinfo{pages}{10772--10781}.
\newblock \URLprefix \url{http://openaccess.thecvf.com/content\_CVPR\_2019/html/Pope\_Explainability\_Methods\_for\_Graph\_Convolutional\_Neural\_Networks\_CVPR\_2019\_paper.html}, \DOIprefix\doi{10.1109/CVPR.2019.01103}.
\bibitem[{Qiao et~al.(2020)Qiao, Luo, Li, Tian and Ma}]{lichenliang_IPM_hgn}
\bibinfo{author}{Qiao, Y.}, \bibinfo{author}{Luo, X.}, \bibinfo{author}{Li, C.}, \bibinfo{author}{Tian, H.}, \bibinfo{author}{Ma, J.}, \bibinfo{year}{2020}.
\newblock \bibinfo{title}{Heterogeneous graph-based joint representation learning for users and pois in location-based social network}.
\newblock \bibinfo{journal}{Information Processing \& Management} \bibinfo{volume}{57}, \bibinfo{pages}{102151}.
\newblock \URLprefix \url{https://www.sciencedirect.com/science/article/pii/S0306457319305114}, \DOIprefix\doi{https://doi.org/10.1016/j.ipm.2019.102151}.
\bibitem[{Ragno et~al.(2022)Ragno, La~Rosa and Capobianco}]{tpami_prototype_gnn}
\bibinfo{author}{Ragno, A.}, \bibinfo{author}{La~Rosa, B.}, \bibinfo{author}{Capobianco, R.}, \bibinfo{year}{2022}.
\newblock \bibinfo{title}{Prototype-based interpretable graph neural networks}.
\newblock \bibinfo{journal}{IEEE Transactions on Artificial Intelligence} , \bibinfo{pages}{1--11}\DOIprefix\doi{10.1109/TAI.2022.3222618}.
\bibitem[{Schlichtkrull et~al.(2018)Schlichtkrull, Kipf, Bloem, Van Den~Berg, Titov and Welling}]{rgcn}
\bibinfo{author}{Schlichtkrull, M.}, \bibinfo{author}{Kipf, T.N.}, \bibinfo{author}{Bloem, P.}, \bibinfo{author}{Van Den~Berg, R.}, \bibinfo{author}{Titov, I.}, \bibinfo{author}{Welling, M.}, \bibinfo{year}{2018}.
\newblock \bibinfo{title}{Modeling relational data with graph convolutional networks}, in: \bibinfo{booktitle}{European semantic web conference}, \bibinfo{organization}{Springer}. pp. \bibinfo{pages}{593--607}.
\bibitem[{Schnake et~al.(2022)Schnake, Eberle, Lederer, Nakajima, Sch{\"{u}}tt, M{\"{u}}ller and Montavon}]{tpami_gnn_lrp}
\bibinfo{author}{Schnake, T.}, \bibinfo{author}{Eberle, O.}, \bibinfo{author}{Lederer, J.}, \bibinfo{author}{Nakajima, S.}, \bibinfo{author}{Sch{\"{u}}tt, K.T.}, \bibinfo{author}{M{\"{u}}ller, K.}, \bibinfo{author}{Montavon, G.}, \bibinfo{year}{2022}.
\newblock \bibinfo{title}{Higher-order explanations of graph neural networks via relevant walks}.
\newblock \bibinfo{journal}{{IEEE} Trans. Pattern Anal. Mach. Intell.} \bibinfo{volume}{44}, \bibinfo{pages}{7581--7596}.
\newblock \URLprefix \url{https://doi.org/10.1109/TPAMI.2021.3115452}, \DOIprefix\doi{10.1109/TPAMI.2021.3115452}.
\bibitem[{Shi et~al.(2021)Shi, Huang, Feng, Zhong, Wang and Sun}]{transformerconv}
\bibinfo{author}{Shi, Y.}, \bibinfo{author}{Huang, Z.}, \bibinfo{author}{Feng, S.}, \bibinfo{author}{Zhong, H.}, \bibinfo{author}{Wang, W.}, \bibinfo{author}{Sun, Y.}, \bibinfo{year}{2021}.
\newblock \bibinfo{title}{Masked label prediction: Unified message passing model for semi-supervised classification}, in: \bibinfo{editor}{Zhou, Z.} (Ed.), \bibinfo{booktitle}{Proceedings of the Thirtieth International Joint Conference on Artificial Intelligence, {IJCAI} 2021, Virtual Event / Montreal, Canada, 19-27 August 2021}, \bibinfo{publisher}{ijcai.org}. pp. \bibinfo{pages}{1548--1554}.
\newblock \URLprefix \url{https://doi.org/10.24963/ijcai.2021/214}, \DOIprefix\doi{10.24963/ijcai.2021/214}.
\bibitem[{Strobl(2022)}]{ML_mix_dag}
\bibinfo{author}{Strobl, V, E.}, \bibinfo{year}{2022}.
\newblock \bibinfo{title}{Causal discovery with a mixture of dags}.
\newblock \bibinfo{journal}{MACHINE LEARNING} \DOIprefix\doi{10.1007/s10994-022-06159-y}.
\bibitem[{Sun et~al.(2011)Sun, Han, Yan, Yu and Wu}]{sun2011pathsim}
\bibinfo{author}{Sun, Y.}, \bibinfo{author}{Han, J.}, \bibinfo{author}{Yan, X.}, \bibinfo{author}{Yu, P.S.}, \bibinfo{author}{Wu, T.}, \bibinfo{year}{2011}.
\newblock \bibinfo{title}{Pathsim: Meta path-based top-k similarity search in heterogeneous information networks}.
\newblock \bibinfo{journal}{Proceedings of the VLDB Endowment} \bibinfo{volume}{4}, \bibinfo{pages}{992--1003}.
\bibitem[{Tan et~al.(2022)Tan, Li, Wang and Chen}]{FinHGNN}
\bibinfo{author}{Tan, J.}, \bibinfo{author}{Li, Q.}, \bibinfo{author}{Wang, J.}, \bibinfo{author}{Chen, J.}, \bibinfo{year}{2022}.
\newblock \bibinfo{title}{Finhgnn: A conditional heterogeneous graph learning to address relational attributes for stock predictions}.
\newblock \bibinfo{journal}{INFORMATION SCIENCES} \bibinfo{volume}{618}, \bibinfo{pages}{317--335}.
\newblock \DOIprefix\doi{10.1016/j.ins.2022.11.013}.
\bibitem[{Vashishth et~al.(2020)Vashishth, Sanyal, Nitin and Talukdar}]{compgcn}
\bibinfo{author}{Vashishth, S.}, \bibinfo{author}{Sanyal, S.}, \bibinfo{author}{Nitin, V.}, \bibinfo{author}{Talukdar, P.P.}, \bibinfo{year}{2020}.
\newblock \bibinfo{title}{Composition-based multi-relational graph convolutional networks}, in: \bibinfo{booktitle}{8th International Conference on Learning Representations, {ICLR} 2020, Addis Ababa, Ethiopia, April 26-30, 2020}, \bibinfo{publisher}{OpenReview.net}.
\newblock \URLprefix \url{https://openreview.net/forum?id=BylA\_C4tPr}.
\bibitem[{Vaswani et~al.(2017)Vaswani, Shazeer, Parmar, Uszkoreit, Jones, Gomez, Kaiser and Polosukhin}]{vaswani2017attention}
\bibinfo{author}{Vaswani, A.}, \bibinfo{author}{Shazeer, N.}, \bibinfo{author}{Parmar, N.}, \bibinfo{author}{Uszkoreit, J.}, \bibinfo{author}{Jones, L.}, \bibinfo{author}{Gomez, A.N.}, \bibinfo{author}{Kaiser, {\L}.}, \bibinfo{author}{Polosukhin, I.}, \bibinfo{year}{2017}.
\newblock \bibinfo{title}{Attention is all you need}, in: \bibinfo{booktitle}{Advances in neural information processing systems}, pp. \bibinfo{pages}{5998--6008}.
\bibitem[{VDong et~al.(2020)VDong, Hu, Wang, Sun and Tang}]{dong2020heterogeneous}
\bibinfo{author}{VDong, Y.}, \bibinfo{author}{Hu, Z.}, \bibinfo{author}{Wang, K.}, \bibinfo{author}{Sun, Y.}, \bibinfo{author}{Tang, J.}, \bibinfo{year}{2020}.
\newblock \bibinfo{title}{Heterogeneous network representation learning}, in: \bibinfo{booktitle}{Proceedings of the Twenty-Ninth International Joint Conference on Artificial Intelligence (IJCAI-20)}, pp. \bibinfo{pages}{4861--4867}.
\bibitem[{Velickovic et~al.(2018)Velickovic, Cucurull, Casanova, Romero, Li{\`{o}} and Bengio}]{gat}
\bibinfo{author}{Velickovic, P.}, \bibinfo{author}{Cucurull, G.}, \bibinfo{author}{Casanova, A.}, \bibinfo{author}{Romero, A.}, \bibinfo{author}{Li{\`{o}}, P.}, \bibinfo{author}{Bengio, Y.}, \bibinfo{year}{2018}.
\newblock \bibinfo{title}{Graph attention networks}, in: \bibinfo{booktitle}{6th International Conference on Learning Representations, {ICLR} 2018, Vancouver, BC, Canada, April 30 - May 3, 2018, Conference Track Proceedings}, \bibinfo{publisher}{OpenReview.net}.
\newblock \URLprefix \url{https://openreview.net/forum?id=rJXMpikCZ}.
\bibitem[{Vowels et~al.(2022)Vowels, Camgoz and Bowden}]{dag_review}
\bibinfo{author}{Vowels, M.J.}, \bibinfo{author}{Camgoz, N.C.}, \bibinfo{author}{Bowden, R.}, \bibinfo{year}{2022}.
\newblock \bibinfo{title}{D’ya like dags? a survey on structure learning and causal discovery}.
\newblock \bibinfo{journal}{ACM Comput. Surv.} \bibinfo{volume}{55}.
\newblock \URLprefix \url{https://doi.org/10.1145/3527154}, \DOIprefix\doi{10.1145/3527154}.
\bibitem[{Wan et~al.(2022)Wan, Yuan, Zhan and Chen}]{ipm-Robust-gcn}
\bibinfo{author}{Wan, Y.}, \bibinfo{author}{Yuan, C.}, \bibinfo{author}{Zhan, M.}, \bibinfo{author}{Chen, L.}, \bibinfo{year}{2022}.
\newblock \bibinfo{title}{Robust graph learning with graph convolutional network}.
\newblock \bibinfo{journal}{Information Processing \& Management} \bibinfo{volume}{59}, \bibinfo{pages}{102916}.
\newblock \URLprefix \url{https://www.sciencedirect.com/science/article/pii/S0306457322000413}, \DOIprefix\doi{https://doi.org/10.1016/j.ipm.2022.102916}.
\bibitem[{Wang et~al.(2023a)Wang, Mi, Guo and Hu}]{hgnn_link_meta_learning_ipm}
\bibinfo{author}{Wang, H.}, \bibinfo{author}{Mi, J.}, \bibinfo{author}{Guo, X.}, \bibinfo{author}{Hu, P.}, \bibinfo{year}{2023}a.
\newblock \bibinfo{title}{Meta-learning adaptation network for few-shot link prediction in heterogeneous social networks}.
\newblock \bibinfo{journal}{Information Processing \& Management} \bibinfo{volume}{60}, \bibinfo{pages}{103418}.
\newblock \URLprefix \url{https://www.sciencedirect.com/science/article/pii/S0306457323001553}, \DOIprefix\doi{https://doi.org/10.1016/j.ipm.2023.103418}.
\bibitem[{Wang et~al.(2019a)Wang, Zheng, Ye, Gan, Li, Song, Zhou, Ma, Yu, Gai, Xiao, He, Karypis, Li and Zhang}]{dgl}
\bibinfo{author}{Wang, M.}, \bibinfo{author}{Zheng, D.}, \bibinfo{author}{Ye, Z.}, \bibinfo{author}{Gan, Q.}, \bibinfo{author}{Li, M.}, \bibinfo{author}{Song, X.}, \bibinfo{author}{Zhou, J.}, \bibinfo{author}{Ma, C.}, \bibinfo{author}{Yu, L.}, \bibinfo{author}{Gai, Y.}, \bibinfo{author}{Xiao, T.}, \bibinfo{author}{He, T.}, \bibinfo{author}{Karypis, G.}, \bibinfo{author}{Li, J.}, \bibinfo{author}{Zhang, Z.}, \bibinfo{year}{2019}a.
\newblock \bibinfo{title}{Deep graph library: A graph-centric, highly-performant package for graph neural networks}.
\newblock \bibinfo{journal}{arXiv preprint arXiv:1909.01315} .
\bibitem[{Wang et~al.(2021)Wang, Mao, Lu, Cao and Li}]{joi_phrase_intersection}
\bibinfo{author}{Wang, S.}, \bibinfo{author}{Mao, J.}, \bibinfo{author}{Lu, K.}, \bibinfo{author}{Cao, Y.}, \bibinfo{author}{Li, G.}, \bibinfo{year}{2021}.
\newblock \bibinfo{title}{Understanding interdisciplinary knowledge integration through citance analysis: A case study on ehealth}.
\newblock \bibinfo{journal}{Journal of Informetrics} \bibinfo{volume}{15}, \bibinfo{pages}{101214}.
\newblock \URLprefix \url{https://www.sciencedirect.com/science/article/pii/S1751157721000857}, \DOIprefix\doi{https://doi.org/10.1016/j.joi.2021.101214}.
\bibitem[{Wang et~al.(2019b)Wang, Ji, Shi, Wang, Ye, Cui and Yu}]{han}
\bibinfo{author}{Wang, X.}, \bibinfo{author}{Ji, H.}, \bibinfo{author}{Shi, C.}, \bibinfo{author}{Wang, B.}, \bibinfo{author}{Ye, Y.}, \bibinfo{author}{Cui, P.}, \bibinfo{author}{Yu, P.S.}, \bibinfo{year}{2019}b.
\newblock \bibinfo{title}{Heterogeneous graph attention network}, in: \bibinfo{booktitle}{The World Wide Web Conference}, pp. \bibinfo{pages}{2022--2032}.
\bibitem[{Wang et~al.(2022a)Wang, Ma, Guo, Jiang, Liu and Xu}]{IPM2023MOOC}
\bibinfo{author}{Wang, X.}, \bibinfo{author}{Ma, W.}, \bibinfo{author}{Guo, L.}, \bibinfo{author}{Jiang, H.}, \bibinfo{author}{Liu, F.}, \bibinfo{author}{Xu, C.}, \bibinfo{year}{2022}a.
\newblock \bibinfo{title}{Hgnn: Hyperedge-based graph neural network for mooc course recommendation}.
\newblock \bibinfo{journal}{Information Processing \& Management} \bibinfo{volume}{59}, \bibinfo{pages}{102938}.
\newblock \URLprefix \url{https://www.sciencedirect.com/science/article/pii/S0306457322000607}, \DOIprefix\doi{https://doi.org/10.1016/j.ipm.2022.102938}.
\bibitem[{Wang et~al.(2022b)Wang, Wu, Zhang, Feng, He and Chua}]{wang2022reinforced}
\bibinfo{author}{Wang, X.}, \bibinfo{author}{Wu, Y.}, \bibinfo{author}{Zhang, A.}, \bibinfo{author}{Feng, F.}, \bibinfo{author}{He, X.}, \bibinfo{author}{Chua, T.S.}, \bibinfo{year}{2022}b.
\newblock \bibinfo{title}{Reinforced causal explainer for graph neural networks}.
\newblock \bibinfo{journal}{IEEE Transactions on Pattern Analysis and Machine Intelligence} .
\bibitem[{Wang et~al.(2023b)Wang, Li, Ota, Dong and Wu}]{IPM_VQA_HGNN}
\bibinfo{author}{Wang, Z.}, \bibinfo{author}{Li, F.}, \bibinfo{author}{Ota, K.}, \bibinfo{author}{Dong, M.}, \bibinfo{author}{Wu, B.}, \bibinfo{year}{2023}b.
\newblock \bibinfo{title}{Regr: Relation-aware graph reasoning framework for video question answering}.
\newblock \bibinfo{journal}{INFORMATION PROCESSING \& MANAGEMENT} \bibinfo{volume}{60}.
\newblock \DOIprefix\doi{10.1016/j.ipm.2023.103375}.
\bibitem[{Wei et~al.(2020)Wei, Gao and Yu}]{nips20_dag_nofears}
\bibinfo{author}{Wei, D.}, \bibinfo{author}{Gao, T.}, \bibinfo{author}{Yu, Y.}, \bibinfo{year}{2020}.
\newblock \bibinfo{title}{Dags with no fears: A closer look at continuous optimization for learning bayesian networks}, in: \bibinfo{editor}{Larochelle, H.}, \bibinfo{editor}{Ranzato, M.}, \bibinfo{editor}{Hadsell, R.}, \bibinfo{editor}{Balcan, M.}, \bibinfo{editor}{Lin, H.} (Eds.), \bibinfo{booktitle}{Advances in Neural Information Processing Systems}, \bibinfo{publisher}{Curran Associates, Inc.}. pp. \bibinfo{pages}{3895--3906}.
\newblock \URLprefix \url{https://proceedings.neurips.cc/paper/2020/file/28a7602724ba16600d5ccc644c19bf18-Paper.pdf}.
\bibitem[{Xian et~al.(2022)Xian, Li, Tang and Ma}]{ieee_reviwer2_path_selection}
\bibinfo{author}{Xian, T.}, \bibinfo{author}{Li, Z.}, \bibinfo{author}{Tang, Z.}, \bibinfo{author}{Ma, H.}, \bibinfo{year}{2022}.
\newblock \bibinfo{title}{Adaptive path selection for dynamic image captioning}.
\newblock \bibinfo{journal}{{IEEE} Trans. Circuits Syst. Video Technol.} \bibinfo{volume}{32}, \bibinfo{pages}{5762--5775}.
\newblock \URLprefix \url{https://doi.org/10.1109/TCSVT.2022.3155795}, \DOIprefix\doi{10.1109/TCSVT.2022.3155795}.
\bibitem[{Xie et~al.(2023)Xie, Li, Tang, Yao and Ma}]{ipm_reviewer2_image_text_graph}
\bibinfo{author}{Xie, X.}, \bibinfo{author}{Li, Z.}, \bibinfo{author}{Tang, Z.}, \bibinfo{author}{Yao, D.}, \bibinfo{author}{Ma, H.}, \bibinfo{year}{2023}.
\newblock \bibinfo{title}{Unifying knowledge iterative dissemination and relational reconstruction network for image–text matching}.
\newblock \bibinfo{journal}{Information Processing \& Management} \bibinfo{volume}{60}, \bibinfo{pages}{103154}.
\newblock \URLprefix \url{https://www.sciencedirect.com/science/article/pii/S0306457322002552}, \DOIprefix\doi{https://doi.org/10.1016/j.ipm.2022.103154}.
\bibitem[{Xie et~al.(2022)Xie, Zhu, Liu, Zhou and Huang}]{IPM_KG_representation}
\bibinfo{author}{Xie, Z.}, \bibinfo{author}{Zhu, R.}, \bibinfo{author}{Liu, J.}, \bibinfo{author}{Zhou, G.}, \bibinfo{author}{Huang, J.X.}, \bibinfo{year}{2022}.
\newblock \bibinfo{title}{An efficiency relation-specific graph transformation network for knowledge graph representation learning}.
\newblock \bibinfo{journal}{INFORMATION PROCESSING \& MANAGEMENT} \bibinfo{volume}{59}.
\newblock \DOIprefix\doi{10.1016/j.ipm.2022.103076}.
\bibitem[{Yang et~al.(2023a)Yang, Yu, Cao, Liu, Wang and Li}]{TKDE_Causal_Representations}
\bibinfo{author}{Yang, S.}, \bibinfo{author}{Yu, K.}, \bibinfo{author}{Cao, F.}, \bibinfo{author}{Liu, L.}, \bibinfo{author}{Wang, H.}, \bibinfo{author}{Li, J.}, \bibinfo{year}{2023}a.
\newblock \bibinfo{title}{Learning causal representations for robust domain adaptation}.
\newblock \bibinfo{journal}{IEEE TRANSACTIONS ON KNOWLEDGE AND DATA ENGINEERING} \bibinfo{volume}{35}, \bibinfo{pages}{2750--2764}.
\newblock \DOIprefix\doi{10.1109/TKDE.2021.3119185}.
\bibitem[{Yang et~al.(2023b)Yang, Yan, Pan, Ye and Fan}]{sehgnn}
\bibinfo{author}{Yang, X.}, \bibinfo{author}{Yan, M.}, \bibinfo{author}{Pan, S.}, \bibinfo{author}{Ye, X.}, \bibinfo{author}{Fan, D.}, \bibinfo{year}{2023}b.
\newblock \bibinfo{title}{Simple and efficient heterogeneous graph neural network}.
\newblock \bibinfo{journal}{Proceedings of the AAAI Conference on Artificial Intelligence} \bibinfo{volume}{37}, \bibinfo{pages}{10816--10824}.
\newblock \URLprefix \url{https://ojs.aaai.org/index.php/AAAI/article/view/26283}, \DOIprefix\doi{10.1609/aaai.v37i9.26283}.
\bibitem[{Ying et~al.(2019)Ying, Bourgeois, You, Zitnik and Leskovec}]{ying2019gnnexplainer}
\bibinfo{author}{Ying, Z.}, \bibinfo{author}{Bourgeois, D.}, \bibinfo{author}{You, J.}, \bibinfo{author}{Zitnik, M.}, \bibinfo{author}{Leskovec, J.}, \bibinfo{year}{2019}.
\newblock \bibinfo{title}{Gnnexplainer: Generating explanations for graph neural networks}.
\newblock \bibinfo{journal}{Advances in neural information processing systems} \bibinfo{volume}{32}.
\bibitem[{You et~al.(2020)You, Chen, Sui, Chen, Wang and Shen}]{nips_graphcl}
\bibinfo{author}{You, Y.}, \bibinfo{author}{Chen, T.}, \bibinfo{author}{Sui, Y.}, \bibinfo{author}{Chen, T.}, \bibinfo{author}{Wang, Z.}, \bibinfo{author}{Shen, Y.}, \bibinfo{year}{2020}.
\newblock \bibinfo{title}{Graph contrastive learning with augmentations}, in: \bibinfo{editor}{Larochelle, H.}, \bibinfo{editor}{Ranzato, M.}, \bibinfo{editor}{Hadsell, R.}, \bibinfo{editor}{Balcan, M.}, \bibinfo{editor}{Lin, H.} (Eds.), \bibinfo{booktitle}{Advances in Neural Information Processing Systems 33: Annual Conference on Neural Information Processing Systems 2020, NeurIPS 2020, December 6-12, 2020, virtual}.
\newblock \URLprefix \url{https://proceedings.neurips.cc/paper/2020/hash/3fe230348e9a12c13120749e3f9fa4cd-Abstract.html}.
\bibitem[{Yu et~al.(2021)Yu, Xu, Rong, Bian, Huang and He}]{yugraph}
\bibinfo{author}{Yu, J.}, \bibinfo{author}{Xu, T.}, \bibinfo{author}{Rong, Y.}, \bibinfo{author}{Bian, Y.}, \bibinfo{author}{Huang, J.}, \bibinfo{author}{He, R.}, \bibinfo{year}{2021}.
\newblock \bibinfo{title}{Graph information bottleneck for subgraph recognition}, in: \bibinfo{booktitle}{International Conference on Learning Representations}.
\bibitem[{Yu et~al.(2023)Yu, Sun, Du, Liu, Lv and Xiong}]{R-HGNN}
\bibinfo{author}{Yu, L.}, \bibinfo{author}{Sun, L.}, \bibinfo{author}{Du, B.}, \bibinfo{author}{Liu, C.}, \bibinfo{author}{Lv, W.}, \bibinfo{author}{Xiong, H.}, \bibinfo{year}{2023}.
\newblock \bibinfo{title}{Heterogeneous graph representation learning with relation awareness}.
\newblock \bibinfo{journal}{IEEE TRANSACTIONS ON KNOWLEDGE AND DATA ENGINEERING} \bibinfo{volume}{35}, \bibinfo{pages}{5935--5947}.
\newblock \DOIprefix\doi{10.1109/TKDE.2022.3160208}.
\bibitem[{Yu et~al.(2019)Yu, Chen, Gao and Yu}]{dag_gnn}
\bibinfo{author}{Yu, Y.}, \bibinfo{author}{Chen, J.}, \bibinfo{author}{Gao, T.}, \bibinfo{author}{Yu, M.}, \bibinfo{year}{2019}.
\newblock \bibinfo{title}{{DAG}-{GNN}: {DAG} structure learning with graph neural networks}, in: \bibinfo{editor}{Chaudhuri, K.}, \bibinfo{editor}{Salakhutdinov, R.} (Eds.), \bibinfo{booktitle}{Proceedings of the 36th International Conference on Machine Learning}, \bibinfo{publisher}{PMLR}. pp. \bibinfo{pages}{7154--7163}.
\newblock \URLprefix \url{https://proceedings.mlr.press/v97/yu19a.html}.
\bibitem[{Yuan et~al.(2020)Yuan, Tang, Hu and Ji}]{xgnn}
\bibinfo{author}{Yuan, H.}, \bibinfo{author}{Tang, J.}, \bibinfo{author}{Hu, X.}, \bibinfo{author}{Ji, S.}, \bibinfo{year}{2020}.
\newblock \bibinfo{title}{{XGNN:} towards model-level explanations of graph neural networks}, in: \bibinfo{editor}{Gupta, R.}, \bibinfo{editor}{Liu, Y.}, \bibinfo{editor}{Tang, J.}, \bibinfo{editor}{Prakash, B.A.} (Eds.), \bibinfo{booktitle}{{KDD} '20: The 26th {ACM} {SIGKDD} Conference on Knowledge Discovery and Data Mining, Virtual Event, CA, USA, August 23-27, 2020}, \bibinfo{publisher}{{ACM}}. pp. \bibinfo{pages}{430--438}.
\newblock \URLprefix \url{https://doi.org/10.1145/3394486.3403085}, \DOIprefix\doi{10.1145/3394486.3403085}.
\bibitem[{Yuan et~al.(2023)Yuan, Yu, Gui and Ji}]{tpami_explainability_survey}
\bibinfo{author}{Yuan, H.}, \bibinfo{author}{Yu, H.}, \bibinfo{author}{Gui, S.}, \bibinfo{author}{Ji, S.}, \bibinfo{year}{2023}.
\newblock \bibinfo{title}{Explainability in graph neural networks: A taxonomic survey}.
\newblock \bibinfo{journal}{IEEE Transactions on Pattern Analysis and Machine Intelligence} \bibinfo{volume}{45}, \bibinfo{pages}{5782--5799}.
\newblock \DOIprefix\doi{10.1109/TPAMI.2022.3204236}.
\bibitem[{Yun et~al.(2019)Yun, Jeong, Kim, Kang and Kim}]{gtn}
\bibinfo{author}{Yun, S.}, \bibinfo{author}{Jeong, M.}, \bibinfo{author}{Kim, R.}, \bibinfo{author}{Kang, J.}, \bibinfo{author}{Kim, H.J.}, \bibinfo{year}{2019}.
\newblock \bibinfo{title}{Graph transformer networks}.
\newblock \bibinfo{journal}{Advances in Neural Information Processing Systems} \bibinfo{volume}{32}, \bibinfo{pages}{11983--11993}.
\bibitem[{Zeng et~al.(2023)Zeng, Li, Tang, Chen and Ma}]{esa_reviewer2_hgnn_sentiment}
\bibinfo{author}{Zeng, Y.}, \bibinfo{author}{Li, Z.}, \bibinfo{author}{Tang, Z.}, \bibinfo{author}{Chen, Z.}, \bibinfo{author}{Ma, H.}, \bibinfo{year}{2023}.
\newblock \bibinfo{title}{Heterogeneous graph convolution based on in-domain self-supervision for multimodal sentiment analysis}.
\newblock \bibinfo{journal}{Expert Systems with Applications} \bibinfo{volume}{213}, \bibinfo{pages}{119240}.
\newblock \URLprefix \url{https://www.sciencedirect.com/science/article/pii/S0957417422022588}, \DOIprefix\doi{https://doi.org/10.1016/j.eswa.2022.119240}.
\bibitem[{Zhai et~al.(2023)Zhai, Yang and Zhang}]{ipm2023causal}
\bibinfo{author}{Zhai, P.}, \bibinfo{author}{Yang, Y.}, \bibinfo{author}{Zhang, C.}, \bibinfo{year}{2023}.
\newblock \bibinfo{title}{Causality-based ctr prediction using graph neural networks}.
\newblock \bibinfo{journal}{Information Processing \& Management} \bibinfo{volume}{60}, \bibinfo{pages}{103137}.
\newblock \URLprefix \url{https://www.sciencedirect.com/science/article/pii/S0306457322002382}, \DOIprefix\doi{https://doi.org/10.1016/j.ipm.2022.103137}.
\bibitem[{Zhang et~al.(2019)Zhang, Song, Huang, Swami and Chawla}]{hetgnn}
\bibinfo{author}{Zhang, C.}, \bibinfo{author}{Song, D.}, \bibinfo{author}{Huang, C.}, \bibinfo{author}{Swami, A.}, \bibinfo{author}{Chawla, N.V.}, \bibinfo{year}{2019}.
\newblock \bibinfo{title}{Heterogeneous graph neural network}, in: \bibinfo{booktitle}{Proceedings of the 25th ACM SIGKDD International Conference on Knowledge Discovery \& Data Mining}, pp. \bibinfo{pages}{793--803}.
\bibitem[{Zhang et~al.(2021)Zhang, Cui, Xu, Zhou, He and Shen}]{zhang2021deep}
\bibinfo{author}{Zhang, X.}, \bibinfo{author}{Cui, P.}, \bibinfo{author}{Xu, R.}, \bibinfo{author}{Zhou, L.}, \bibinfo{author}{He, Y.}, \bibinfo{author}{Shen, Z.}, \bibinfo{year}{2021}.
\newblock \bibinfo{title}{Deep stable learning for out-of-distribution generalization}, in: \bibinfo{booktitle}{Proceedings of the IEEE/CVF Conference on Computer Vision and Pattern Recognition}, pp. \bibinfo{pages}{5372--5382}.
\bibitem[{Zhang et~al.(2022)Zhang, Liu, Wang, Lu and Lee}]{aaai_ProtGNN}
\bibinfo{author}{Zhang, Z.}, \bibinfo{author}{Liu, Q.}, \bibinfo{author}{Wang, H.}, \bibinfo{author}{Lu, C.}, \bibinfo{author}{Lee, C.}, \bibinfo{year}{2022}.
\newblock \bibinfo{title}{Protgnn: Towards self-explaining graph neural networks}, in: \bibinfo{booktitle}{Thirty-Sixth {AAAI} Conference on Artificial Intelligence, {AAAI} 2022, Thirty-Fourth Conference on Innovative Applications of Artificial Intelligence, {IAAI} 2022, The Twelveth Symposium on Educational Advances in Artificial Intelligence, {EAAI} 2022 Virtual Event, February 22 - March 1, 2022}, \bibinfo{publisher}{{AAAI} Press}. pp. \bibinfo{pages}{9127--9135}.
\newblock \URLprefix \url{https://doi.org/10.1609/aaai.v36i8.20898}, \DOIprefix\doi{10.1609/AAAI.V36I8.20898}.
\bibitem[{Zhao et~al.(2021)Zhao, Liu, Neves, Woodford, Jiang and Shah}]{aaai_data_aug}
\bibinfo{author}{Zhao, T.}, \bibinfo{author}{Liu, Y.}, \bibinfo{author}{Neves, L.}, \bibinfo{author}{Woodford, O.J.}, \bibinfo{author}{Jiang, M.}, \bibinfo{author}{Shah, N.}, \bibinfo{year}{2021}.
\newblock \bibinfo{title}{Data augmentation for graph neural networks}, in: \bibinfo{booktitle}{Thirty-Fifth {AAAI} Conference on Artificial Intelligence, {AAAI} 2021, Thirty-Third Conference on Innovative Applications of Artificial Intelligence, {IAAI} 2021, The Eleventh Symposium on Educational Advances in Artificial Intelligence, {EAAI} 2021, Virtual Event, February 2-9, 2021}, \bibinfo{publisher}{{AAAI} Press}. pp. \bibinfo{pages}{11015--11023}.
\newblock \URLprefix \url{https://doi.org/10.1609/aaai.v35i12.17315}, \DOIprefix\doi{10.1609/AAAI.V35I12.17315}.
\bibitem[{Zheng et~al.(2018)Zheng, Aragam, Ravikumar and Xing}]{notears}
\bibinfo{author}{Zheng, X.}, \bibinfo{author}{Aragam, B.}, \bibinfo{author}{Ravikumar, P.}, \bibinfo{author}{Xing, E.P.}, \bibinfo{year}{2018}.
\newblock \bibinfo{title}{Dags with {NO} {TEARS:} continuous optimization for structure learning}, in: \bibinfo{editor}{Bengio, S.}, \bibinfo{editor}{Wallach, H.M.}, \bibinfo{editor}{Larochelle, H.}, \bibinfo{editor}{Grauman, K.}, \bibinfo{editor}{Cesa{-}Bianchi, N.}, \bibinfo{editor}{Garnett, R.} (Eds.), \bibinfo{booktitle}{Advances in Neural Information Processing Systems 31: Annual Conference on Neural Information Processing Systems 2018, NeurIPS 2018, December 3-8, 2018, Montr{\'{e}}al, Canada}, pp. \bibinfo{pages}{9492--9503}.
\newblock \URLprefix \url{https://proceedings.neurips.cc/paper/2018/hash/e347c51419ffb23ca3fd5050202f9c3d-Abstract.html}.
\bibitem[{Zheng et~al.(2020)Zheng, Dan, Aragam, Ravikumar and Xing}]{pmlr_noparam_dag}
\bibinfo{author}{Zheng, X.}, \bibinfo{author}{Dan, C.}, \bibinfo{author}{Aragam, B.}, \bibinfo{author}{Ravikumar, P.}, \bibinfo{author}{Xing, E.}, \bibinfo{year}{2020}.
\newblock \bibinfo{title}{Learning sparse nonparametric dags}, in: \bibinfo{editor}{Chiappa, S.}, \bibinfo{editor}{Calandra, R.} (Eds.), \bibinfo{booktitle}{Proceedings of the Twenty Third International Conference on Artificial Intelligence and Statistics}, \bibinfo{publisher}{PMLR}. pp. \bibinfo{pages}{3414--3425}.
\newblock \URLprefix \url{https://proceedings.mlr.press/v108/zheng20a.html}.
\bibitem[{Zhu et~al.(2021)Zhu, Rossi, Rao, Mai, Lipka, Ahmed and Koutra}]{homo_aaai}
\bibinfo{author}{Zhu, J.}, \bibinfo{author}{Rossi, R.A.}, \bibinfo{author}{Rao, A.}, \bibinfo{author}{Mai, T.}, \bibinfo{author}{Lipka, N.}, \bibinfo{author}{Ahmed, N.K.}, \bibinfo{author}{Koutra, D.}, \bibinfo{year}{2021}.
\newblock \bibinfo{title}{Graph neural networks with heterophily}, in: \bibinfo{booktitle}{Thirty-Fifth {AAAI} Conference on Artificial Intelligence, {AAAI} 2021, Thirty-Third Conference on Innovative Applications of Artificial Intelligence, {IAAI} 2021, The Eleventh Symposium on Educational Advances in Artificial Intelligence, {EAAI} 2021, Virtual Event, February 2-9, 2021}, \bibinfo{publisher}{{AAAI} Press}. pp. \bibinfo{pages}{11168--11176}.
\newblock \URLprefix \url{https://ojs.aaai.org/index.php/AAAI/article/view/17332}.
\bibitem[{Zhu et~al.(2020a)Zhu, Yan, Zhao, Heimann, Akoglu and Koutra}]{nips_homo}
\bibinfo{author}{Zhu, J.}, \bibinfo{author}{Yan, Y.}, \bibinfo{author}{Zhao, L.}, \bibinfo{author}{Heimann, M.}, \bibinfo{author}{Akoglu, L.}, \bibinfo{author}{Koutra, D.}, \bibinfo{year}{2020}a.
\newblock \bibinfo{title}{Beyond homophily in graph neural networks: Current limitations and effective designs}, in: \bibinfo{editor}{Larochelle, H.}, \bibinfo{editor}{Ranzato, M.}, \bibinfo{editor}{Hadsell, R.}, \bibinfo{editor}{Balcan, M.}, \bibinfo{editor}{Lin, H.} (Eds.), \bibinfo{booktitle}{Advances in Neural Information Processing Systems}, \bibinfo{publisher}{Curran Associates, Inc.}. pp. \bibinfo{pages}{7793--7804}.
\newblock \URLprefix \url{https://proceedings.neurips.cc/paper/2020/file/58ae23d878a47004366189884c2f8440-Paper.pdf}.
\bibitem[{Zhu et~al.(2020b)Zhu, Ng and Chen}]{2020Causal_huawei_rl}
\bibinfo{author}{Zhu, S.}, \bibinfo{author}{Ng, I.}, \bibinfo{author}{Chen, Z.}, \bibinfo{year}{2020}b.
\newblock \bibinfo{title}{Causal discovery with reinforcement learning}, in: \bibinfo{booktitle}{International Conference on Learning Representations}.
\bibitem[{Zhu et~al.(2019)Zhu, Zhou, Pan, Zhu and Wang}]{rshn}
\bibinfo{author}{Zhu, S.}, \bibinfo{author}{Zhou, C.}, \bibinfo{author}{Pan, S.}, \bibinfo{author}{Zhu, X.}, \bibinfo{author}{Wang, B.}, \bibinfo{year}{2019}.
\newblock \bibinfo{title}{Relation structure-aware heterogeneous graph neural network}, in: \bibinfo{editor}{Wang, J.}, \bibinfo{editor}{Shim, K.}, \bibinfo{editor}{Wu, X.} (Eds.), \bibinfo{booktitle}{2019 {IEEE} International Conference on Data Mining, {ICDM} 2019, Beijing, China, November 8-11, 2019}, \bibinfo{publisher}{{IEEE}}. pp. \bibinfo{pages}{1534--1539}.
\newblock \URLprefix \url{https://doi.org/10.1109/ICDM.2019.00203}, \DOIprefix\doi{10.1109/ICDM.2019.00203}.

\end{thebibliography}
